\documentclass[journal]{IEEEtran}
\usepackage{lineno}
\modulolinenumbers[5]

\usepackage{xpatch}

\usepackage{bm}
\usepackage{array}
\usepackage{graphicx}
\usepackage{amsmath,amssymb,amsthm}
\usepackage{siunitx}
\usepackage{algpseudocode}
\usepackage{algorithmicx}
\usepackage{algorithm}
\usepackage{booktabs}
\usepackage{color}
\usepackage{changepage}
\usepackage{xr}
\usepackage{xr-hyper}
\usepackage{graphicx}%图片设置
\usepackage{subfigure}
\usepackage{caption}%注释设置
\usepackage{multirow}
\usepackage{float}
\usepackage{soul}
\usepackage[hidelinks]{hyperref}
\usepackage[numbers,sort&compress]{natbib}
\usepackage{bigstrut} %表格大竖线
\usepackage[table]{xcolor} %表格单元格颜色
\usepackage{enumitem} %enumerate标签样式\usepackage{listings}
\usepackage{listings} %listing代码
\usepackage[resetlabels]{multibib}
\newcites{supp}{Supplement References}

\definecolor{dkgreen}{rgb}{0,0.5,0}
\definecolor{gray}{rgb}{0.5,0.5,0.5}
\definecolor{mauve}{rgb}{0.58,0,0.82}
\definecolor{dkblue}{rgb}{0,0,0.6}

\AtBeginDocument{%
 \abovedisplayskip=5pt plus 4pt minus 2pt
 \abovedisplayshortskip=5pt plus 4pt minus 4pt
 \belowdisplayskip=5pt plus 4pt minus 2pt
 \belowdisplayshortskip=5pt plus 4pt minus 4pt
}

\ifCLASSINFOpdf
\else
\fi

\begin{document}
\captionsetup{font={footnotesize}}
\captionsetup[table]{labelformat=simple, labelsep=newline, textfont=sc, justification=centering}
\title{MetaDE: Evolving Differential Evolution by Differential Evolution}

\author{Minyang Chen, Chenchen Feng,
        and Ran Cheng

        \thanks{
        Minyang Chen was with the Department of Computer Science and Engineering, Southern University of Science and Technology, Shenzhen 518055, China. E-mail: cmy1223605455@gmail.com. }
        \thanks{
        Chenchen Feng is with the Department of Computer Science and Engineering, Southern University of Science and Technology, Shenzhen 518055, China. E-mail: chenchenfengcn@gmail.com. 
        }
        \thanks{
       Ran Cheng is with the Department of Data Science and Artificial Intelligence, and the Department of Computing, The Hong Kong Polytechnic University, Hong Kong SAR, China. E-mail: ranchengcn@gmail.com. (\emph{Corresponding author: Ran Cheng})
        }
        }% <-this % stops a space

\markboth{Bare Demo of IEEEtran.cls for IEEE Journals}%
{Shell \MakeLowercase{\textit{et al.}}: Bare Demo of IEEEtran.cls for IEEE Journals}




\maketitle

\begin{abstract}
As a cornerstone in the Evolutionary Computation (EC) domain, Differential Evolution (DE) is known for its simplicity and effectiveness in handling challenging black-box optimization problems.
While the advantages of DE are well-recognized, achieving peak performance heavily depends on its hyperparameters such as the mutation factor, crossover probability, and the selection of specific DE strategies.
Traditional approaches to this hyperparameter dilemma have leaned towards parameter tuning or adaptive mechanisms.
However, identifying the optimal settings tailored for specific problems remains a persistent challenge.
In response, we introduce MetaDE, an approach that evolves DE's intrinsic hyperparameters and strategies using DE itself at a meta-level.
A pivotal aspect of MetaDE is a specialized parameterization technique, which endows it with the capability to dynamically modify DE's parameters and strategies throughout the evolutionary process.
To augment computational efficiency, MetaDE incorporates a design that leverages parallel processing through a GPU-accelerated computing framework.
Within such a framework, DE is not just a solver but also an optimizer for its own configurations, thus streamlining the process of hyperparameter optimization and problem-solving into a cohesive and automated workflow.
Extensive evaluations on the CEC2022 benchmark suite demonstrate MetaDE's promising performance.
Moreover, when applied to robot control via evolutionary reinforcement learning, MetaDE also demonstrates promising performance.
The source code of MetaDE is publicly accessible at: \url{https://github.com/EMI-Group/metade}.
\end{abstract}


% \fi
\IEEEpeerreviewmaketitle

\section{Introduction}
\IEEEPARstart{T}{he} Differential Evolution (DE) \cite{DE1996,DEcontest1996,DEusage1996,DE1997} algorithm, introduced by Storn and Price in 1995, has emerged as a cornerstone in the realm of evolutionary computation (EC) for its prowess in addressing complex optimization problems across diverse domains of science and engineering.
DE's comparative advantage over other evolutionary algorithms is evident in its streamlined design, robust performance, and ease of implementation.
Notably, with just three primary control parameters, i.e., scaling factor, crossover rate, and population size, DE operates efficiently.
This minimalistic design, paired with a lower algorithmic complexity, positions DE as an ideal candidate for large-scale optimization problems.
Its influential role in the optimization community is further cemented by its extensive research attention and successful applications over the past decades \cite{DEsurvey2011, DEsurvey2016, DEpapersurvey2020}, with DE and its derivatives often securing top positions in the IEEE Congress on Evolutionary Computation (CEC) competitions.

Despite the well recognized performance, DE is not without limitations.
Particularly, some studies indicate that DE's optimization process may stagnate if it fails to generate offspring solutions superior to their parents \cite{DEStagnation, neriDEsurvey}.
To avert this stagnation, selecting an appropriate parameter configuration to enhance DE's search capabilities becomes crucial.

However, the No Free Lunch (NFL) theorem \cite{NFL} suggests that a universally optimal parameter configuration is unattainable.
For example, while a higher mutation factor may aid in escaping local optima, a lower crossover probability might be preferable for problems with separability characteristics.

To address the intricate challenge of parameter configuration in DE, researchers often gravitate towards two predominant strategies: \emph{parameter control} and \emph{parameter tuning} \cite{param1999,paramTun2012,paramTun2020}.
Parameter control is a dynamic approach wherein the algorithm's parameters are adjusted on-the-fly during its execution.
This adaptability allows the algorithm to respond to the evolving characteristics of the problem landscape, enhancing its chance of finding optimal or near-optimal solutions.
Notably, DE has incorporated this strategy in several of its variants.
For instance, jDE \cite{jDE2006} adjusts the mutation factor and crossover rate during the run, while SaDE \cite{SaDE2008} dynamically chooses a mutation strategy based on its past success rates. Similarly, JaDE \cite{JADE2009} and CoDE \cite{CoDE2011} employ adaptive mechanisms to modify control parameters and mutation strategies, respectively.

In contrast, parameter tuning is a more static methodology, wherein the optimal configuration is established prior to the algorithm's initiation.
It aims to discover a parameter set that consistently demonstrates robust performance across various runs and problem instances.
Despite its potential for reliable outcomes, parameter tuning is known for its computational intensity, often necessitating dedicated optimization efforts or experimental designs to identify the optimal parameters, which may explain its limited exploration in the field.
Viewed as an optimization challenge, parameter tuning is also referred to as meta-optimization \cite{metaEAPhD2010}.
This perspective gave rise to \emph{MetaEA}, which optimizes the parameters of an EA using another EA.

Despite MetaEA's methodological elegance and simplicity, it confronts the significant challenge of depending on extensive function evaluations.
Fortunately, the inherent parallelism within MetaEA, across both meta-level and base-level populations, renders it particularly amenable to parallel computing environments.
However, a notable disparity exists between methodological innovations and the availability of advanced computational infrastructures, thus limiting MetaEA's potential due to the lack of advanced hardware accelerations such as GPUs.
To bridge this gap, we introduce the \emph{MetaDE} approach, which embodies the MetaEA paradigm by employing DE in a meta-level to guide the evolution of a specially tailored Parameterized Differential Evolution (PDE).

Designed with adaptability in mind, PDE can flexibly adjust its parameters and strategies, paving the way for a wide range of DE configurations.
As PDE interacts with the optimization problem at hand, the meta-level DE observes and refines PDE's settings to better align with the problem's characteristics.
Amplifying the efficiency of this nested optimization approach, MetaDE is integrated with a GPU-accelerated EC framework, thus weaving together parameter refinement and direct problem-solving into a seamless end-to-end approach to black-box optimization.
In summary, our main contributions are as follows.

\begin{itemize}
\item \textbf{Parameterized Differential Evolution:}
We have introduced Parameterized Differential Evolution (PDE), a variant of DE with augmented parameterization.
Unlike traditional DE algorithms that come with fixed mutation and crossover strategies, PDE’s architecture offers users the flexibility to adjust these parameters and strategies to fit the problem at hand.
This design not only allows for the creation of diverse DE configurations tailored for specific challenges but also ensures efficient computation.
To achieve this, all core operations of PDE, including mutation, crossover, and evaluation, have been optimized for parallel execution to harness advancement of GPU acceleration.

\item \textbf{MetaDE:}
Building on the MetaEA paradigm, we have designed the MetaDE approach.
Specifically, MetaDE employs a meta-level DE as an \texttt{evolver} to iteratively refine PDE's hyperparameters, which is guided by performance feedback from multiple PDE instances acting as \texttt{executors}.
This continuous optimization ensures PDE's configurations remain aligned with evolving problem landscapes.
Moreover, we have incorporated several specialized methods to further enhance the performance of MetaDE.

\item \textbf{GPU-accelerated Implementation:}
Breaking away from the limitations of conventional parameter tuning, we integrate MetaDE with a GPU-accelerated computing framework
 -- EvoX~\cite{evox}, which enhances MetaDE's computational prowess for facilitating swifter evaluations and algorithmic refinements.
With this specialized implementation, MetaDE provides an efficient and automated end-to-end approach to black-box optimization.
\end{itemize}

The subsequent sections are organized as follows. Section \ref{section_Preliminary} presents some preliminary knowledge for this work.
Section \ref{section_The proposed metade} elucidates the intricacies of the proposed approach, including PDE and the MetaDE.
Section \ref{section_Experimental_study} showcases the experimental results.
Finally, Section \ref{section Conclusion} wraps up the discourse and points towards avenues for future work.

\section{Preliminaries}\label{section_Preliminary}

\subsection{DE and its Parameter Adaption}\label{subsection_DE and Parameter adaption}
\subsubsection{Overview of DE}
As a typical EC algorithm, DE's essence lies in its differential mutation mechanism that drives the evolution of a population.
The operational cycle of DE unfolds iteratively, with each iteration embodying specific phases, as elaborated in Algorithm \ref{Alg_DE}:
\begin{enumerate}[label=\arabic*.]
  \item \textbf{Initialization} (Line 1):
  The algorithm initializes a set of potential solutions.
  Each of these solutions, representing vectors of decision variables, is randomly generated within the search space boundaries.
  \item \textbf{Mutation} (Lines 6-7):
  Each solution undergoes mutation to produce a mutant vector.
  This mutation process involves combinations of different individuals to form the mutant vectors.
  \item \textbf{Crossover} (Line 8):
  The crossover operation interchanges components between mutants and the original solutions to generate a trial vector.
  \item \textbf{Selection} (Lines 10-12): The trial vector competes against the original solution based on fitness, with the better solution progressing to the next generation.
\end{enumerate}

DE progresses through cycles of mutation, crossover, and selection, persisting until it encounters a termination criterion.
This could manifest as either reaching a predefined number of generations or achieving a target fitness threshold.
The algorithm's adaptability allows for the spawning of myriad DE variants by merely tweaking its mutation and crossover operations.
Specifically, DE variants follow a unified naming convention: \texttt{DE/x/y/z}, where \texttt{x} identifies the base vector used for mutation, \texttt{y} quantifies the number of difference involved, and \texttt{z} typifies the crossover method employed.
For example, the DE variant as presented in Algorithm \ref{Alg_DE} is named as \texttt{DE/rand/1/bin}.

{\linespread{1.1}
\begin{algorithm}
\small
\caption{DE}\label{Alg_DE}
\begin{algorithmic}[1]
  \Require {$D$, $NP$, $F$, $CR$, $G_{max}$}
  \State Initialize population $\mathbf{X} = \{\mathbf{x}_1, \mathbf{x}_2, \dots, \mathbf{x}_{\scalebox{0.5}{$\textit{NP}$}}\}$
  \State Evaluate the fitness of each individual in the population
  \State $g = 0$
  \While{$g \leq G_{max}$}
    \For{$i = 1$ to $NP$}
      \State Randomly select $\mathbf{x}_{r_1}$, $\mathbf{x}_{r_2}$, and $\mathbf{x}_{r_3}$ from $\mathbf{X}$,
      \Statex \qquad \quad such that $r_1 \neq r_2 \neq r_3 \neq i$
      \State Compute the mutant vector: $\mathbf{v}_i = \mathbf{x}_{r_1} + F \cdot (\mathbf{x}_{r_2} - \mathbf{x}_{r_3})$
      \State Perform crossover for each variable between $\mathbf{x}_i$ and $\mathbf{v}_i$:
      \begin{align*}
        \qquad \quad u_{i,j}=\begin{cases}
          v_{i,j},\ \text{if } \text{rand}(0, 1) \leq CR \text{ or } j = \text{randint}(1, D) \\
          x_{i,j},\ \text{otherwise}
        \end{cases}
      \end{align*}
      \State Evaluate the fitness of $\mathbf{u}_i$
      \If{$\textrm{f}(\mathbf{u}_i) \leq \textrm{f}(\mathbf{x}_i)$}
        \State Replace $\mathbf{x}_i$ with $\mathbf{u}_i$ in the population
      \EndIf
    \EndFor
    \State $g = g + 1$
  \EndWhile
  \State\Return the best fitness
\end{algorithmic}
\end{algorithm}
{\linespread{1}

\subsubsection{Parameter Modulation in DE}
DE employs a unique mutation mechanism, which adapts to the problem's natural scaling.
By adjusting the mutation step's size and orientation to the objective function landscape, DE embraces the \emph{contour matching principle} \cite{DEbook2006}, which promotes basin-to-basin transfer for enhancing the convergence of the algorithm.

At the core of DE's mutation is the scaling factor \( F \).
This factor not only determines the mutation's intensity but also governs its trajectory and ability to bypass local optima.
Commonly, \( F \) is set within the $[0.5, 1]$ interval, with a starting point often at 0.5.
While values outside the $[0.4, 1]$ range can sometimes yield good results, an \( F \) greater than 1 tends to slow convergence.
Conversely, values up to 1 generally promise swifter and more stable outcomes \cite{EPSDE2011}.
Nonetheless, to deter settling at suboptimal solutions too early, \( F \) should be adequately elevated.

Parallel to mutation, DE incorporates a uniform crossover operator, which is often labeled as discrete recombination or binomial crossover in the GA lexicon.
The crossover constant \( CR \) also plays a pivotal role, which determines the proportion of decision variables to be exchanged during the generation of offspring.
A low value for \( CR \) ensures only a small portion of decision variables are modified per iteration, thus leading to axis-aligned search steps.
As \( CR \) increases closer to 1, offspring tend to increasingly reflect their mutant parent, thereby curbing the generation of orthogonal search steps \cite{DEsurvey2011}.

For classical DE configurations, such as \texttt{DE/rand/1/bin}, rotational invariance is achieved only when \( CR \) is maxed out at 1.
Here, the crossover becomes wholly vector-driven, and offspring effectively mirror their mutants.
However, the optimal \( CR \) is intrinsically problem-dependent.
Empirical studies recommend a \( CR \) setting within the $[0, 0.2]$ range for problems characterized by separable decision variables.
Conversely, for problems with non-separable decision variables, a \( CR \) in the proximity of $[0.9, 1]$ is more effective \cite{DEsurvey2011}.

The adaptability of DE is evident in its wide spectrum of variants, each distinct in its mutation and crossover strategies with delicate parameter modulations.
In the following, we will detail seven mutation strategies and three crossover strategies, all of which are widely-recognized in state-of-the-art DE variants.
Here, the subscript notation in \( \textbf{x} \) specifies the individual selection technique.
For instance, \( \textbf{x}_r \) and \( \textbf{x}_{best} \) correspond to randomly selected and best-performing individuals respectively, whereas
\( \textbf{x}_i \) represents the currently evaluated individual.

\textbf{Mutation Strategies}:

\begin{enumerate}[label=\arabic*.]
  \item \texttt{DE/rand/1}:
        \begin{eqnarray}\label{equ_mutation rand}
        \begin{aligned}
        \mathbf{v}_i=\mathbf{x}_{r_1}+F \cdot\left(\mathbf{x}_{r_2}-\mathbf{x}_{r_3}\right).
        \end{aligned}
        \end{eqnarray}

  \item \texttt{DE/best/1}:
        \begin{eqnarray}\label{equ_mutation best}
        \begin{aligned}
        \mathbf{v}_i=\mathbf{x}_{\text {best }}+F \cdot\left(\mathbf{x}_{r_1}-\mathbf{x}_{r_2}\right).
        \end{aligned}
        \end{eqnarray}

  \item \texttt{DE/rand/2}:
        \begin{eqnarray}\label{equ_mutation rand2}
        \begin{aligned}
        \mathbf{v}_i & =\mathbf{x}_{r_1}+F \cdot\left(\mathbf{x}_{r_2}-\mathbf{x}_{r_3}\right)+F \cdot\left(\mathbf{x}_{r_4}-\mathbf{x}_{r_5}\right).
        \end{aligned}
        \end{eqnarray}

  \item \texttt{DE/best/2}:
        \begin{eqnarray}\label{equ_mutation best2}
        \begin{aligned}
        \mathbf{v}_i & =\mathbf{x}_{\text {best}}+F \cdot\left(\mathbf{x}_{r_1}-\mathbf{x}_{r_2}\right)+F \cdot\left(\mathbf{x}_{r_3}-\mathbf{x}_{r_4}\right).
        \end{aligned}
        \end{eqnarray}

  \item \texttt{DE/current-to-best/1}:
        \begin{eqnarray}\label{equ_mutation current2best}
        \begin{aligned}
        \mathbf{v}_i =\mathbf{x}_i+F \cdot\left(\mathbf{x}_{\text {best }}-\mathbf{x}_i\right)+F \cdot\left(\mathbf{x}_{r_1}-\mathbf{x}_{r_2}\right).
        \end{aligned}
        \end{eqnarray}
    The above five classical mutation strategies, introduced by Storn and Price \cite{DEbook2006}, cater to various problem landscapes.
    For instance, the `rand' variants help maintain population diversity, while strategies using two differences typically produce more diverse offspring than those relying on a single difference.

  \item \texttt{DE/current-to-pbest/1}:
        \begin{eqnarray}\label{equ_mutation current2pbest}
        \begin{aligned}
        \mathbf{v}_i =\mathbf{x}_i+F \cdot\left(\mathbf{x}_{\text {pbest}}-\mathbf{x}_i\right)+F \cdot\left(\mathbf{x}_{r_1}-\mathbf{x}_{r_2}\right).
        \end{aligned}
        \end{eqnarray}
    This strategy originates from JaDE \cite{JADE2009}. $\mathbf{x}_{\text{pbest}}$ is randomly selected from the top \emph{p}\% of individuals in the population (typically the top 10\%) to strike a balance between exploration and exploitation.

  \item \texttt{DE/current-to-rand/1}:
        \begin{eqnarray}\label{equ_mutation current2rand}
        \begin{aligned}
        &\mathbf{u}_i =\mathbf{x}_i+K_i\cdot\left(\mathbf{x}_{r_1}-\mathbf{x}_i\right)+F \cdot\left(\mathbf{x}_{r_2}-\mathbf{x}_{r_3}\right).
        \end{aligned}
        \end{eqnarray}
    Here, \(K_i\) is a random number from \(U(0,1)\).
    This strategy, originally proposed in \cite{DEintro1999}, emphasizes rotational invariance. By bypassing the crossover phase, it directly yields the trial vector \(\mathbf{u}_i\). Thus, it is ideal for addressing non-separable rotation challenges and has been a cornerstone for multiple adaptive DE variations.

  \end{enumerate}

\textbf{Crossover Strategies:}

\begin{enumerate}[label=\arabic*.]
  \item Binomial Crossover:
    \begin{eqnarray}\label{equ_cross bin}
    \begin{aligned}
    u_{i, j}= \begin{cases}v_{i, j}, & \text { if } r \leq C R \text { or } j=j_{\mathrm{rand}} \\ x_{i, j}, & \text {otherwise},\end{cases}
    \end{aligned}
    \end{eqnarray}
    where \(j_{\mathrm{rand}}\) is a random integer between 1 and \( D \). This strategy is a cornerstone in DE.

  \item Exponential Crossover:
  \begin{eqnarray}\label{equ_cross exp}
    \begin{aligned}
    \small
    u_{i, j}= \begin{cases}v_{i, j}{ } & \text { if } j=\langle n\rangle_d,\langle n+1\rangle_d,...,\langle n+L-1\rangle_d \\ x_{i, j} & \text {otherwise},\end{cases}
    \end{aligned}
    \end{eqnarray}
    where \(\langle \rangle_d\) is a modulo operation with \(D\) and \(L\) representing the crossover length, following a censored geometric distribution with a limit of \(D\) and probability of \(CR\).
    By focusing on consecutive variables, this strategy excels in handling problems with contiguous variable dependencies.

  \item Arithmetic Recombination:
  \begin{eqnarray}\label{equ_cross arith}
    \begin{aligned}
    \mathbf{u}_i=\mathbf{x}_i + K_i\cdot(\mathbf{v}_i - \mathbf{x}_i),
    \end{aligned}
    \end{eqnarray}
    where \(K_i\) is a random value from \(U(0,1)\).
    Exhibiting rotational invariance, this strategy, when combined with the \texttt{DE/rand/1} mutation, results in the \texttt{DE/current-to-rand/1} strategy \cite{DEsurvey2011}, as described by:
    \begin{eqnarray}\label{equ_rand_1_arith}
    \begin{aligned}
    \mathbf{u}_i&=\mathbf{x}_i + K_i\cdot(\mathbf{v}_i - \mathbf{x}_i)\\
    &=\mathbf{x}_i + K_i(\mathbf{x}_{r_1}+F\cdot(\mathbf{x}_{r_2}-\mathbf{x}_{r_3}) - \mathbf{x}_i)\\
    &=\mathbf{x}_i + K_i(\mathbf{x}_{r_1} - \mathbf{x}_i)+ K_i\cdot F(\mathbf{x}_{r_2}-\mathbf{x}_{r_3}),
    \end{aligned}
    \end{eqnarray}
    which is equivalent to Eq. (\ref{equ_mutation current2rand}).

\end{enumerate}

\subsubsection{Adaptive DE}
The development of parameter adaption in DE has witnessed significant advancements over time, from initial endeavors in parameter adaptation to recent sophisticated methods that merge multiple strategies.
This subsection traces the chronological advancements, emphasizing the pivotal contributions and their respective impacts on adaptive DE.

The earliest phase in DE's adaption centered on the modification of the crossover rate \( CR \) .
Pioneering algorithms such as SPDE \cite{SPDE2003} incorporated \( CR \) within the parameter set of individuals, enabling its simultaneous evolution with the decision variables of the problem to be solved.
This strategy was further refined by SDE \cite{SDE2005}, which assigned \( CR \) for each individual based on a normal distribution. Subsequent research efforts shifted focus to the scaling factor \( F \).
In this context, DETVSF \cite{DETVSF2005} dynamically adjusted \( F \), fostering exploration during the algorithm's nascent stages and pivoting to exploitation in later iterations.
Building on this, FaDE \cite{FaDE2005} employed fuzzy logic controllers to optimize mutation and crossover parameters.

The DESAP \cite{DESAP2006} algorithm marked a significant paradigm shift by introducing self-adapting populations and encapsulating control parameters within individuals.
Successive contributions like jDE \cite{jDE2006}, SaDE \cite{SaDE2008}, and JaDE \cite{JADE2009} accentuated the significance of parameter encoding, integrated innovative mutation strategies, and emphasized archiving optimization trajectories using external repositories.
Further, EPSDE \cite{EPSDE2011} and CoDE \cite{CoDE2011} enhanced the offspring generation process, amalgamating multiple strategies with randomized parameters.

The contemporary landscape of adaptive DE is characterized by complex methodologies and refined strategies.
Algorithms such as SHADE \cite{SHADE2013} and LSHADE \cite{LSHADE2014} championed the utilization of success-history mechanisms and dynamic population size modifications.
Notable developments like ADE \cite{ADE2014} introduced a biphasic parameter adaptation mechanism.
The domain further expanded with algorithms like LSHADE-RSP \cite{LSHADE-RSP2018}, IMODE \cite{IMODE2020}, and LADE \cite{LADE2023}, emphasizing mechanisms such as selective pressure, the integration of multiple DE variants, and the automation of the learning process.

Undoubtedly, the adaptive DE domain has witnessed transformative growth, with each phase of its evolution contributing to its current sophistication.
However, despite these advancements, many adaptive strategies remain empirical and hinge on manual designs, while their effectiveness is not universally guaranteed.

\subsection{Distributed DE}
The integration of distributed (i.e., multi-population) strategies also significantly enhances the efficacy of DE. 
Leading this advancement, Weber \textit{et al.} conducted extensive research on scale factor interactions and mechanisms within a distributed DE framework \cite{weber2010study, weber2011study, weber2013study}, followed by ongoing developments along the pathway \cite{DEpapersurvey2020}.
For example, some works such as EDEV \cite{EDEV2018}, MPEDE \cite{MPEDE2015} and IMPEDE \cite{IMPEDE} adopted multi-population frameworks to ensemble various DE variants/operators,
while the other works  such as DDE-AMS \cite{DDE-AMS} and DDE-ARA \cite{DDE-ARA} employed multiple populations for adaptive resource allocations.

Despite the achievements, current implementations of distributed DE often focus predominantly on algorithmic improvements, while overlooking potential enhancements from advanced hardware accelerations such as GPU computing. 
Besides, the design of these distributed strategies often features intricate and rigid configurations that lack proper flexibility.

\subsection{MetaEAs}\label{subsection_Meta-EA}

Generally, the term \emph{meta} refers to a higher-level abstraction of an underlying concept, often characterized by its \emph{recursive} nature.
In the context of EC, inception of the Meta Evolutionary Algorithms (MetaEAs) can be traced back to the pioneering works of Mercer and Sampson \cite{metaplan1978} in the late 1970s.
Under the initiative termed \emph{meta-plan}, their pioneering efforts aimed at enhancing EA performance by optimizing its parameters through another EA.
Although sharing similarities with hyperheuristics \cite{Hyperheu2013,Hyperheu2020,NeriHyperspam}, a major difference distinguishes MetaEAs: while hyperheuristics often delve into selecting and fine-tuning a set of predefined algorithms, MetaEAs concentrate on the paradigm of refining the parameters of EAs by EAs.
Notably, MetaEAs are also akin to ensemble of algorithms, such as EDEV \cite{EDEV2018} and CoDE \cite{CoDE2011}, which amalgamate diverse algorithms to ascertain the most efficacious among them.

Advancing the meta-plan concept, MetaGA \cite{MetaGA1986} emerged as a significant milestone.
Here, a genetic algorithm (GA) was deployed to fine-tune six intrinsic control parameters, namely: population size, crossover rate, mutation rate, generation gap, scaling window, and selection strategy.
The efficacy of this approach was gauged using dual metrics: online and offline performance.

The evolution of the concept continued with MetaEP \cite{metaEP1991}, which offers a meta-level evolutionary programming (EP) that could concurrently evolve optimal parameter settings.
Another pivotal contribution was the Parameter Relevance Estimation and Value Calibration (REVAC) \cite{REVAC2007}, which served as a meta estimation of distribution algorithm (MetaEDA).
Utilizing a GA at its core, REVAC iteratively discerned promising parameter value distributions within the configuration space.

Innovations in the domain persisted with the Gender-based GA (GGA) \cite{GGA2009}, inspired by natural gender differentiation, and other notable methods like MetaCMAES \cite{metaCMAES2012}.
As articulated in the PhD thesis by Pedersen \cite{metaEAPhD2010}, a profound insight into MetaEA revealed that while contemporary optimizers endowed with adaptive behavioral parameters offered advantages, they were often eclipsed by streamlined optimizers under appropriate parameter tuning.
This thesis, which embraced DE as one of its optimization tools, employed the Local Unimodal Sampling (LUS) heuristic for tuning parameters such as \( NP \), \( F \), and \( CR \).

Culminating the discourse, the work in \cite{metaEAdistributed} demonstrated the scalability of MetaEAs by harnessing it within a large-scale distributed computing environment.
With the ($\mu$, $\lambda$)$-$ES steering the meta-level tuning, base-level algorithms like GA, ES, and DE were adeptly optimized.
For DE, parameters optimized encompassed \( NP \), mutation operator, \( F \), \( CR \), and \( PF \) (parameter for the \emph{either-or} strategy), enhancing MetaEAs' prowess in addressing intricate, large-scale optimization problems.
Recently, the MetaEA paradigm has also been employed for automated design of ensemble DE \cite{EDE}.

The field of MetaEAs has shown steady progress since its inception in the 1970s.
However, despite the achievements, the landscape of MetaEAs research still confronts certain limitations.
Notably, the research, while promising, has predominantly remained confined to smaller-scale implementations.
The anticipated leap to large-scale experiments, especially those that might benefit from GPU acceleration, remains largely uncharted. This underscores an imperative need for more extensive empirical validations and the exploration of contemporary computational resources to fully realize the potential of MetaEAs.

\subsection{GPU-accelerated EC Framework}\label{subsection_Meta-optimization}
To capitalize on the advancements of modern computing infrastructures, we have seamlessly integrated our proposed MetaDE with EvoX~\cite{evox}, a distributed GPU-accelerated computing framework for scalable EC.
This integration ensures that MetaDE enables efficient execution and optimization for large-scale evaluations.

The EvoX framework provides several distinctive features.
Primarily, it is designed for optimal performance across diverse distributed systems and is tailored to manage large-scale challenges.
Its user-friendly functional programming model simplifies the EC algorithm development process, reducing inherent complexities.
The framework cohesively integrates data streams and functional elements into a comprehensive workflow, underpinned by a sophisticated hierarchical state management system.
Moreover, EvoX features a rich library of EC algorithms, proficient in addressing a wide array of tasks, from black-box optimization to advanced areas such as deep neuroevolution and evolutionary reinforcement learning.

\section{Proposed Approach}\label{section_The proposed metade}
The foundational premise of MetaDE is to utilize a core DE algorithm to evolve an ensemble of parameterized DE variants.
Within this framework, the core DE, which tunes the parameters, is termed the \texttt{evolver}.
In contrast, each parameterized DE variant, which optimizes the problem at hand, is termed the \texttt{executor}.
This section commences by augmenting the parameterization of DE in a more general manner, such that DE is made \emph{evolvable} by spawning various DE variants by modulating the parameters.
Then, this section details the integration of proposed MetaDE within a meta-framework, together with a brief introduction to the GPU-accelerated implementation.

\subsection{Augmented Parameterization of DE}\label{section_PDE}
To make DE evolvable, this subsection introduces the Parameterized Differential Evolution (PDE), an extension of the standard DE designed to augment its flexibility through the parameterization of mutation and crossover strategies.
While PDE retains the foundational principles of standard DE, its distinctiveness lies in its capability to generate a multitude of strategies by modulating the parameters.

In the standard DE framework, tunability is constrained to the adjustments of the \( F \) and \( CR \) parameters, and strategies are bound by predefined rules.
To augment this limited flexibility, PDE introduces a more granular parameterization.
Building upon DE's notation \texttt{DE/x/y/z}, PDE encompasses six parameters: \( F \) (scale factor), \( CR \) (crossover rate), \( bl \) (base vector left), \( br \) (base vector right), \( dn \) (difference number), and \( cs \) (crossover scheme).
The combined roles of \( bl \) and \( br \) determine the base vector, leading to a strategy notation for PDE expressed as \texttt{DE/bl-to-br/dn/cs}.

Each parameter's nuances and the array of strategy combinations they enable are elaborated upon in the subsequent sections.
Fig. \ref{Figure_structure} provides a visual representation of these parameters, delineated within dashed boxes.

\begin{figure*}[!h]
\centering
\includegraphics[scale=0.6]{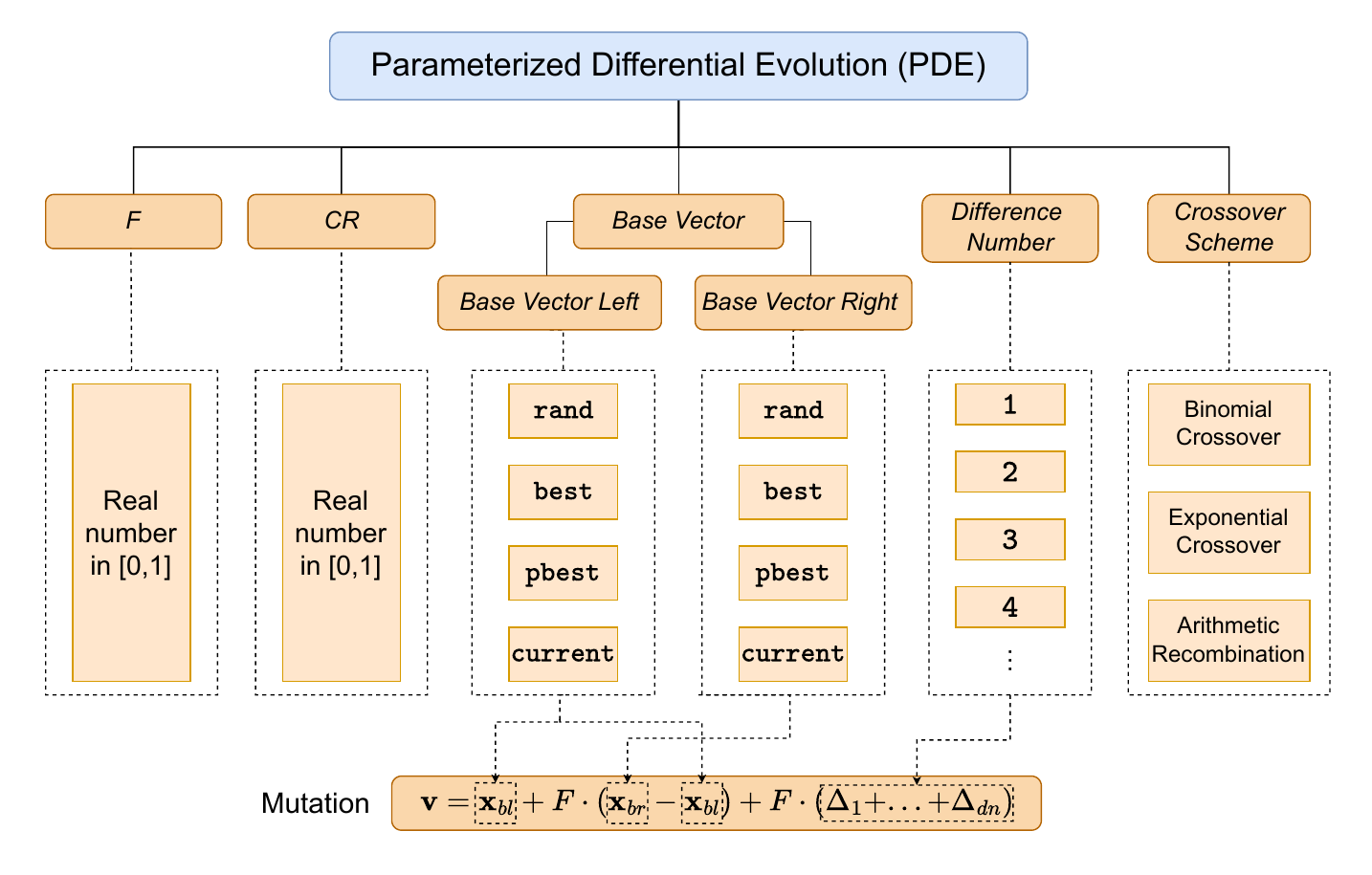}
\caption{
Parameter delineation of PDE and their respective domains.
PDE comprehensively parameterizes DE, endorsing unrestricted parameter and strategy modifications.
In this schema, \( F \) and \( CR \) are continuous parameters, whereas others are categorical.
The dashed-line boxes exhibit their specific value ranges.
The mutation function is derived from the base vector left, base vector right, and difference number parameters.
}
\label{Figure_structure}
\end{figure*}

\subsubsection{Scale Factor \( F \) and Crossover Rate \( CR \)}
The scale factor \( F \) and crossover rate \( CR \) serve as pivotal parameters in DE, both represented as real numbers.

The \( F \) parameter regulates the differential variation among population entities.
Elevated values induce exploratory search behaviors, while lower values encourage more exploitation.
Although Storn and Price originally identified [0, 2] as an effective domain for \( F \) \cite{DE1997}, contemporary DE variants deem \( F \leq 1 \) as more judicious \cite{JADE2009,CoDE2011,EPSDE2011,SHADE2013,LSHADE2014}.
Consequently, PDE constrains \( F \) within [0,1].

On the other hand, \( CR \) controls the recombination extent during crossover.
Higher \( CR \) make it more likely to try out new gene combinations from changes, while lower values keep the genes more stable and similar to the original.
As a rate (or probability), \( CR \) is confined to the [0, 1] interval.

\subsubsection{Augmented Parameterization of Mutation}
The distinctiveness of PDE lies in its ability to generate a multitude of mutation strategies by modulating parameters $bl, br$, and $dn$.
Parameters $bl$ (base vector left) and $br$ (base vector right) select one of four possible vectors:
\begin{itemize}
  \item  \texttt{rand}: A randomly chosen individual from the population.
  \item  \texttt{best}: The best individual in the population.
  \item  \texttt{pbest}: A random selection from the top \( p\% \) of individuals.
  \item  \texttt{current}: The present parent individual.
\end{itemize}
Parameter \( dn \), controls the number of differences in mutation, assumes values in the set \{1, 2, 3, 4\}.
Specifically, each difference \( \Delta \) captures the difference between two unique and randomly selected individuals from the population.
Therefore, the mutation formulation \texttt{DE/bl-to-br/dn} is:
\begin{align}\label{equ_PDE_mutation}
\mathbf{v} = \mathbf{x}_{bl} + F \cdot (\mathbf{x}_{br} - \mathbf{x}_{bl}) + F \cdot (\Delta_1 + ... +\Delta_{dn}).
\end{align}
In the implementation of PDE, \( bl \) and \( br \) are encoded as: 1: \texttt{rand}, 2: \texttt{best},  3: \texttt{pbest}, and   4: \texttt{current}.
In addition, if both \( bl \) and \( br \) assume identical values, the term $F \cdot (\mathbf{x}_{br} - \mathbf{x}_{bl})$ will disappear. The base vector takes the value of $\mathbf{x}_{bl}$ directly and the mutation strategy then becomes non-directional (e.g., \texttt{DE/rand/1}).

\subsubsection{Augmented Parameterization of Crossover}
Parameter \( cs \) defines the crossover strategies in PDE, comprising those elaborated in Section~\ref{subsection_DE and Parameter adaption}: binomial crossover, exponential crossover, and arithmetic recombination.
Specifically, \( cs \) is encoded as:   1: \texttt{bin}, 2: \texttt{exp}, and 3: \texttt{arith}, representing binomial crossover, exponential crossover, and arithmetic recombination respectively.

As a result, the augmented parameterizations for mutation and crossover, denoted as \texttt{DE/bl-to-br/dn/cs}, will give rise to a spectrum of 192 distinctive strategies in total.
This breadth allows for the encapsulation of mainstream strategies detailed in Section~\ref{subsection_DE and Parameter adaption}, as illustrated in Table \ref{tab_param encoding}.
When synergized with \( F \) and \( CR \), this culminates in a comprehensive parameter configuration landscape.

\begin{table}[htbp]
  \centering
  \caption{The encoding of typical DE variants by the proposed PDE.}
         \renewcommand{\arraystretch}{1.1}
 \renewcommand{\tabcolsep}{10pt}
    \begin{tabular}{ccccc}
    \toprule
    strategy & $bl$  & $br$  & $dn$  & $cs$ \\
    \midrule
    \texttt{DE/rand/1/bin} & 1     & 1     & 1     & 1 \\
    \midrule
    \texttt{DE/best/1/bin} & 2     & 2     & 1     & 1 \\
    \midrule
    \texttt{DE/current-to-best/1/bin} & 4     & 2     & 1     & 1 \\
    \midrule
    \texttt{DE/rand/2/bin} & 1     & 1     & 2     & 1 \\
    \midrule
    \texttt{DE/best/2/bin} & 2     & 2     & 2     & 1 \\
    \midrule
    \texttt{DE/current-to-pbest/1/bin} & 4     & 3     & 1     & 1 \\
    \midrule
    \texttt{DE/current-to-rand/1*} & 1     & 1     & 1     & 3 \\
    \bottomrule
    \end{tabular}%
  \label{tab_param encoding}%
    \vspace{0.5em}

      \footnotesize
    \textsuperscript{*} As per Eq. (\ref{equ_rand_1_arith}), \texttt{DE/current-to-rand/1} is equivalent to \texttt{DE/rand/1/arith}.\\
\end{table}%

Algorithm \ref{Alg_PDE} details PDE's procedure.
Its distinct attribute is Line 3, where the mutation function is shaped by \( bl \), \( br \), and \( dn \).
Line 4 designates the crossover function based on \( cs \).
The evolutionary phase commences at Line 6.
Notably, PDE's concurrent mutation and crossover operations (Lines 7-10) are {tensorized} to facilitate parallel offspring generation, which deviate from the conventional sequential operations in standard DE.
With such a tailored procedure, the computational efficiency can be substantially improved.

\begin{algorithm}
\small
\caption{Parameterized DE (PDE)}\label{Alg_PDE}
\begin{algorithmic}[1]
  \Require {$D$, $NP$, $G_{max}$, $F$, $CR$, $bl$ (base vector left), $br$ (base vector right), $dn$ (difference number), $cs$ (crossover scheme)}
  \State Initialize population $\mathbf{X} = \{\mathbf{x}_1, \mathbf{x}_2, \dots, \mathbf{x}_{\scalebox{0.5}{$\textit{NP}$}}\}$
  \State Evaluate the fitness of each individual in the population
  \State Generate mutation function $M(\mathbf{X})$ according to $bl$, $br$, $dn$:
  \Statex $\mathbf{v} = \mathbf{x}_{bl} + F \cdot (\mathbf{x}_{br} - \mathbf{x}_{bl}) + F \cdot (\Delta_1 + ... +\Delta_{dn})$
  \State According to $cs$, choose a crossover function $C(\mathbf{V}, \mathbf{X})$ in (\ref{equ_cross bin}-\ref{equ_cross arith})
  \State $g = 0$
  \While{$g \leq G_{max}$}
  \State Generate $NP$ mutant vectors: $\mathbf{V} = M(\mathbf{X})$
  \State Perform crossover for all mutant vectors: $\mathbf{U} = C(\mathbf{V}, \mathbf{X})$
  \State Evaluate the fitness of $\mathbf{U}$
  \State Make selection between $\mathbf{U}$ and $\mathbf{X}$
  \State $g = g + 1$
  \EndWhile
  \State\Return the best fitness
\end{algorithmic}
\end{algorithm}

\subsection{Architecture of MetaDE}
Atop the proposed PDE, this subsection further introduces the architecture of MetaDE.
The main target of MetaDE is to evolve the parameters of PDE through an external DE, empowering PDE to identify optimal parameters tailored to the target problem.
% Every computational phase of MetaDE benefits from extensive parallelization with GPU acceleration integration.

\begin{figure}[t]
\centering
\includegraphics[scale=0.9]{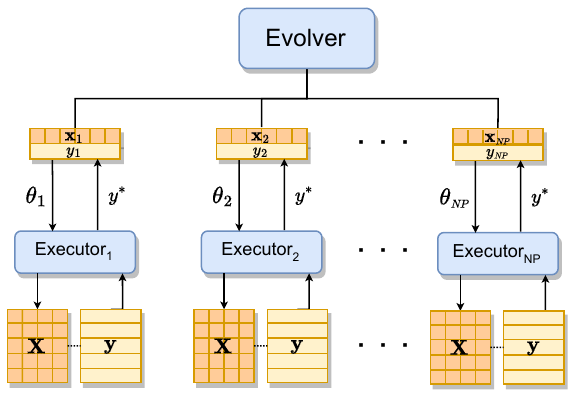}
\caption{
Architecture of MetaDE.
Within this architecture, a conventional DE algorithm operates as an \texttt{evolver}, where its individual $\mathbf{x}_i$ represents a distinct parameter configuration $\mathbf{\theta}_i$.
These configurations are relayed to PDE  to instantiate diverse DE variants as the \texttt{executors}.
Each \texttt{executor} then evolves its distinct population and returns the best fitness $y^*$ as identified, which is subsequently set as the fitness of $\mathbf{x}_i$.
}
\label{Figure_MetaDE}
\end{figure}

As illustrated in Fig. \ref{Figure_MetaDE}, MetaDE is structured with a two-tiered optimization architecture.
The upper tier, termed the \texttt{evolver}, leverages DE to evolve the parameters of PDE. In contrast, the lower tier consists of a collection of \texttt{executors} that each run the parameterized PDE instance to optimize the objective function.
Every individual in the \texttt{evolver}, represented as $\mathbf{x}_i$, is decoded into a parameter configuration $\bm{\theta}_i$ with six elements: $F$, $CR$, $bl$, $br$, $dn$, and $cs$.
For the evaluation of each individual, the configuration $\bm{\theta}_i$ is directed to its respective \texttt{executor} $\textrm{PDE}_i$ for objective function optimization.
The final fitness $y^*$ as identified by each \texttt{executor}, is subsequently set as the fitness of the corresponding $\mathbf{x}_i$ individual.

The architecture of MetaDE is streamlined for simplicity.
Building upon this architecture, MetaDE integrates two tailored components: the \emph{one-shot evaluation method} and the \emph{power-up strategy}.
These components further enhance the adaptability and efficiency of the \texttt{executors}, thereby elevating the overall performance of MetaDE.

\subsection{One-shot Evaluation Method}
Within the context of an \texttt{executor} driven by DE itself, the inherent stochastic nature can lead to variability in the optimal fitness values returned.
Historically, several evaluation techniques, such as repeated evaluation~\cite{metaCMAES2012}, F-racing~\cite{FRacing}, and intensification~\cite{intens}, have been put forth to tackle this inconsistency.
Yet, these often come at the cost of an exorbitant number of functional evaluations (FEs).
To address this issue, we introduce the one-shot evaluation method.

Specifically, the method mandates each \texttt{executor} to undertake a singular, comprehensive independent run, subsequently returning its best-found solution.
A distinguishing aspect of this method is the consistent allocation of the same initial random seed to every \texttt{executor}.
As the algorithm progresses, this uniform seed ensures that the PDE fine-tunes its parameters in a consistent manner, thereby identifying optimal parameters tailored to the given seed environment.
Essentially, this strategy embeds the seed as an integral facet of the problem domain.

\subsection{Power-up Strategy}
During the independent runs of an \texttt{executor}, the allocation of FEs plays a pivotal role in determining both the quality of solutions and computational efficiency. Allocating an excessive number of FEs indiscriminately can lead to undue computational resource consumption without necessarily improving solution quality.
To address this issue, we propose the power-up strategy.

The essence of this strategy is dynamic FE allocation: while earlier iterations receive a moderate number of FEs to ensure resource efficiency, a more generous allocation (fivefold) is reserved for the terminal iteration within the evolutionary process.
This strategy ensures that the \texttt{executor} has the resources for a thorough and comprehensive evaluation during its most crucial phase -- the final generation of the \texttt{evolver}.

\subsection{Implementation}
As outlined in Algorithm~\ref{Alg_MetaDE}, MetaDE draws its simple algorithmic workflow from conventional DE. MetaDE adopts \texttt{DE/rand/1/bin} as the \texttt{evolver}.
The initialization phase (Line 1) spawns the MetaDE population within the parameter boundaries \( [\mathbf{lb}, \mathbf{ub}] \).
During the evaluation phase (Lines 6-11), each individual is decoded into a parameter blueprint and directed to an independent PDE instance (\texttt{executor}) for problem resolution.
Running for a predetermined iteration count \( G' \), each \texttt{executor} subsequently reports the best fitness.
Notably, Line 10 encapsulates the essence of the power-up strategy: for MetaDE's concluding iteration (\( g == G_{max} \)), the evaluation quota is amplified to \( 5 \times G' \) for the \texttt{executors}.

\begin{algorithm}
\small
\caption{MetaDE}
\label{Alg_MetaDE}
\begin{algorithmic}[1]
  \Require {$D$, $NP$, $G_{max}$, $\mathbf{lb}$ (lower boundaries of PDE's parameters), $\mathbf{ub}$ (upper boundaries of PDE's parameters), $NP'$ (population size of PDE), $G'$ (max generations of PDE)}
  \State Initialize population $\mathbf{X} = \{\mathbf{x}_1, \mathbf{x}_2, \dots, \mathbf{x}_{\scalebox{0.5}{$\textit{NP}$}}\}$ between $[\mathbf{lb}, \mathbf{ub}]$
  \State Initialize the fitness of $\mathbf{X}$: $\mathbf{y}=\mathbf{inf}$
  \State $g = 0$
  \While{$g \leq G_{max}$}
  \State Generate trial vectors $\mathbf{U}$ by mutation and crossover
  \Statex \quad \ /* The mutation and crossover scheme used is rand/1/bin\ */
  \State Decode each trial vector $\mathbf{u}$ into parameters:
  \Statex \quad \ $F=\mathbf{u}[1], CR=\mathbf{u}[2], bl=\textrm{floor}(\mathbf{u}[3]), br=\textrm{floor}(\mathbf{u}[4])$,
  \Statex \quad \ $dn=\textrm{floor}(\mathbf{u}[5]), cs=\textrm{floor}(\mathbf{u}[6])$
  %\Statex /*Evaluate each $\mathbf{u}$ by running PDE for certain generations*/
  \If {$g < G_{max}$}
    \State $\mathbf{y} = \textrm{PDE}(D, NP', G', F, CR, bl, br, dn, cs)$
  \Else
    \State $\mathbf{y} = \textrm{PDE}(D, NP', 5 * G', F, CR, bl, br, dn, cs)$
  \EndIf
  \State Make selection between $\mathbf{U}$ and $\mathbf{X}$
  \State $g = g + 1$
  \EndWhile
  \State\Return the best individual and fitness
\end{algorithmic}
\end{algorithm}

Evidently, the algorithmic design of MetaDE provides an automated end-to-end approach to black-box optimization.
However, the computational demands of MetaDE, particularly in terms of FEs, cannot be understated.
In historical computational contexts, such intensive demands might have posed significant impediments.
Fortunately, contemporary advancements in computational infrastructures, coupled with the ubiquity of high-performance computational apparatuses such as GPUs, have substantially alleviated such a challenge.

Hence, we leverage the GPU-accelerated framework of EvoX~\cite{evox} for the implementation of MetaDE.
Thanks to the inherently parallel nature of MetaDE, computational tasks can be judiciously delegated to GPUs to engender optimized runtime performance.
Specifically, the parallelism in MetaDE manifests in three distinct facets:
\begin{itemize}
  \item \textbf{Parallel Initialization and Execution:}
 The multiple \texttt{executors} are instantiated and operated concurrently, each tailored by a unique parameter configuration derived from the MetaDE ensemble.
  This simultaneous operation enables comprehensive exploration across varied parameter landscapes.

  \item \textbf{Parallel Offspring Generation:}
  Both the \texttt{evolver} and \texttt{executors} adhere to parallel strategies for offspring inception.
  By synchronizing and coordinating mutations and crossover operations in their respective populations, MetaDE is able to rapidly produce offspring, thereby accelerating the evolutionary process.

  \item \textbf{Parallel Fitness Evaluations:}
  Each \texttt{executor} conducts fitness evaluations concurrently across its member individuals.
  Given the substantial number of the individuals within the populations of the \texttt{executors}, this parallel strategy significantly enhances the overall efficiency of MetaDE.
\end{itemize}

\lstset{
  language=Python,
  aboveskip=3mm,
  belowskip=3mm,
  showstringspaces=false,
  columns=flexible,
  basicstyle={\footnotesize\ttfamily},
  numbers=left,
  numberstyle=\tiny\color{gray},
  xleftmargin=2em,
  keywordstyle=\bfseries\color{dkgreen},
  commentstyle=\color{gray}\itshape,
  stringstyle=\color{mauve},
  breaklines=true,
  breakatwhitespace=true,
  tabsize=3,
  emph={[1]evox,BatchExecutor,MetaProblem},          % Emphasize numpy
  emphstyle={[1]\bfseries\color{dkblue}},  % Set the style for emphasized words
  emph={[2]__init__, min, evaluate, reproduce, init},
  emphstyle={[2]\color{dkblue}},
  emph={[3]self,super},          % Emphasize numpy
  emphstyle={[3]\color{dkgreen}},
  morekeywords={from,import},  % Add more keywords
  captionpos=b,             % Caption position
      frame=lines,
  framesep=2mm,
}
\
\begin{lstlisting}[caption={Demonstrative implementation of MetaDE leveraging the computational workflow of EvoX. The implementation is distinctly divided into four pivotal components: Workflow Initialization, Meta Problem Transformation, Computing Workflow Creation, and Execution.}, label={lst:python_example}, float=!t]
from evox import algorithms, problems, ...

### Initialization ###
evolver = algorithm.DE()  # specify evolver
executor = algorithm.PDE()  # specify executor
problem = ...  # specify optimization problem

### Meta Problem Transformation ###
class MetaProblem(Problem):
    def __init__(self, batch_executor, ... ):
        # vectorize fitness evaluations
        self.batch_evaluate = vectorize(vectorize(problem.evaluate))

    def evaluate(self, state, ...):
        ...
        # run executors
        while ...:
            ...
            batch_fits, ... = self.batch_evaluate(...)
        # return fitness
        return min(min(batch_fits))

### Computing Workflow Creation ###
batch_executor = create_batch_executor(...)
meta_problem = MetaProblem(batch_executor, ...)
workflow = workflow.UniWorkflow(
            algorithm = evolver,
            pop_transform = decoder,
            problem = meta_problem,
        )

### Execution ###
while ...:
    state = workflow.step(state)
\end{lstlisting}

MetaDE adheres rigorously to the functional programming paradigm, capitalizing on automatic vectorization for parallel execution. Core algorithmic components, including crossover, mutation, and evaluation, are constructed using pure functions. Subsequently, the entire program is mapped to a GPU-based computation graph, ushering in accelerated processing. EvoX's adept state management ensures a seamless transfer of the algorithm's prevailing state, encompassing aspects like population, fitness, hyperparameters, and auxiliary data.

Listing \ref{lst:python_example} elucidates a representative implementation of MetaDE underpinned by the EvoX framework, which is meticulously segmented into four salient phases:

\begin{itemize}
    \item \textbf{Initialization}: Herein, primary entities like the \texttt{evolver} (employing the traditional DE) and the \texttt{executor} (utilizing the proposed PDE) are instantiated. Concurrently, the target optimization problem is defined.

    \item \textbf{Meta Problem Transformation}: Within this phase, the original optimization problem is transformed to align with the meta framework. This metamorphosis is realized via the \texttt{\textbf{MetaProblem}} class, where the evaluation function undergoes vectorization, priming it for efficient batch assessments and facilitating concurrent evaluations of manifold configurations.

    \item \textbf{Computing Workflow Creation}: Post transformation, the workflow is architected to seamlessly amalgamate the initialized components. The \texttt{batch\_executor} is crafted for batched operations of DE variants, and the \texttt{\textbf{MetaProblem}} is instantiated therewith. The holistic workflow, embodied by the \texttt{UniWorkflow} class, is then constructed, weaving together the \texttt{evolver}, the transformed problem, and a (\texttt{decoder}) which transforms the \texttt{evolver}'s population into specific hyperparameters for instantiating  the DE variant of each \texttt{executor}.

    \item \textbf{Execution}: Having established the groundwork, MetaDE's execution phase is triggered, autonomously driving the computing workflow across distributed GPUs.
    This workflow is traversed iteratively, culminating once a predefined termination criterion is met.
\end{itemize}

\section{Experimental Study}\label{section_Experimental_study}
In this section, we conduct detailed experimental assessments of MetaDE's capabilities.
First, we comprehensively benchmark MetaDE against several representative DE variants and CEC2022 top algorithms to gauge its relative performance on the CEC2022 benchmark suite \cite{CEC2022SO}.
Then, we investigate the optimal DE variants obtained by MetaDE in the benchmark experiment.
Finally, we apply MetaDE to robot control tasks.
All experiments were conducted on a system equipped with an Intel Core i9-10900X CPU and  an NVIDIA RTX 3090 GPU.
For GPU acceleration, all the algorithms and test functions were implemented within EvoX \cite{evox}.

\subsection{Benchmarks against Representative DE Variants}\label{section_Comparison with Classic DE Variants}

\subsubsection{Experimental Setup}
The CEC2022 benchmark suite for single-objective black-box optimization was utilized for this study.
This suite includes basic ($F_1-F_5$), hybrid ($F_6-F_8$), and composition functions ($F_9-F_{12}$), catering to various optimization characteristics such as unimodality/multimodality and separability/non-separability.

For benchmark comparisons, we selected seven representative DE variants: DE (\texttt{rand/1/bin}) \cite{DE1997}, SaDE \cite{SaDE2008}, JaDE \cite{JADE2009}, CoDE \cite{CoDE2011}, SHADE \cite{SHADE2013}, LSHADE-RSP \cite{LSHADE-RSP2018}, and EDEV \cite{EDEV2018}, which encapsulate a spectrum of mutation, crossover, and adaptation strategies. All algorithms were reimplemented using EvoX, with each capable of running in parallel, including the concurrent evaluation and reproduction.

Their respective descriptions are as follows:
\begin{itemize}
  \item \texttt{DE/rand/1/bin} is a foundational DE variant, which leverages a random mutation strategy coupled with binomial crossover.
  \item SaDE maintains an archive for tracking successful strategies and \( CR \) values and exhibits adaptability in strategy selection and parameter adjustments throughout the optimization process.
  \item JaDE relies on the \texttt{current-to-pbest} mutation strategy and dynamically adjusts its \( F \) and \( CR \) parameters during the optimization trajectory.
  \item CoDE infuses generational diversity by composing three disparate strategies, each complemented with randomized parameters, for offspring generation.
  \item SHADE employs the current-to-pbest mutation strategy and integrates a success-history mechanism to fine-tune its \( F \) and \( CR \)  parameters adaptively.
  \item LSHADE-RSP, as one of the most competitive DE variants, employs delicate strategies such as linear population size reduction and ranking-based mutation.
  \item EDEV adopts a distributed framework that ensembles three classic DE variants: JaDE, CoDE, and EPSDE.
\end{itemize}

The population size for all comparative algorithms was uniformly set to 100, except for experiments involving large populations. The other parameters for these algorithms were adopted as per their default settings described in their respective publications.

In our MetaDE configuration, on one hand, the \texttt{evolver} had a population size of 100 and adopted the vanilla \texttt{rand/1/bin} strategy with $F=0.5$ and $CR=0.9$;
On the other hand, each \texttt{executor} maintained a population of 100, iterating 1000 times for all the problems.
For simplicity, any result exceeding the precision of $10^{-8}$ was truncated to 0.
All statistical results were obtained via 31 independent runs\footnote{Full results, including the statistical results applying Wilcoxon rank-sum tests with a a significance level of 0.05, can be found in the Supplementary Document.}.

\begin{figure}[!t]
\centering
\includegraphics[scale=0.3]{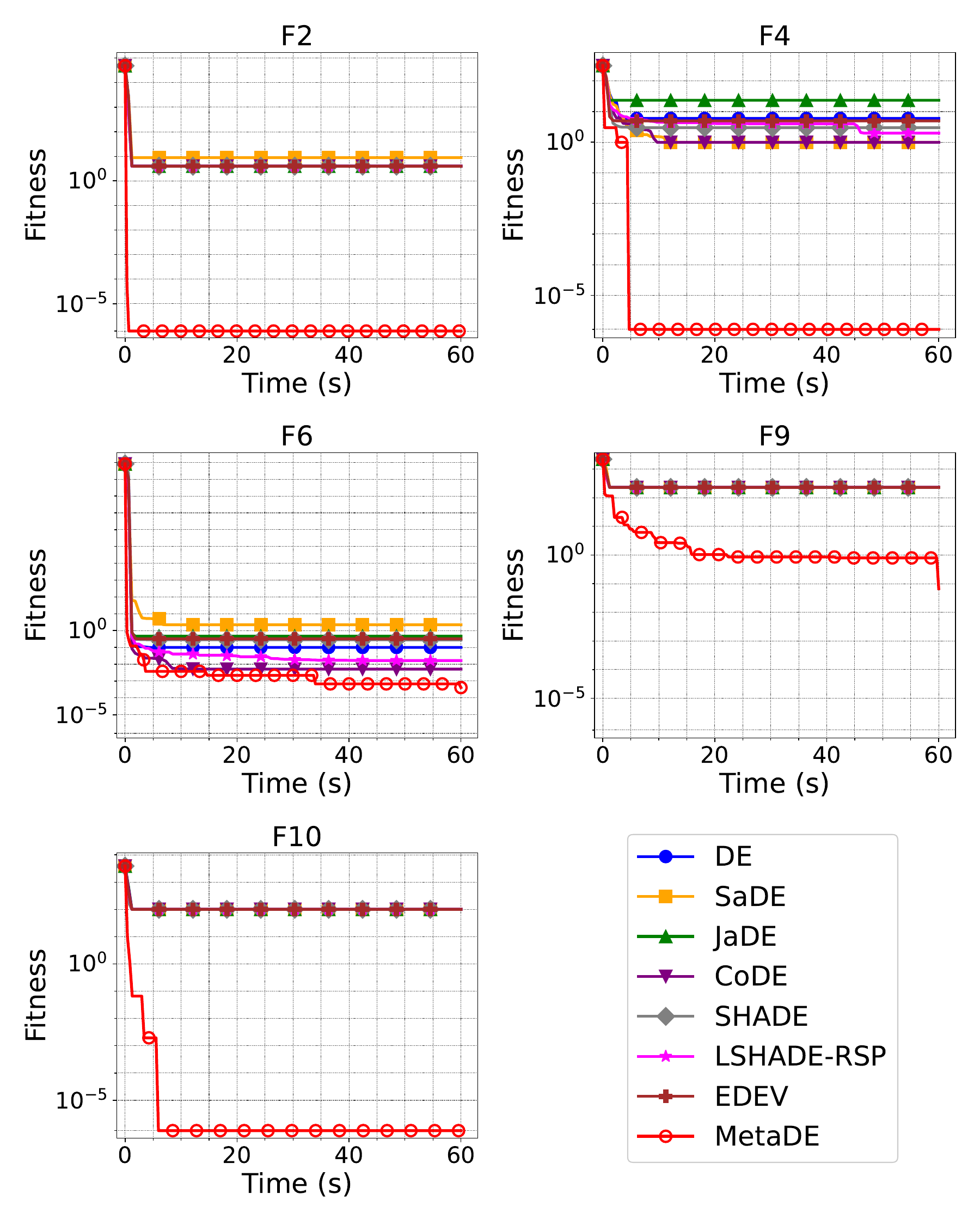}
\caption{Convergence curves on 10D problems in CEC2022 benchmark suite. The peer DE variants are set with population size of 100.}
\label{Figure_convergence_10D}
\end{figure}

\begin{figure}[!t]
\centering
\includegraphics[scale=0.3]{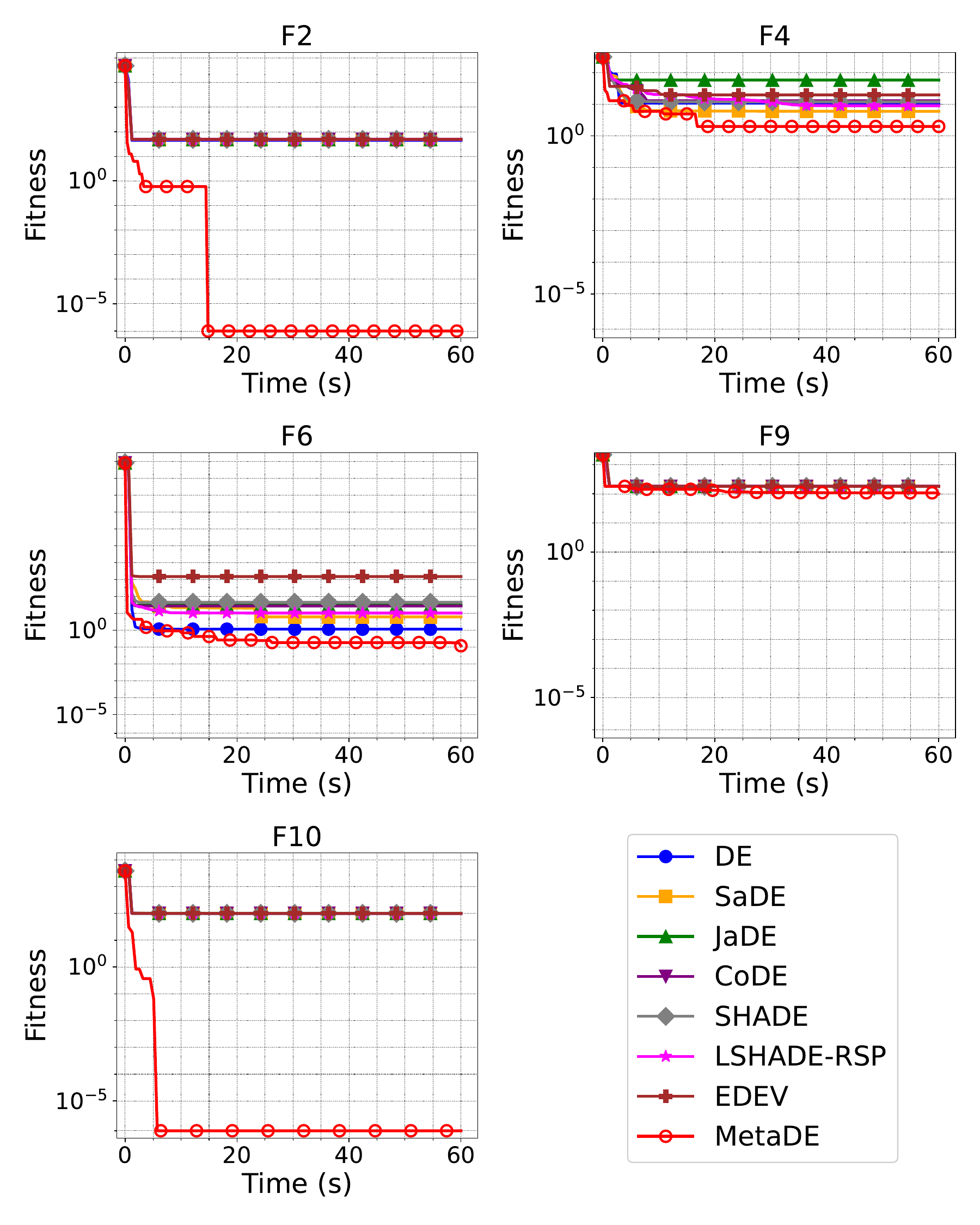}
\caption{Convergence curves on 20D problems in CEC2022 benchmark suite. The peer DE variants are set with population size of 100.}
\label{Figure_convergence_20D}
\end{figure}

\subsubsection{Performance under Equal Wall-clock Time}\label{sec:expereiment_time}
In this part, we set equal  wall-clock time (\SI{60}{\second}) as the termination condition for running each test. 
{
This approach aligns with the practical constraints of modern GPU computing, where execution time serves as a more meaningful and comparable measure of performance across algorithms. Since all algorithms in our experiments are implemented with GPU parallelism, this setup ensures fairness by standardizing the computational resources and focusing on efficiency within the same time budget.
}

As shown in Figs. \ref{Figure_convergence_10D} and \ref{Figure_convergence_20D}, we selected five challenging problems, specifically $F_2$, $F_4$, $F_6$, $F_9$, and $F_{10}$, to demonstrate the convergence profiles.
Notably, MetaDE's convergence curve is observably more favorable, consistently registering lower errors than its counterparts across the majority of the problems.
Particularly, on $F_2$, $F_4$, $F_9$, and $F_{10}$, MetaDE exhibits resilience against local optima entrapment and subsequent convergence stagnation.
This is attributed to MetaDE's capability to identify optimal algorithm settings tailored for diverse problems, rather than merely tweaking parameters based on isolated segments of the optimization trajectory, as is the case with some DE variants.
An intriguing characteristic of MetaDE's convergence, evident in functions like $F_9$ (refer to Fig. \ref{Figure_convergence_10D}), is its pronounced performance surge in the optimization's terminal phase.
This enhancement can be linked to MetaDE's power-up strategy of allocating bonus computational resources in its final phase (as per Line 10 of Algorithm \ref{Alg_MetaDE}).

\begin{figure}[!h]
\centering
\includegraphics[scale=0.3]{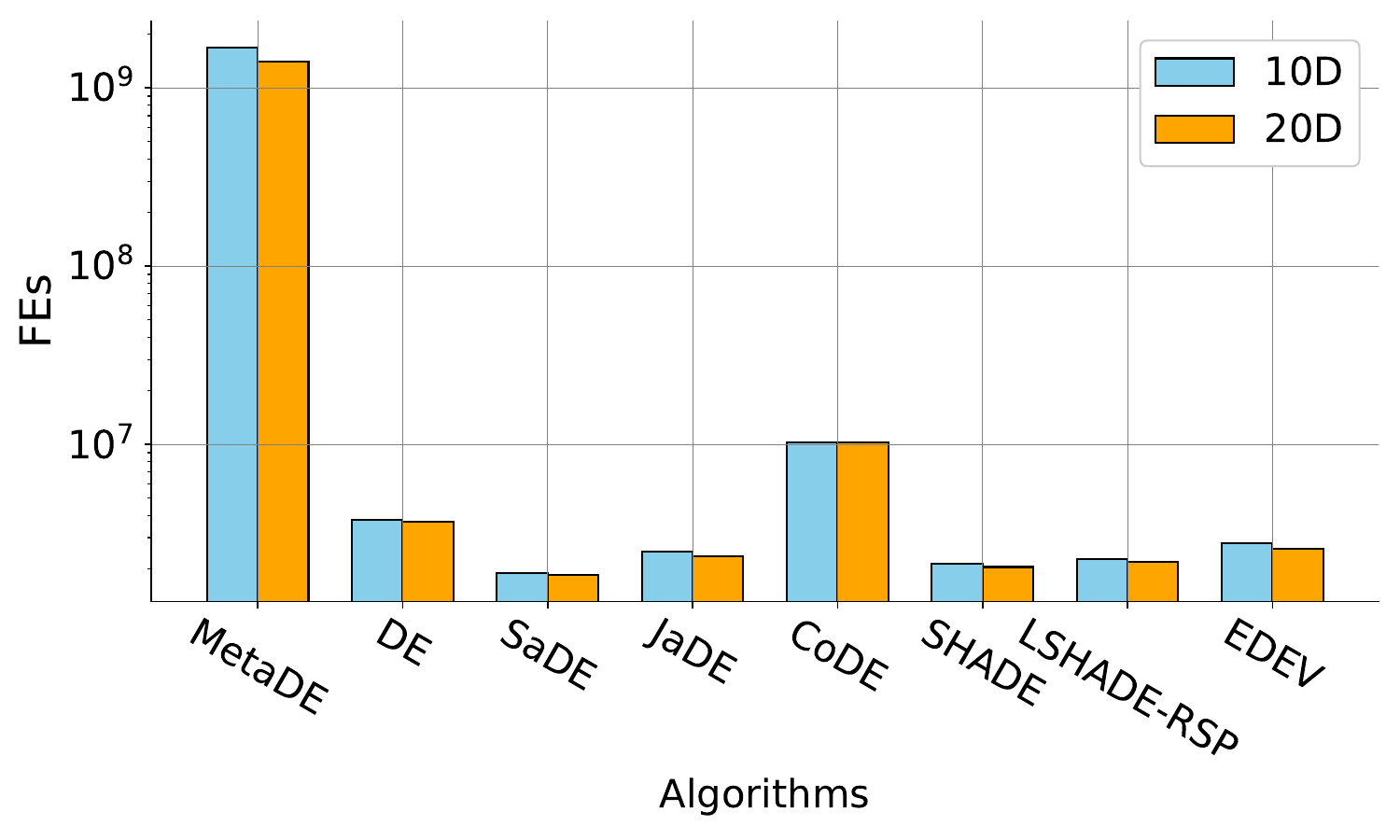}
\caption{The number of FEs achieved by each algorithm within \SI{60}{\second}. The results are averaged on all 10D and 20D problems in the CEC2022 benchmark suite.}
\label{fig:maxFEs}
\end{figure}

{
Furthermore, to assess the concurrency of the algorithms, the number of FEs achieved by each algorithm within 60 seconds is shown in Table~\ref{tab:FEs} and Fig~\ref{fig:maxFEs}.
}
The results indicate that MetaDE achieves approximately $10^9$ FEs within 60 seconds, while the other algorithms manage to attain only around $10^7$ FEs in the same time frame.
The results demonstrate the high concurrency of MetaDE, which is particularly favorable in GPU computing.

\begin{table}[htbp]
  \centering
  
  \caption{{The number of FEs achieved by each algorithm within \SI{60}{\second}.}}
 
\scriptsize                   %设置字体大小
\renewcommand{\arraystretch}{1}
\renewcommand{\tabcolsep}{2.5pt}   %pt越大字越小
\resizebox{\linewidth}{!}{
% Table generated by Excel2LaTeX from sheet 'Experiment1 60S'
\begin{tabular}{cccccccccc}
\toprule
Dim   & Func  & MetaDE & DE    & SaDE  & JaDE  & CoDE  & SHADE &LSHADE-RSP&EDEV\\
\midrule
\multirow{12}[2]{*}{10D} & $F_{1}$ & \textbf{1.85E+09} & 4.28E+06 & 1.96E+06 & 2.55E+06 & 1.09E+07 & 2.24E+06&2.57E+06&2.79E+06 \\
      & $F_{2}$ & \textbf{1.84E+09} & 4.19E+06 & 1.89E+06 & 2.42E+06 & 1.11E+07 & 2.18E+06&2.58E+06&2.82E+06 \\
      & $F_{3}$ & \textbf{1.50E+09} & 4.00E+06 & 1.89E+06 & 2.50E+06 & 1.11E+07 & 2.10E+06& 2.46E+06&2.68E+06\\
      & $F_{4}$ & \textbf{1.84E+09} & 4.11E+06 & 2.02E+06 & 2.61E+06 & 1.11E+07 & 2.15E+06&2.60E+06&2.88E+06 \\
      & $F_{5}$ & \textbf{1.83E+09} & 4.13E+06 & 2.00E+06 & 2.60E+06 & 1.15E+07 & 2.17E+06& 2.53E+06&2.96E+06\\
      & $F_{6}$ & \textbf{1.84E+09} & 4.31E+06 & 1.96E+06 & 2.65E+06 & 1.07E+07 & 2.14E+06&2.55E+06&2.95E+06\\
      & $F_{7}$ & \textbf{1.74E+09} & 3.35E+06 & 1.90E+06 & 2.55E+06 & 9.96E+06 & 2.14E+06& 2.41E+06&2.87E+06\\
      & $F_{8}$ & \textbf{1.72E+09} & 3.34E+06 & 1.83E+06 & 2.53E+06 & 9.60E+06 & 2.17E+06&2.34E+06&2.74E+06 \\
      & $F_{9}$ & \textbf{1.78E+09} & 3.35E+06 & 1.84E+06 & 2.52E+06 & 9.69E+06 & 2.18E+06&2.44E+06 &2.82E+06\\
      & $F_{10}$ & \textbf{1.44E+09} & 3.32E+06 & 1.83E+06 & 2.46E+06 & 9.00E+06 & 2.12E+06& 2.30E+06&2.70E+06\\
      & $F_{11}$ & \textbf{1.46E+09} & 3.55E+06 & 1.88E+06 & 2.34E+06 & 9.66E+06 & 2.13E+06&2.34E+06& 2.63E+06\\
      & $F_{12}$ & \textbf{1.43E+09} & 3.46E+06 & 1.83E+06 & 2.41E+06 & 9.51E+06 & 2.09E+06&2.32E+06 &2.67E+06\\
\midrule
\midrule
\multirow{12}[2]{*}{20D} & $F_{1}$ & \textbf{1.66E+09} & 4.32E+06 & 1.92E+06 & 2.46E+06 & 1.13E+07 & 2.21E+06&2.68E+06&2.80E+06 \\
      & $F_{2}$ & \textbf{1.66E+09} & 3.91E+06 & 1.88E+06 & 2.37E+06 & 1.17E+07 & 2.09E+06& 2.66E+06&2.74E+06\\
      & $F_{3}$ & \textbf{1.18E+09} & 3.76E+06 & 1.77E+06 & 2.33E+06 & 9.75E+06 & 1.95E+06&2.62E+06& 2.64E+06\\
      & $F_{4}$ & \textbf{1.65E+09} & 3.50E+06 & 1.87E+06 & 2.28E+06 & 1.07E+07 & 2.00E+06&2.63E+06 &2.80E+06\\
      & $F_{5}$ & \textbf{1.64E+09} & 3.57E+06 & 1.86E+06 & 2.32E+06 & 1.07E+07 & 2.05E+06& 2.55E+06&2.74E+06\\
      & $F_{6}$ & \textbf{1.64E+09} & 3.89E+06 & 1.90E+06 & 2.34E+06 & 1.16E+07 & 2.09E+06& 2.62E+06&2.80E+06\\
      & $F_{7}$ & \textbf{1.45E+09} & 4.15E+06 & 1.84E+06 & 2.36E+06 & 1.01E+07 & 1.99E+06&2.32E+06& 2.80E+06\\
      & $F_{8}$ & \textbf{1.44E+09} & 3.42E+06 & 1.82E+06 & 2.31E+06 & 9.51E+06 & 2.12E+06&2.18E+06&2.30E+06 \\
      & $F_{9}$ & \textbf{1.57E+09} & 3.30E+06 & 1.77E+06 & 2.33E+06 & 9.48E+06 & 2.03E+06&2.59E+06&2.75E+06 \\
      & $F_{10}$ & \textbf{9.80E+08} & 3.67E+06 & 1.82E+06 & 2.43E+06 & 1.01E+07 & 2.08E+06& 2.09E+06&2.35E+06\\
      & $F_{11}$ & \textbf{1.00E+09} & 3.51E+06 & 1.95E+06 & 2.41E+06 & 9.81E+06 & 2.07E+06&2.17E+06& 2.37E+06\\
      & $F_{12}$ & \textbf{9.90E+08} & 3.44E+06 & 1.85E+06 & 2.37E+06 & 9.51E+06 & 2.07E+06& 2.17E+06&2.39E+06\\
\bottomrule
\end{tabular}%
}
  \label{tab:FEs}%
\end{table}%

\subsubsection{Performance under Equal FEs}\label{sec:expereiment_FEs}

\begin{table*}[htbp]
  %\centering
\caption{Comparisons between MetaDE and other DE variants under equal FEs. 
The mean and standard deviation (in parentheses) of the results over multiple runs are displayed in pairs. 
Results with the best mean values are highlighted.}
  \resizebox{\linewidth}{!}{
  \renewcommand{\arraystretch}{1.2}
 \renewcommand{\tabcolsep}{2pt}
% Table generated by Excel2LaTeX from sheet 'Exp3 same FEs'
\begin{tabular}{cccccccccc}
\toprule
\multicolumn{2}{c}{Func} & MetaDE & DE    & SaDE  & JaDE  & CoDE  & SHADE & LSHADE-RSP & EVDE \\
\midrule
\multirow{3}[2]{*}{10D} & $F_{2}$ & \textbf{0.00E+00 (0.00E+00)} & 4.52E+00 (2.36E+00)$-$ & 6.85E+00 (3.52E+00)$-$ & 6.31E+00 (3.05E+00)$-$ & 5.78E+00 (2.37E+00)$-$ & 4.33E+00 (3.81E+00)$-$ & 2.35E+00 (3.44E+00)$-$ & 5.86E+00 (2.99E+00)$-$ \\
      & $F_{6}$ & \textbf{5.50E-04 (3.96E-04)} & 1.13E-01 (7.83E-02)$-$ & 3.54E+01 (1.11E+02)$-$ & 2.02E+00 (3.34E+00)$-$ & 6.96E-03 (5.99E-03)$-$ & 9.27E-01 (1.31E+00)$-$ & 3.10E-02 (4.83E-02)$-$ & 1.46E+00 (2.76E+00)$-$ \\
      & $F_{10}$ & \textbf{0.00E+00 (0.00E+00)} & 1.00E+02 (4.40E-02)$-$ & 1.00E+02 (6.10E-02)$-$ & 1.21E+02 (4.35E+01)$-$ & 1.00E+02 (6.88E-02)$-$ & 1.29E+02 (4.67E+01)$-$ & 1.09E+02 (2.87E+01)$-$ & 1.10E+02 (3.05E+01)$-$ \\
\midrule
\midrule
\multirow{3}[2]{*}{20D} & $F_{2}$ & \textbf{1.26E-02 (3.74E-02)} & 4.72E+01 (2.09E+00)$-$ & 4.76E+01 (2.02E+00)$-$ & 1.34E+01 (2.19E+01)$-$ & 4.91E+01 (1.70E-06)$-$ & 4.91E+01 (3.40E-06)$-$ & 4.84E+01 (1.62E+00)$-$ & 4.47E+01 (1.41E+01)$-$ \\
      & $F_{6}$ & \textbf{1.16E-01 (2.79E-02)} & 7.28E-01 (5.22E-01)$-$ & 3.20E+01 (1.61E+01)$-$ & 4.90E+01 (3.31E+01)$-$ & 2.26E+01 (1.80E+01)$-$ & 5.67E+01 (3.90E+01)$-$ & 1.15E+01 (8.38E+00)$-$ & 2.92E+03 (5.82E+03)$-$ \\
      & $F_{10}$ & \textbf{0.00E+00 (0.00E+00)} & 1.07E+02 (2.07E+01)$-$ & 1.00E+02 (2.72E-02)$-$ & 1.01E+02 (3.64E-02)$-$ & 1.00E+02 (3.71E-02)$-$ & 1.43E+02 (5.64E+01)$-$ & 1.21E+02 (4.52E+01)$-$ & 1.14E+02 (4.65E+01)$-$ \\
\midrule
\multicolumn{2}{c}{$+$ / $\approx$ / $-$} & --    & 0/0/6 & 0/0/6 & 0/0/6 & 0/0/6 & 0/0/6 & 0/0/6 & 0/0/6 \\
\bottomrule
\end{tabular}%

}
\label{tab:sameFEs}%

\footnotesize
\textsuperscript{*} The Wilcoxon rank-sum tests (with a significance level of 0.05) were conducted between MetaDE and each algorithm individually.
The final row displays the number of problems where the corresponding algorithm performs statistically better ($+$),  similar ($\thickapprox$), or worse ($-$) compared to MetaDE.
\end{table*}%

In the preceding part, the performance benchmarking of MetaDE with other algorithms was anchored to equal wall-clock durations.
However, to ensure a comprehensive assessment, it is imperative to evaluate their performances under equivalent FEs.
In this part, we run each algorithm using the FEs achieved by MetaDE in \SI{60}{\second} (i.e., $1.84\times10^9/1.66\times10^9$, $1.84\times10^9/1.64\times10^9$, and $1.44\times10^9/9.8\times10^8$) on $F_2$, $F_6$, and $F_{10}$ for 10D/20D cases.
These selected functions collectively epitomize the basic, hybrid, and composition challenges within the CEC2022 benchmark suite.

As summarized in Table \ref{tab:sameFEs}, MetaDE consistently demonstrates the best performance, even when other algorithms are endowed with comparable FEs.
The reason can be traced to the inherent stagnation tendencies of other algorithms: after a certain point, additional FEs may not contribute to performance improvements.
This behavioral pattern is also lucidly captured in the convergence curves as presented in Figs. \ref{Figure_convergence_10D} and \ref{Figure_convergence_20D}.

Another noteworthy observation is the extended computation time required for a singular run of the comparison algorithms under these enhanced FEs, often extending to several hours or even transcending a day (e.g., running a single run of DE can take up to seven hours).
This elongated computational span can largely be attributed to their low concurrency, which struggles to benefit the parallelism of GPU computing.

\subsubsection{Performance with Large Populations}\label{sec:expereiment_large_pop}
Since a large population size could potentially increase the concurrency of fitness evaluations, for rigorousness, we further investigate the performance of the algorithms with large populations.

Specifically, MetaDE adopted the same population size setting as in previous experiments (i.e., 100 for both \texttt{evolver} and \texttt{executor}), while the population size of the other DE variants was increased to 1,000. 
This adjustment significantly enhances the concurrency of the other DE variants when utilizing GPU accelerations, thereby preventing insufficient convergence.

As evidenced in Figs. \ref{Figure_convergence_10D_NP10k}-\ref{Figure_convergence_20D_NP10k}, MetaDE still outperforms the other DE variants across all problems.
However, the performances of the other DE variants did not show significant improvements, which can be attributed to two factors.
First, since the conventional DE variants were not tailored for large populations, simply enlarging the populations may not help.
Second, since the sorting and archiving operations in some DE variants (e.g., SaDE) suffer from high computational complexities related to the population size, enlarging the populations brings additional computation overheads, thus limiting their performances under fixed wall-clock time.

By contrast, the large population in MetaDE is delicately organized in a \emph{hierarchical} manner: the \texttt{executor} maintains a population of moderate size, with each individual initializing an \texttt{executor} with a normal population.
This strategy not only capitalizes on the small-population advantage of conventional DE, but also benefits the concurrency brought by large populations.

\begin{figure}[!t]
\centering
\includegraphics[scale=0.3]{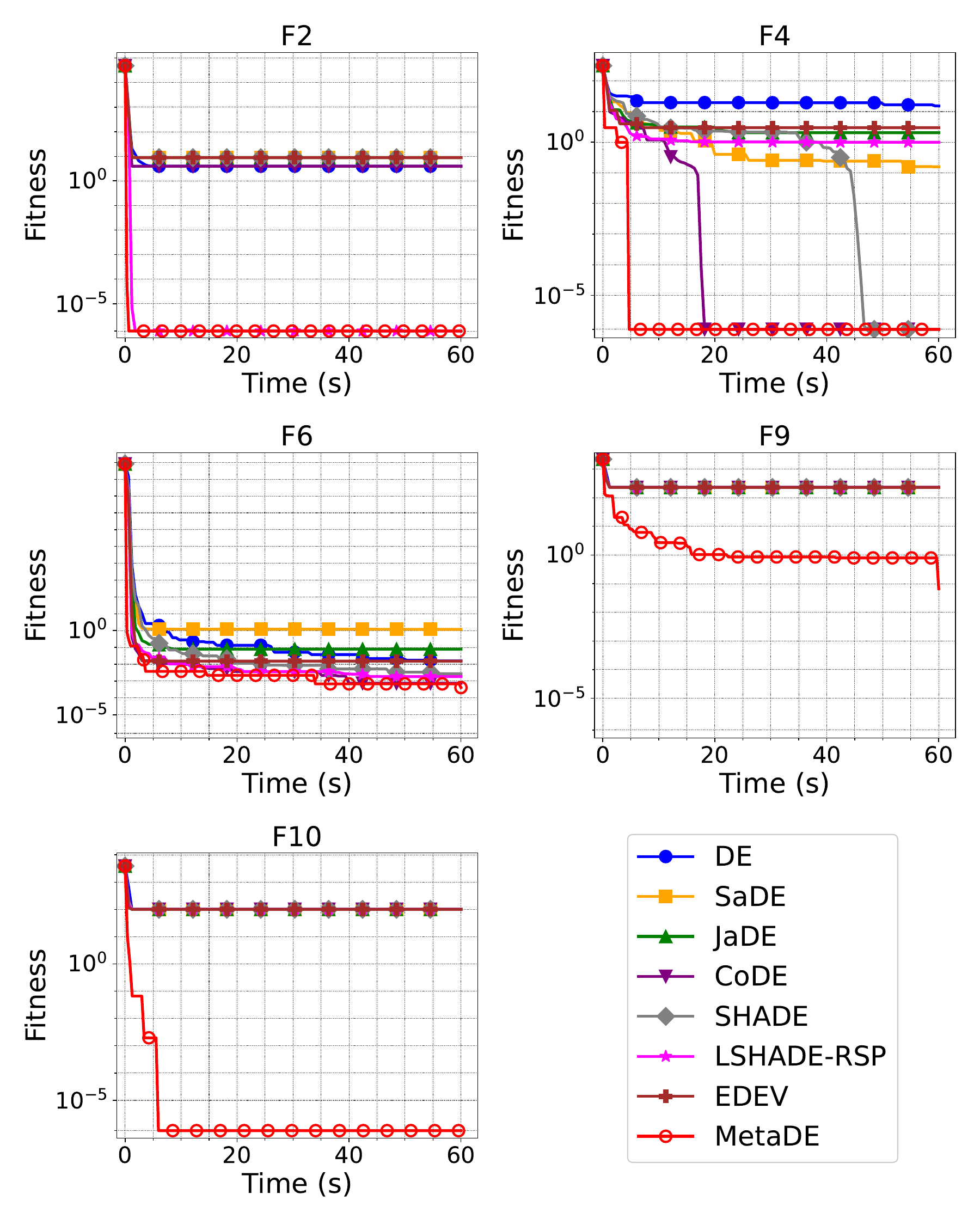}
\caption{Convergence curves on 10D problems in CEC2022 benchmark suite. The peer DE variants are set with population size of 1,000.}
\label{Figure_convergence_10D_NP10k}
\end{figure}

\begin{figure}[!t]
\centering
\includegraphics[scale=0.3]{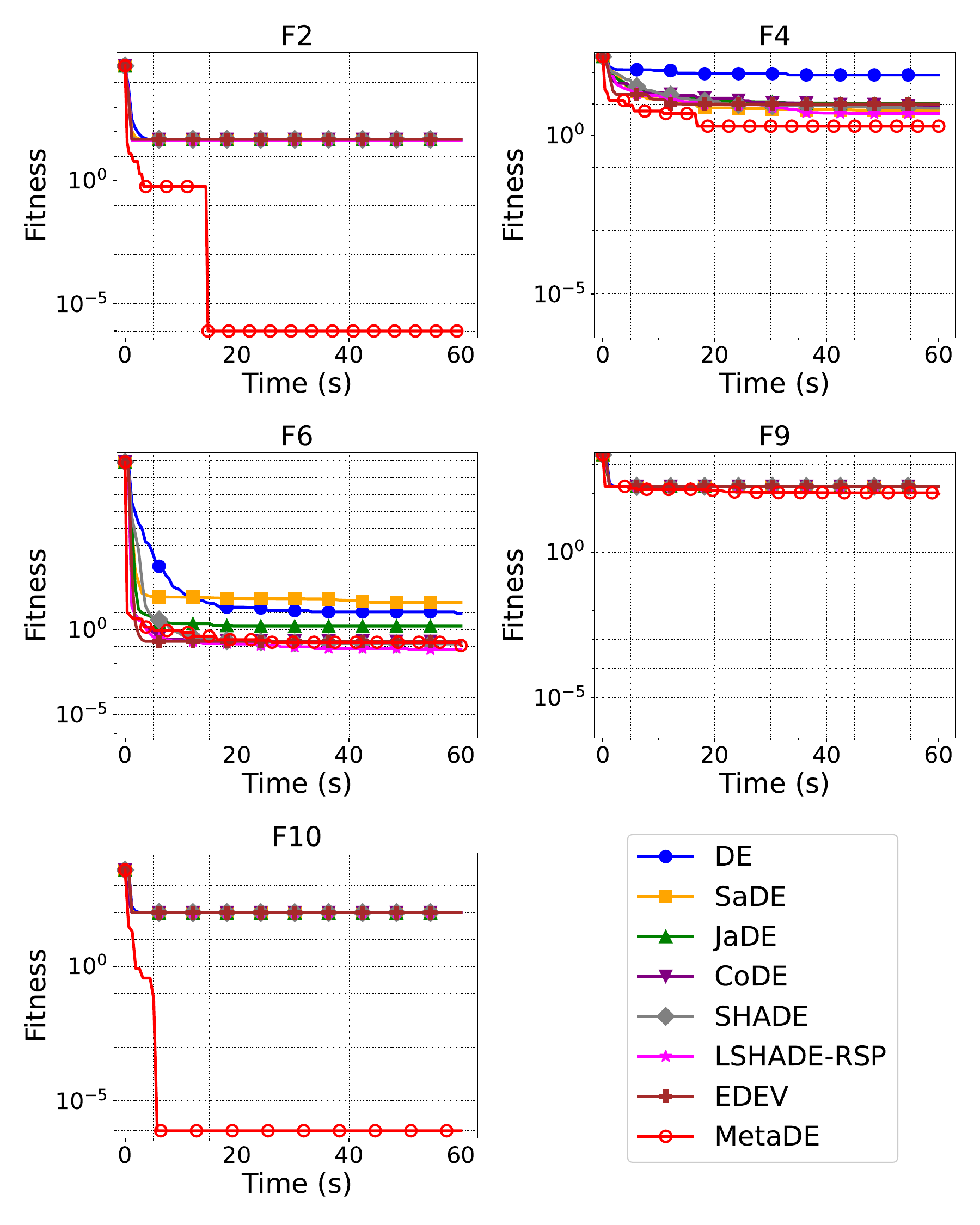}
\caption{Convergence curves on 20D problems in CEC2022 benchmark suite. The peer DE variants are set with population size of 1,000.}
\label{Figure_convergence_20D_NP10k}
\end{figure}

{
\subsection{Comparisons with Top Algorithms in CEC2022 Competition}\label{section_Comparison with Top Algorithms of CEC Competition}

\begin{table*}[htbp]
  \centering
  
  \caption{{Comparisons between MetaDE and the top 4 algorithms from CEC2022 Competition (10D). 
The mean and standard deviation (in parentheses) of the results over multiple runs are displayed in pairs. 
Results with the best mean values are highlighted. }
  }
\footnotesize
% Table generated by Excel2LaTeX from sheet 'Exp 7 vs CECtop'
\begin{tabular}{cccccc}
\toprule
Func  & MetaDE & EA4eig & NL-SHADE-LBC & NL-SHADE-RSP & S-LSHADE-DP \\
\midrule
$F_{1}$ & \textbf{0.00E+00 (0.00E+00)} & \boldmath{}\textbf{0.00E+00 (0.00E+00)$\approx$}\unboldmath{} & \boldmath{}\textbf{0.00E+00 (0.00E+00)$\approx$}\unboldmath{} & \boldmath{}\textbf{0.00E+00 (0.00E+00)$\approx$}\unboldmath{} & \boldmath{}\textbf{0.00E+00 (0.00E+00)$\approx$}\unboldmath{} \\
$F_{2}$ & \textbf{0.00E+00 (0.00E+00)} & 7.97E-01 (1.78E+00)$-$ & 7.97E-01 (1.78E+00)$-$ & \boldmath{}\textbf{0.00E+00 (0.00E+00)$\approx$}\unboldmath{} & \boldmath{}\textbf{0.00E+00 (0.00E+00)$\approx$}\unboldmath{} \\
$F_{3}$ & \textbf{0.00E+00 (0.00E+00)} & \boldmath{}\textbf{0.00E+00 (0.00E+00)$\approx$}\unboldmath{} & \boldmath{}\textbf{0.00E+00 (0.00E+00)$\approx$}\unboldmath{} & \boldmath{}\textbf{0.00E+00 (0.00E+00)$\approx$}\unboldmath{} & \boldmath{}\textbf{0.00E+00 (0.00E+00)$\approx$}\unboldmath{} \\
$F_{4}$ & \textbf{0.00E+00 (0.00E+00)} & 9.95E-01 (1.22E+00)$-$ & 1.99E-01 (4.45E-01)$-$ & 2.98E+00 (1.15E+00)$-$ & \boldmath{}\textbf{0.00E+00 (0.00E+00)$\approx$}\unboldmath{} \\
$F_{5}$ & \textbf{0.00E+00 (0.00E+00)} & \boldmath{}\textbf{0.00E+00 (0.00E+00)$\approx$}\unboldmath{} & \boldmath{}\textbf{0.00E+00 (0.00E+00)$\approx$}\unboldmath{} & \boldmath{}\textbf{0.00E+00 (0.00E+00)$\approx$}\unboldmath{} & \boldmath{}\textbf{0.00E+00 (0.00E+00)$\approx$}\unboldmath{} \\
$F_{6}$ & 5.50E-04 (3.96E-04) & 7.53E-04 (5.52E-04)$\approx$ & 8.93E-02 (1.18E-01)$-$ & 4.37E-02 (5.41E-02)$-$ & \textbf{5.84E-05 (4.73E-05)$+$} \\
$F_{7}$ & \textbf{0.00E+00 (0.00E+00)} & \boldmath{}\textbf{0.00E+00 (0.00E+00)$\approx$}\unboldmath{} & \boldmath{}\textbf{0.00E+00 (0.00E+00)$\approx$}\unboldmath{} & \boldmath{}\textbf{0.00E+00 (0.00E+00)$\approx$}\unboldmath{} & \boldmath{}\textbf{0.00E+00 (0.00E+00)$\approx$}\unboldmath{} \\
$F_{8}$ & 5.52E-03 (4.41E-03) & 1.01E-04 (1.66E-04)$+$ & 3.96E-04 (4.23E-04)$+$ & 3.13E-01 (3.60E-01)$-$ & \textbf{1.26E-05 (1.56E-05)$+$} \\
$F_{9}$ & \textbf{3.36E+00 (1.77E+01)} & 1.86E+02 (0.00E+00)$-$ & 2.29E+02 (3.18E-14)$-$ & 8.03E+01 (1.08E+02)$-$ & 2.23E+02 (1.31E+01)$-$ \\
$F_{10}$ & \textbf{0.00E+00 (0.00E+00)} & 1.00E+02 (0.00E+00)$-$ & 1.00E+02 (0.00E+00)$-$ & 1.56E-02 (3.12E-02)$-$ & \boldmath{}\textbf{0.00E+00 (0.00E+00)$\approx$}\unboldmath{} \\
$F_{11}$ & \textbf{0.00E+00 (0.00E+00)} & \boldmath{}\textbf{0.00E+00 (0.00E+00)$\approx$}\unboldmath{} & \boldmath{}\textbf{0.00E+00 (0.00E+00)$\approx$}\unboldmath{} & \boldmath{}\textbf{0.00E+00 (0.00E+00)$\approx$}\unboldmath{} & \boldmath{}\textbf{0.00E+00 (0.00E+00)$\approx$}\unboldmath{} \\
$F_{12}$ & \textbf{1.39E+02 (4.63E+01)} & 1.48E+02 (5.98E+00)$-$ & 1.65E+02 (0.00E+00)$-$ & 1.62E+02 (2.15E+00)$-$ & 1.59E+02 (0.00E+00)$-$ \\
\midrule
$+$ / $\approx$ / $-$ & --    & 1/6/5 & 1/5/6 & 0/6/6 & 2/8/2 \\
\bottomrule
\end{tabular}%

\footnotesize
\textsuperscript{*} The Wilcoxon rank-sum tests (with a significance level of 0.05) were conducted between MetaDE and each algorithm individually.
The final row displays the number of problems where the corresponding algorithm performs statistically better ($+$),  similar ($\thickapprox$), or worse ($-$) compared to MetaDE.\\

\label{tab:vsCECTop 10D}%
\end{table*}%

\begin{table*}[htbp]
  \centering
  
  \caption{{Comparisons between MetaDE and the top 4 algorithms from CEC2022 Competition (20D). 
The mean and standard deviation (in parentheses) of the results over multiple runs are displayed in pairs. 
Results with the best mean values are highlighted.
  }
  }
  %\resizebox{\linewidth}{!}{
  %       \renewcommand{\arraystretch}{1}
 %\renewcommand{\tabcolsep}{3pt}
\footnotesize
% Table generated by Excel2LaTeX from sheet 'Exp 7 vs CECtop'
\begin{tabular}{cccccc}
\toprule
Func  & MetaDE & EA4eig & NL-SHADE-LBC & NL-SHADE-RSP & S-LSHADE-DP \\
\midrule
$F_{1}$ & \textbf{0.00E+00 (0.00E+00)} & \boldmath{}\textbf{0.00E+00 (0.00E+00)$\approx$}\unboldmath{} & \boldmath{}\textbf{0.00E+00 (0.00E+00)$\approx$}\unboldmath{} & \boldmath{}\textbf{0.00E+00 (0.00E+00)$\approx$}\unboldmath{} & \boldmath{}\textbf{0.00E+00 (0.00E+00)$\approx$}\unboldmath{} \\
$F_{2}$ & 3.83E-04 (2.10E-03) & \textbf{0.00E+00 (0.00E+00)$+$} & 4.91E+01 (0.00E+00)$-$ & \textbf{0.00E+00 (0.00E+00)$+$} & \textbf{0.00E+00 (0.00E+00)$+$} \\
$F_{3}$ & \textbf{0.00E+00 (0.00E+00)} & \boldmath{}\textbf{0.00E+00 (0.00E+00)$\approx$}\unboldmath{} & \boldmath{}\textbf{0.00E+00 (0.00E+00)$\approx$}\unboldmath{} & \boldmath{}\textbf{0.00E+00 (0.00E+00)$\approx$}\unboldmath{} & \boldmath{}\textbf{0.00E+00 (0.00E+00)$\approx$}\unboldmath{} \\
$F_{4}$ & 1.96E+00 (7.76E-01) & 7.36E+00 (2.06E+00)$-$ & \boldmath{}\textbf{1.59E+00 (5.45E-01)$\approx$}\unboldmath{} & 1.07E+02 (1.54E+02)$-$ & 3.20E+00 (1.94E+00)$-$ \\
$F_{5}$ & \textbf{0.00E+00 (0.00E+00)} & \boldmath{}\textbf{0.00E+00 (0.00E+00)$\approx$}\unboldmath{} & \boldmath{}\textbf{0.00E+00 (0.00E+00)$\approx$}\unboldmath{} & 2.27E-01 (4.54E-01)$-$ & \boldmath{}\textbf{0.00E+00 (0.00E+00)$\approx$}\unboldmath{} \\
$F_{6}$ & \textbf{1.38E-01 (5.56E-02)} & 2.54E-01 (4.28E-01)$-$ & 3.06E-01 (2.01E-01)$-$ & 2.08E-01 (9.78E-02)$-$ & 5.02E-01 (5.34E-01)$-$ \\
$F_{7}$ & 8.42E-02 (1.01E-01) & 1.37E+00 (1.10E+00)$-$ & \boldmath{}\textbf{6.24E-02 (1.40E-01)$\approx$}\unboldmath{} & 1.28E+00 (1.95E+00)$-$ & 9.83E-01 (8.12E-01)$-$ \\
$F_{8}$ & 2.66E+00 (3.83E+00) & 2.02E+01 (1.28E-01)$-$ & \textbf{1.01E-01 (1.41E-01)$+$} & 1.99E+01 (4.97E-01)$-$ & 2.30E-01 (1.82E-01)$+$ \\
$F_{9}$ & \textbf{1.32E+02 (3.43E+01)} & 1.65E+02 (0.00E+00)$-$ & 1.81E+02 (0.00E+00)$-$ & 1.81E+02 (0.00E+00)$-$ & 1.81E+02 (0.00E+00)$-$ \\
$F_{10}$ & \textbf{0.00E+00 (0.00E+00)} & 1.23E+02 (5.12E+01)$-$ & 1.00E+02 (9.27E-03)$-$ & \boldmath{}\textbf{0.00E+00 (0.00E+00)$\approx$}\unboldmath{} & \boldmath{}\textbf{0.00E+00 (0.00E+00)$\approx$}\unboldmath{} \\
$F_{11}$ & 1.74E-03 (7.97E-03) & 3.20E+02 (4.47E+01)$-$ & 3.00E+02 (0.00E+00)$-$ & \textbf{0.00E+00 (0.00E+00)$+$} & \textbf{0.00E+00 (0.00E+00)$+$} \\
$F_{12}$ & 2.29E+02 (9.70E-01) & \textbf{2.00E+02 (2.04E-04)$+$} & 2.37E+02 (3.17E+00)$-$ & 2.34E+02 (1.46E+00)$-$ & 2.34E+02 (4.51E+00)$-$ \\
\midrule
$+$ / $\approx$ / $-$ & --    & 2/3/7 & 1/5/6 & 2/3/7 & 3/4/5 \\
\bottomrule
\end{tabular}%

\footnotesize
\textsuperscript{*} The Wilcoxon rank-sum tests (with a significance level of 0.05) were conducted between MetaDE and each algorithm individually.
The final row displays the number of problems where the corresponding algorithm performs statistically better ($+$),  similar ($\thickapprox$), or worse ($-$) compared to MetaDE.\\

\label{tab:vsCECTop 20D}%
\end{table*}%

To further assess the performance of MetaDE, we compare it with the top 4 algorithms from the CEC2022 Competition on Single Objective Bound Constrained Numerical Optimization\footnote{\url{https://github.com/P-N-Suganthan/2022-SO-BO}}.
For each algorithm, we set equal FEs as achieved by MetaDE within 60 seconds (refer to Table~\ref{tab:FEs} for details).

The top 4 algorithms from the CEC2022 Competition are {EA4eig}~\cite{EA4eig}, {NL-SHADE-LBC}~\cite{NL-SHADE-LBC}, {NL-SHADE-RSP-MID}~\cite{NL-SHADE-RSP}, and {S-LSHADE-DP}~\cite{S_LSHADE_DP}:
\begin{itemize}
  \item {EA4eig} combines the strengths of four evolutionary algorithms (CMA-ES, CoBiDE, an adaptive variant of jSO, and IDE) using Eigen crossover.
  \item {NL-SHADE-LBC} is a dynamic DE variant that integrates linear bias changes for parameter adaptation, repeated point generation to handle boundary constraints, non-linear population size reduction, and a selective pressure mechanism.
  \item {NL-SHADE-RSP-MID} is an advanced version of NL-SHADE-RSP, which estimates the optimum using the population midpoint, incorporates a restart mechanism, and improves boundary constraint handling.
  \item {S-LSHADE-DP} focuses on maintaining population diversity through dynamic perturbation, adjusting noise intensity to enhance exploration.
\end{itemize}

The experimental results are summarized in Tables \ref{tab:vsCECTop 10D} and \ref{tab:vsCECTop 20D}.
On 10D problems, MetaDE outperforms EA4eig, NL-SHADE-LBC, and NL-SHADE-RSP, while achieving comparable performance to S-LSHADE-DP. 
On 20D problems, MetaDE consistently outperforms the four algorithms.
An additional noteworthy observation is that S-LSHADE-DP exhibits promising performance under a large number of FEs.
}

\subsection{Investigation of Optimal DE Variants}\label{section_Optimal Parameter Analysis}

\begin{table}[h]
  \centering
  \caption{Optimal DE variants obtained by MetaDE on each problem of the CEC2022 benchmark suite. FDC and RIE are two fitness landscape characteristics that measure the difficulty and ruggedness of the problem.}
% Table generated by Excel2LaTeX from sheet 'Exp4 param'
\resizebox{\columnwidth}{!}{
\begin{tabular}{cccccccc}
\toprule
\multicolumn{2}{c}{Problem} & F     & CR    & \multicolumn{2}{c}{Strategy} & FDC & RIE \\
\midrule
\multirow{4}[2]{*}{10D} & $F_{6}$ & 0.70  & 0.99  & \multicolumn{2}{c}{\texttt{rand-to-pbest/1/arith}} & 0.61  & 0.81  \\
      & $F_{8}$ & 0.51  & 0.44  & \multicolumn{2}{c}{\texttt{pbest-to-best/1/bin}} & 0.27  & 0.62  \\
      & $F_{9}$ & 0.02  & 0.03  & \multicolumn{2}{c}{\texttt{current/2/bin}} & 0.08  & 0.82  \\
      & $F_{12}$ & 0.16  & 0.00  & \multicolumn{2}{c}{\texttt{current-to-best/4/bin}} & -0.15  & 0.78  \\
\midrule
\multirow{6}[2]{*}{20D} & $F_{4}$ & 0.13  & 0.71  & \multicolumn{2}{c}{\texttt{rand-to-best/3/bin}} & 0.90  & 0.79  \\
      & $F_{6}$ & 0.67  & 0.99  & \multicolumn{2}{c}{\texttt{pbest-to-rand/1/bin}} & 0.48  & 0.80  \\
      & $F_{7}$ & 0.27  & 0.93  & \multicolumn{2}{c}{\texttt{rand/2/bin}} & 0.26  & 0.78  \\
      & $F_{8}$ & 0.65  & 0.00  & \multicolumn{2}{c}{\texttt{pbest/1/exp}} & 0.12  & 0.40  \\
      & $F_{9}$ & 0.06  & 0.00  & \multicolumn{2}{c}{\texttt{current/2/bin}} & -0.17  & 0.84  \\
      & $F_{12}$ & 0.33  & 0.44  & \multicolumn{2}{c}{\texttt{rand-to-best/2/bin}} & -0.16  & 0.85  \\
\bottomrule
\end{tabular}%
}
\label{tab:optimal_param}
\end{table}%

This part provides an in-depth examination of the optimal DE variants obtained by MetaDE in Section~\ref{sec:expereiment_time}, as summarized in Table \ref{tab:optimal_param}.
The optimal parameters correspond to the best individual in the final population of MetaDE.
The table only displays the optimal parameters for the ten listed problems, as the remaining problems are relatively simpler, with numerous DE variants capable of locating the optimal solutions of the problems. Furthermore, the optimal parameters presented in the table represent the best results of MetaDE derived from the finest run out of 31 independent trials.

All the problems in the table are characterized by both multimodality and non-separability.
Additionally, to further depict the characteristics of the problems' fitness landscapes, we computed both the fitness distance correlation (FDC) \cite{FDC} and the ruggedness of information entropy (RIE) \cite{RIE}; the former measures the complexity (difficulty) of the problems, while the latter characterizes the ruggedness of the landscape.

Analyzing the obtained data, it is evident that no single set of parameters or strategies consistently excels across all problems.
Parameters such as \(F\) and \(CR\) exhibit variability across problems without adhering to a specific trend.
Similarly, the selection of base vectors ($bl$ and $br$) does not show a uniform preference either.
Regarding the fitness landscape characteristics of each problem, the selection of parameters exhibits distinct patterns.
The FDC indicates problem complexity; with simpler problems (higher FDC), such as 10-dimensional $F_6$, $F_8$ and 20-dimensional $F_4$, $F_6$, $F_7$, a larger \(CR\) value is favored. Conversely, smaller \(CR\) values are chosen for problems with lower FDC. A larger \(CR\) tends to facilitate convergence, whereas a \(CR\) close to 0 leads to offspring that change incrementally, dimension by dimension. However, the other characteristic, RIE, does not seem to have a clear association with parameter choices.
The optimal strategies for identical problems across different dimensions exhibit closeness, with $F_8$, $F_9$, and $F_{12}$ demonstrating notably parallel strategies between their 10D and 20D problems.
In terms of crossover strategies ($cs$), it seems to have a preference for binomial crossover. This aligns with the traditional DE configurations.

These observations align with the No Free Lunch (NFL) theorem \cite{NFL}, thus underscoring the importance of distinct optimization strategies tailored for diverse problems.
Conventionally, the optimization strategies have oscillated between seeking a generalist set of parameters for broad applicability and a specialist set tailored for specific problems. However, the dynamic nature of optimization problems, where even minute changes like a different random seed can pivot the problem's dynamics, highlights the challenges of a generalist approach.
In contrast, MetaDE provides a simple yet effective approach, showing promising generality and adaptability.

\subsection{Application to Robot Control}\label{sec:expereiment_brax}
In this experiment, we demonstrate the extended application of MetaDE to robot control.
Specifically, we adopted the evolutionary reinforcement learning paradigm~\cite{ERL} as illustrated in Fig.~\ref{Figure_EvoRL}.
The experiment was conducted on Brax \cite{brax} for robotics simulations with GPU acceleration.

\begin{figure}[!h]
\centering
\includegraphics[scale=0.38]{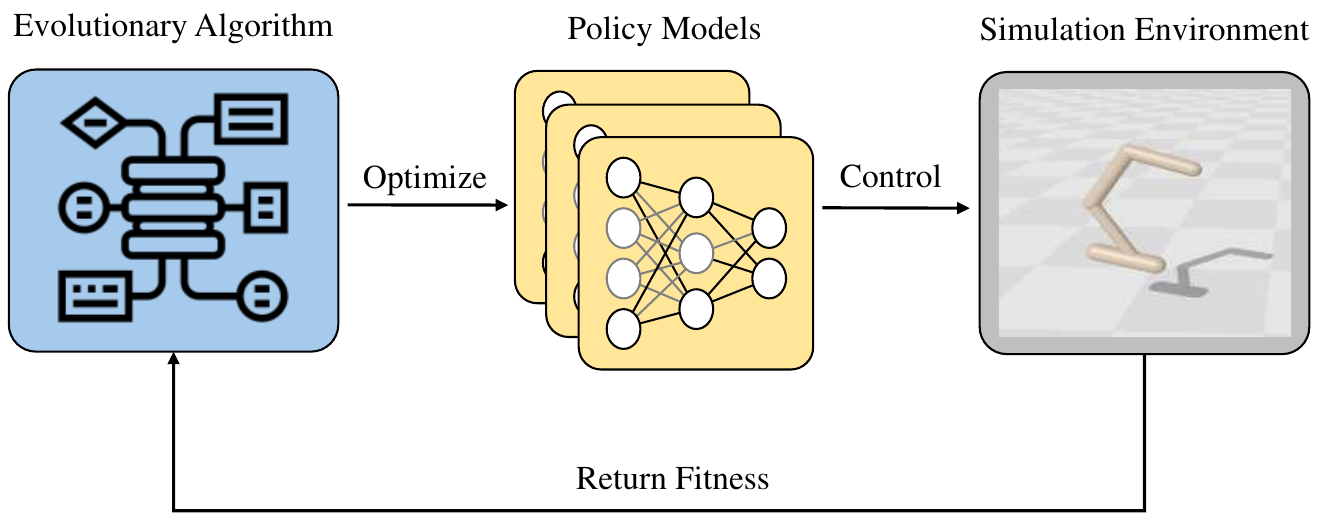}
\caption{Illustration of robot control via evolutionary reinforcement learning. The evolutionary algorithm optimizes the parameters of a population of candidate policy models for controlling the robotics behaviors. The simulation environment returns rewards achieved by the candidate policy models to the evolutionary algorithm as fitness values.}
\label{Figure_EvoRL}
\end{figure}

This experiment involved three robot control tasks: ``swimmer'', ``hopper'', and ``reacher''.
As summarized in Table \ref{tab:Neural network structures}, we adopted similar policy models for these three tasks, each consisting of a multilayer perceptron (MLP) with three fully connected layers, but with different input and output dimensions.
Consequently, the three policy models comprise 1410, 1539, and 1506 parameters for optimization respectively, where the optimization objective is to achieve maximum reward of each task.
MetaDE, vanilla DE \cite{DE1996}, SHADE~\cite{SHADE2013}, LSHADE-RSP~\cite{LSHADE-RSP2018}, EDEV~\cite{EDEV2018}, CSO \cite{CSO}\footnote{The competitive swarm optimizer (CSO) is a tailored PSO variant for large-scale optimization.}, and CMA-ES~\cite{CMAES} were applied as the optimizer respectively.

%The policy models  were initialized with identical random parameters.
The iteration count for PDE within MetaDE was set to 50, while other algorithms maintained a population size of 100.
Each algorithm was run independently 15 times.
Considering the time-intensive nature of the robotics simulations, we set 60 minutes as the termination condition for each run.

\begin{table}[htbp]
\centering
\caption{Neural network structure of the policy model for each robot control task}
\label{tab:Neural network structures}
\resizebox{\columnwidth}{!}{%
% Table generated by Excel2LaTeX from sheet 'Exp6 brax'
\begin{tabular}{cccccc}
\toprule
\textbf{Task} & \textbf{D} & \textbf{Input} & \textbf{Hidden Layers} &   \textbf{Output}    & \textbf{Overview of objectives} \\
\midrule
Hopper & 1539  & 11    & 32$\times$32 & 3     & balance and jump \\
Swimmer & 1410  & 8     & 32$\times$32 & 2     & maximizing movement \\
Reacher & 1506  & 11    & 32$\times$32 & 2     & precise reaching \\
\bottomrule
\end{tabular}%
}
\end{table}

\begin{figure}[!h]
\centering
\includegraphics[scale=0.3]{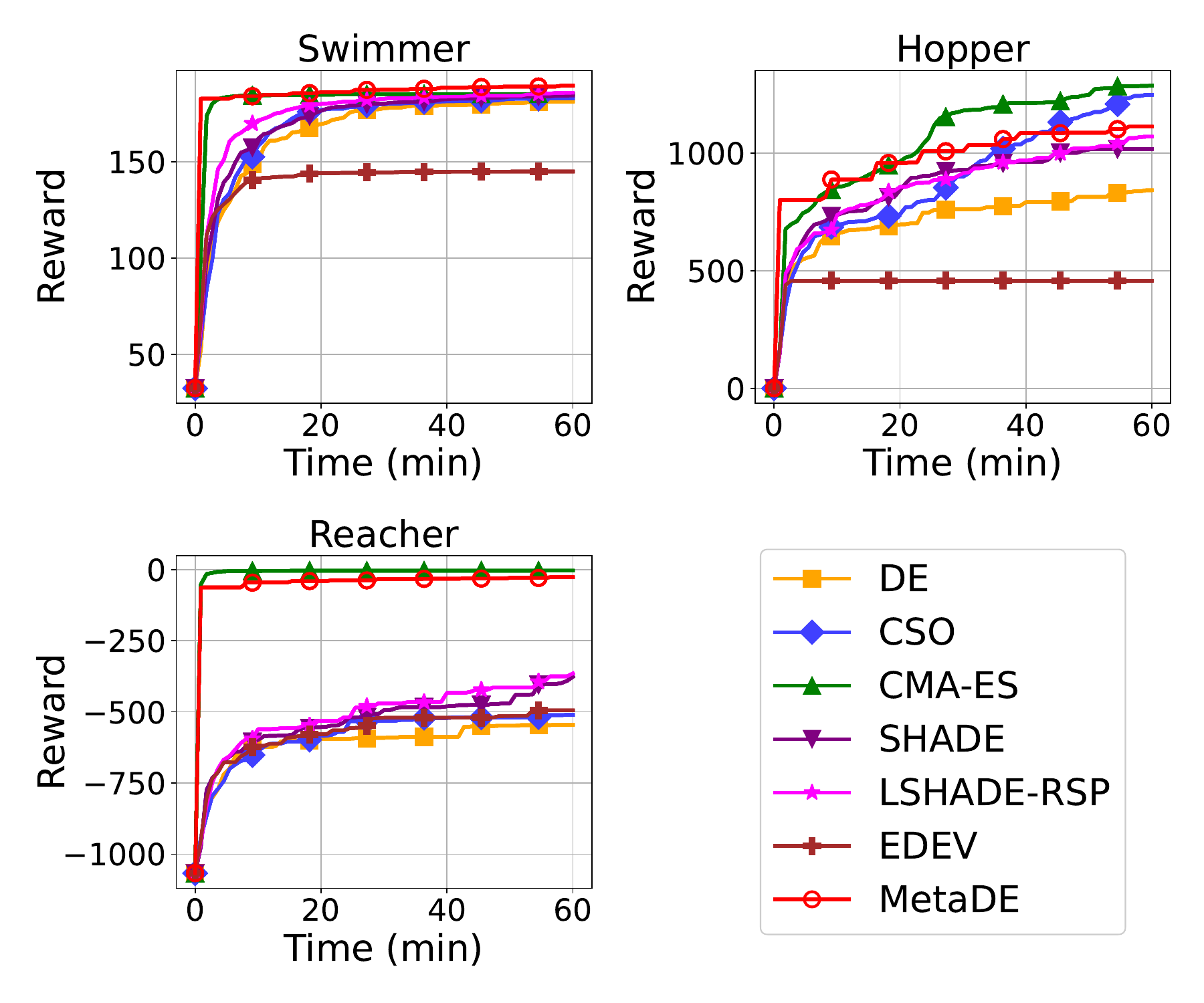}
\caption{The reward curves achieved by MetaDE and peer evolutionary algorithms when applied to each robot control task. }
\label{Figure_brax_all_convergence}
\end{figure}

\begin{figure}[!h]
\centering
\includegraphics[width=\linewidth]{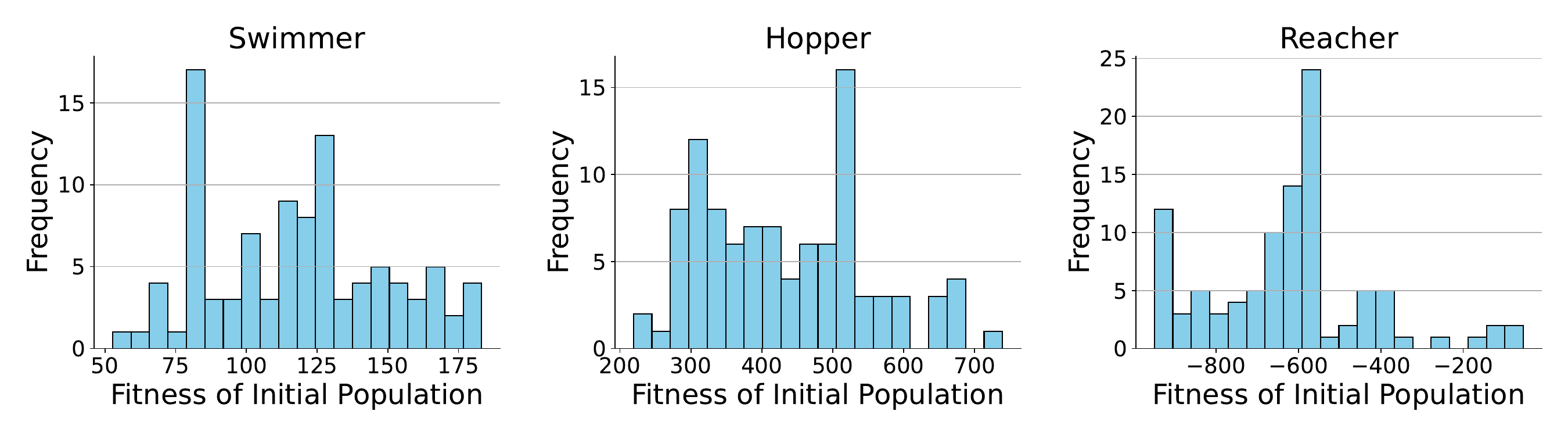}
\caption{The fitness distribution of MetaDE's initial population when applied to each robot control task.}
\label{Figure_brax_distribution}
\end{figure}

As shown in Fig. \ref{Figure_brax_all_convergence}, it is evident that MetaDE achieves the best performance in the Swimmer tasks, while slightly outperformed by CMA-ES and CSO in the Hopper and Reacher task.
An interesting observation from the reward curves is that MetaDE almost reaches optimality nearly at the first generation and does not show further significant improvements thereafter.
To elucidate this phenomenon, Fig.~\ref{Figure_brax_distribution} provides the fitness distribution of MetaDE's initial population, indicating that MetaDE harbored several individuals with considerably high fitness from the initial generation.
In other words, MetaDE was able to generate high-performance DE variants for these problems even by random sampling.
This can be attributed to the unique nature of neural network optimization.
As widely acknowledged, the neural network optimization typically features numerous plateaus in the fitness landscape, thus making it relatively easy to find one of the local optima.
MetaDE provides unbiased sampling of parameter settings for generating diverse DE variants.
Even without further evolution, some of the randomly sampled DE variants are very likely to reach the plateaus.
In contrast, the other algorithms are specially tailored with biases; in such large-scale optimization scenarios, the biases can be further amplified, thus making them ineffective.

\section{Conclusion}\label{section Conclusion}
In this paper, we introduced MetaDE, a method that leverages the strengths of DE not only to address optimization tasks but also to adapt and refine its own strategies. This meta-evolutionary approach demonstrates how DE can autonomously evolve its parameter configurations and strategies. 
Our experiments demonstrate that MetaDE has robust performance across various benchmarks, as well as the application in robot control through evolutionary reinforcement learning. 
Nevertheless, the study also emphasizes the complexity of finding universally optimal parameter configurations. The intricate balance between generalization and specialization remains a challenge, and MetaDE has shed light on further research into self-adapting algorithms. 
We anticipate that the insights gained from this work will inspire the development of more advanced meta-evolutionary approaches, pushing the boundaries of evolutionary optimization in even more complex and dynamic environments.

\footnotesize

% \bibliography{manuscript_references}

\newpage

\renewcommand\thealgorithm{S.\arabic{algorithm}}
\renewcommand\thetable{S.\arabic{table}}
\renewcommand\thefigure{S.\arabic{figure}}
\renewcommand\thesection{S.\roman{section}}
\renewcommand\theequation{S.\arabic{equation}}

\title{Supplementary Document for ``MetaDE: Evolving Differential Evolution by Differential Evolution"}
\author{Minyang Chen, Chenchen Feng,
        and Ran Cheng
        \thanks{
        Minyang Chen was with the Department of Computer Science and Engineering, Southern University of Science and Technology, Shenzhen 518055, China. E-mail: cmy1223605455@gmail.com. }
        \thanks{
        Chenchen Feng is with the Department of Computer Science and Engineering, Southern University of Science and Technology, Shenzhen 518055, China. E-mail: chenchenfengcn@gmail.com. 
        }
        \thanks{
       Ran Cheng is with the Department of Data Science and Artificial Intelligence, and the Department of Computing, The Hong Kong Polytechnic University, Hong Kong SAR, China. E-mail: ranchengcn@gmail.com. (\emph{Corresponding author: Ran Cheng})
        }
        }

\onecolumn{}

\markboth{Journal of \LaTeX\ Class Files,~Vol.~0, No.~0, 0~0}%
{Shell \MakeLowercase{\textit{et al.}}: Bare Demo of IEEEtran.cls for IEEE Journals}




\maketitle



% \fi
\IEEEpeerreviewmaketitle

% \clearpage
% \begin{titlepage}
% \centering
% {\LARGE Supplementary Document for ``MetaDE: Evolving Differential Evolution by Differential Evolution"}\\[1.5cm]
% {\large Minyang Chen, Chenchen Feng, and Ran Cheng}\\[1cm]
% {\small
% Minyang Chen was with the Department of Computer Science and Engineering, Southern University of Science and Technology, Shenzhen 518055, China. E-mail: cmy1223605455@gmail.com.\\[0.2cm]
% Chenchen Feng is with the Department of Computer Science and Engineering, Southern University of Science and Technology, Shenzhen 518055, China. E-mail: chenchenfengcn@gmail.com.\\[0.2cm]
% Ran Cheng is with the Department of Data Science and Artificial Intelligence, and the Department of Computing, The Hong Kong Polytechnic University, Hong Kong SAR, China. E-mail: ranchengcn@gmail.com. (Corresponding author: Ran Cheng)
% }\\[2cm]
% \end{titlepage}

\clearpage
\begin{center}
  {\Huge Supplementary Document for ``MetaDE: Evolving \\[0.3em]
  Differential Evolution by Differential Evolution"}\\[2em]
  {\Large Minyang Chen, Chenchen Feng, and Ran Cheng}\\[6em]
\end{center}

\section{Supplementary Experimental data}\label{section:FEs}

\subsection{Supplementary Figures}\label{section:FEs}

\begin{figure*}[htpb]
\centering
\includegraphics[scale=0.3]{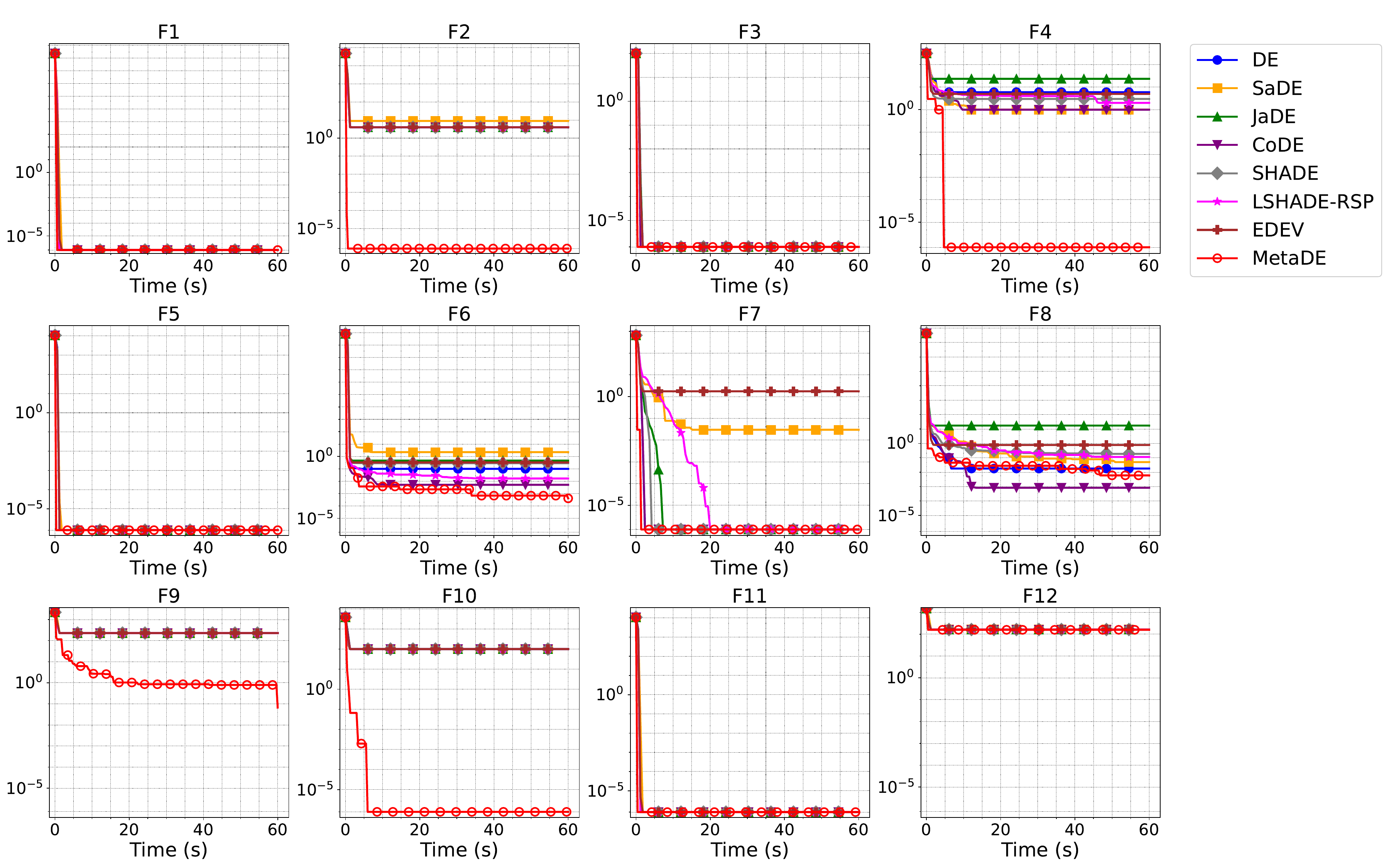}
\caption{Convergence curves on 10D problems in CEC2022 benchmark suite. The peer DE variants are set with population size of 100.}
\label{Figure_convergence_10D_supp}
\end{figure*}

\vfill 
{\small
\noindent

Minyang Chen was with the Department of Computer Science and Engineering, Southern University of Science and Technology, Shenzhen 518055, China. E-mail: cmy1223605455@gmail.com.

Chenchen Feng is with the Department of Computer Science and Engineering, Southern University of Science and Technology, Shenzhen 518055, China. E-mail: chenchenfengcn@gmail.com.

Ran Cheng is with the Department of Data Science and Artificial Intelligence, and the Department of Computing, The Hong Kong Polytechnic University, Hong Kong SAR, China. E-mail: ranchengcn@gmail.com. \textit{(Corresponding author: Ran Cheng)}
}

\begin{figure*}[htpb]
\centering
\includegraphics[scale=0.3]{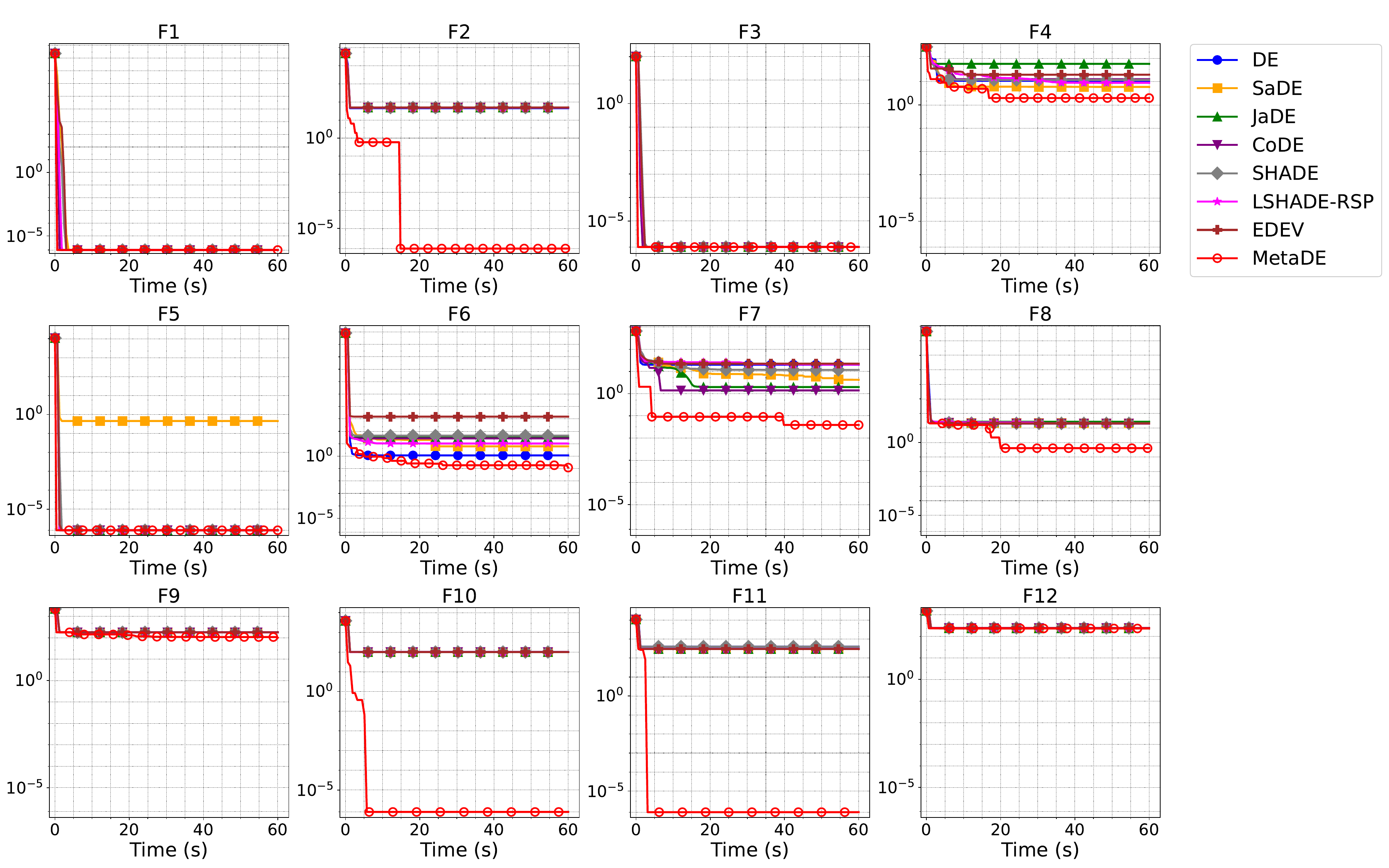}
\caption{Convergence curves on 20D problems in CEC2022 benchmark suite. The peer DE variants are set with population size of 100.}
\label{Figure_convergence_20D_supp}
\end{figure*}

\begin{figure*}[htpb]
\centering
\includegraphics[scale=0.3]{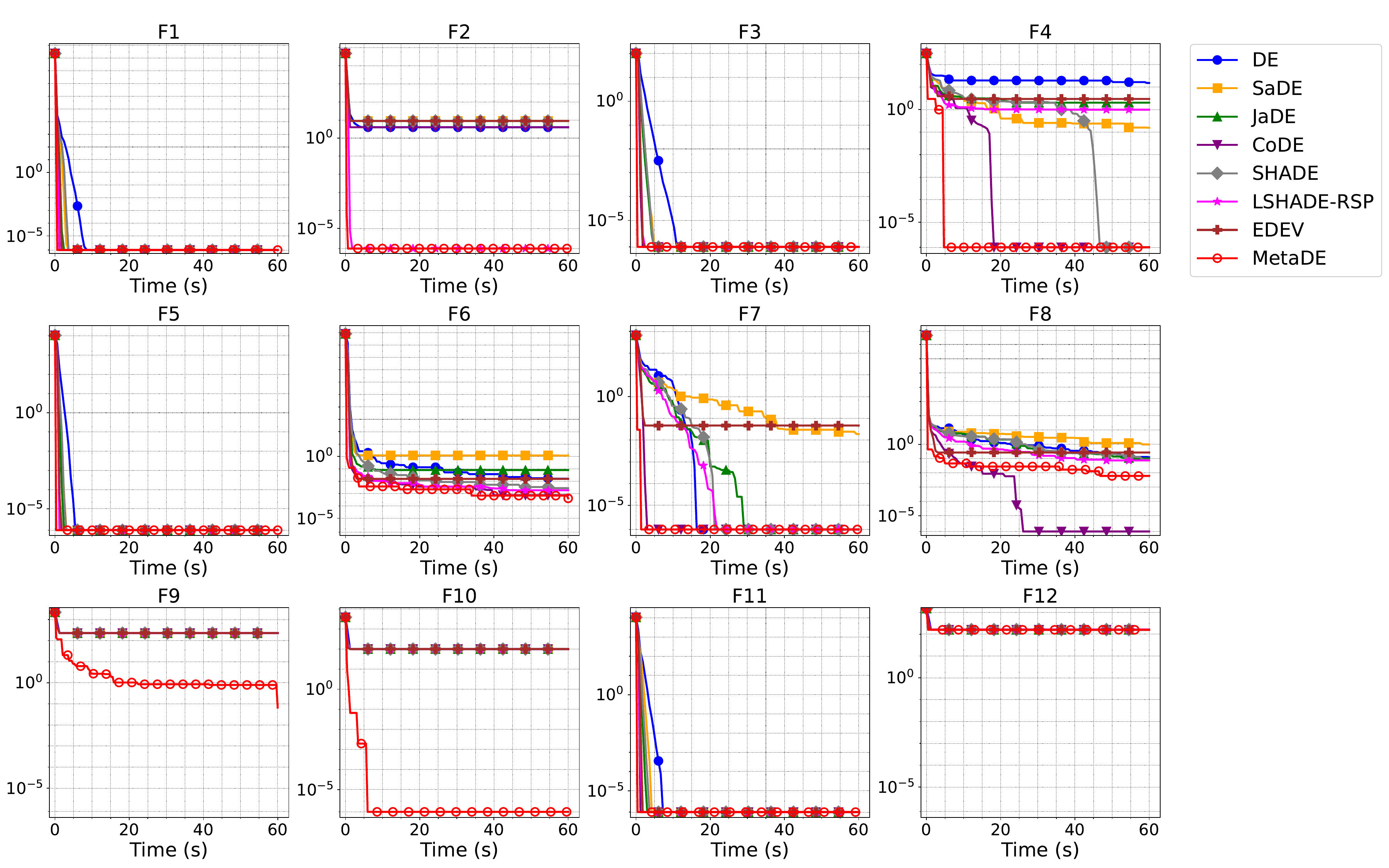}
\caption{Convergence curves on 10D problems in CEC2022 benchmark suite. The peer DE variants are set with population size of 1,000.}
\label{Figure_convergence_10D_NP10k_supp}
\end{figure*}

\begin{figure*}[htpb]
\centering
\includegraphics[scale=0.3]{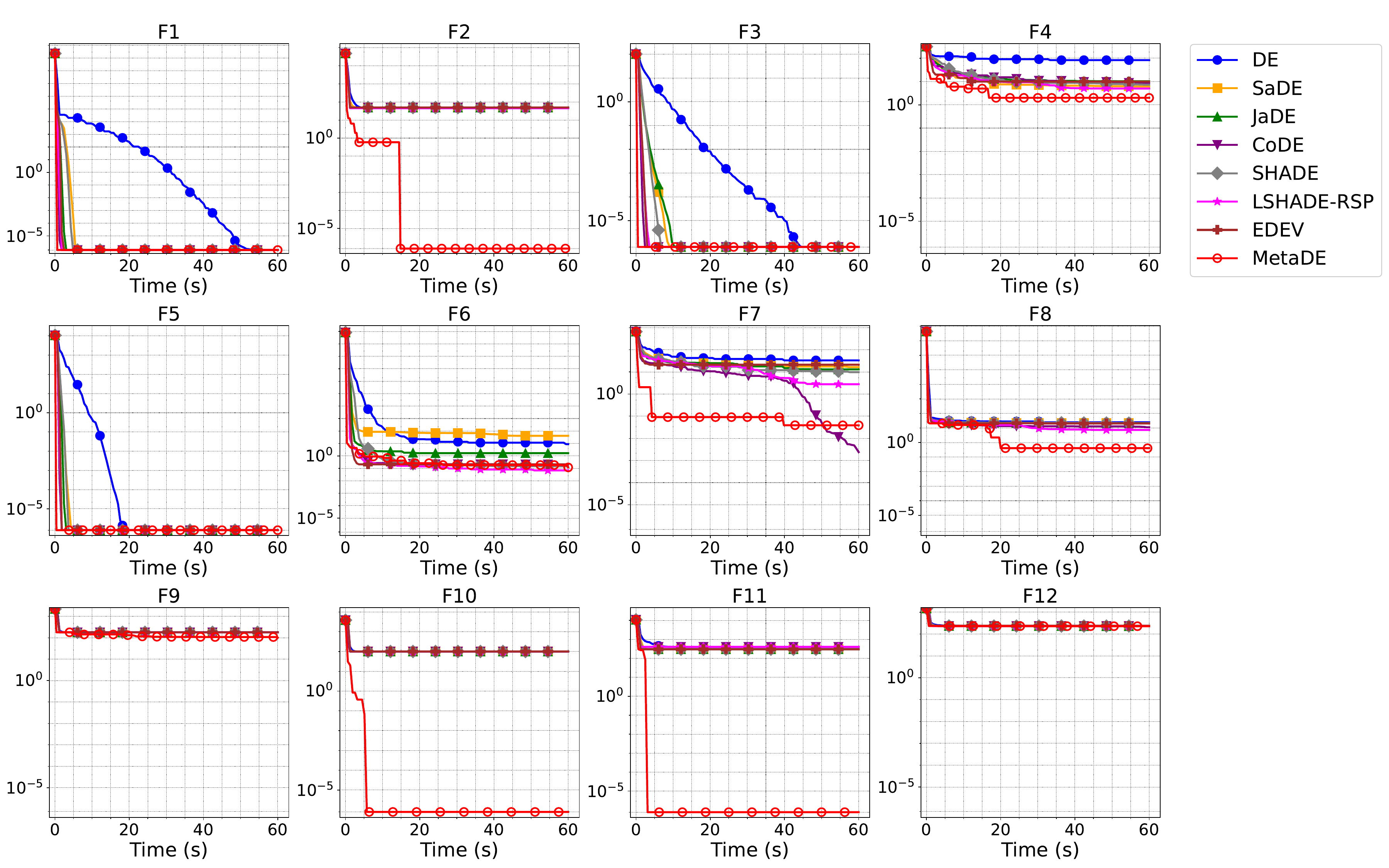}
\caption{Convergence curves on 20D problems in CEC2022 benchmark suite. The peer DE variants are set with population size of 1,000.}
\label{Figure_convergence_20D_NP10k_supp}
\end{figure*}

\clearpage

\subsection{Detailed Experimental Results}\label{section:FEs_supp}

% Tables \ref{tab:vsClass10D} and \ref{tab:vsClass20D} present the detailed results of MetaDE compared to other algorithms on 10-dimensional and 20-dimensional problems from CEC2022 within a 60-second time frame. The convergence curves for all problems are shown in Figs. \ref{Figure_convergence_10D} and \ref{Figure_convergence_20D}.

% Tables \ref{tab:NP10000 10D} and \ref{tab:NP10000 20D} display the detailed results for MetaDE versus comparison algorithms with equal concurrency (population size = 10,000) on 10-dimensional and 20-dimensional problems in CEC2022, all within a span of 60 seconds. The convergence curves for all problems are shown in Figs. \ref{Figure_convergence_10D_NP10k} and \ref{Figure_convergence_20D_NP10k}.

\begin{table}[htbp]
  \centering
  \caption{
  Detailed results on 10D problems in CEC2022 benchmark suite. The peer DE variants are set with population size of 100.
The mean and standard deviation (in parentheses) of the results over multiple runs are displayed in pairs. 
Results with the best mean values are highlighted.
  }
  \resizebox{\textwidth}{!}{
   \renewcommand{\arraystretch}{1.2}
% Table generated by Excel2LaTeX from sheet 'Experiment1 60S'
\begin{tabular}{ccccccccc}
\toprule
Func  & MetaDE & DE    & SaDE  & JaDE  & CoDE  & SHADE & LSHADE-RSP & EDEV \\
\midrule
$F_{1}$ & \textbf{0.00E+00 (0.00E+00)} & \boldmath{}\textbf{0.00E+00 (0.00E+00)$\approx$}\unboldmath{} & \boldmath{}\textbf{0.00E+00 (0.00E+00)$\approx$}\unboldmath{} & \boldmath{}\textbf{0.00E+00 (0.00E+00)$\approx$}\unboldmath{} & \boldmath{}\textbf{0.00E+00 (0.00E+00)$\approx$}\unboldmath{} & \boldmath{}\textbf{0.00E+00 (0.00E+00)$\approx$}\unboldmath{} & \boldmath{}\textbf{0.00E+00 (0.00E+00)$\approx$}\unboldmath{} & \boldmath{}\textbf{0.00E+00 (0.00E+00)$\approx$}\unboldmath{} \\
$F_{2}$ & \textbf{0.00E+00 (0.00E+00)} & 6.05E+00 (2.43E+00)$-$ & 4.86E+00 (4.29E+00)$-$ & 4.89E+00 (3.74E+00)$-$ & 4.71E+00 (2.55E+00)$-$ & 5.47E+00 (3.62E+00)$-$ & 2.38E+00 (2.58E+00)$-$ & 6.11E+00 (2.87E+00)$-$ \\
$F_{3}$ & \textbf{0.00E+00 (0.00E+00)} & \boldmath{}\textbf{0.00E+00 (0.00E+00)$\approx$}\unboldmath{} & \boldmath{}\textbf{0.00E+00 (0.00E+00)$\approx$}\unboldmath{} & \boldmath{}\textbf{0.00E+00 (0.00E+00)$\approx$}\unboldmath{} & \boldmath{}\textbf{0.00E+00 (0.00E+00)$\approx$}\unboldmath{} & \boldmath{}\textbf{0.00E+00 (0.00E+00)$\approx$}\unboldmath{} & \boldmath{}\textbf{0.00E+00 (0.00E+00)$\approx$}\unboldmath{} & \boldmath{}\textbf{0.00E+00 (0.00E+00)$\approx$}\unboldmath{} \\
$F_{4}$ & \textbf{0.00E+00 (0.00E+00)} & 6.90E+00 (4.01E+00)$-$ & 1.03E+00 (8.93E-01)$-$ & 2.31E+01 (1.19E+01)$-$ & 8.34E-01 (7.62E-01)$-$ & 3.05E+00 (9.43E-01)$-$ & 2.12E+00 (6.56E-01)$-$ & 6.52E+00 (4.51E+00)$-$ \\
$F_{5}$ & \textbf{0.00E+00 (0.00E+00)} & \boldmath{}\textbf{0.00E+00 (0.00E+00)$\approx$}\unboldmath{} & \boldmath{}\textbf{0.00E+00 (0.00E+00)$\approx$}\unboldmath{} & \boldmath{}\textbf{0.00E+00 (0.00E+00)$\approx$}\unboldmath{} & \boldmath{}\textbf{0.00E+00 (0.00E+00)$\approx$}\unboldmath{} & \boldmath{}\textbf{0.00E+00 (0.00E+00)$\approx$}\unboldmath{} & \boldmath{}\textbf{0.00E+00 (0.00E+00)$\approx$}\unboldmath{} & \boldmath{}\textbf{0.00E+00 (0.00E+00)$\approx$}\unboldmath{} \\
$F_{6}$ & \textbf{5.50E-04 (3.96E-04)} & 1.11E-01 (8.92E-02)$-$ & 6.04E+01 (2.25E+02)$-$ & 1.60E+00 (2.48E+00)$-$ & 9.08E-03 (1.17E-02)$-$ & 1.33E+00 (2.22E+00)$-$ & 3.84E-02 (5.60E-02)$-$ & 9.06E-01 (1.76E+00)$-$ \\
$F_{7}$ & \textbf{0.00E+00 (0.00E+00)} & 5.18E-02 (1.59E-01)$-$ & 2.15E-02 (1.39E-02)$-$ & \boldmath{}\textbf{0.00E+00 (0.00E+00)$\approx$}\unboldmath{} & 1.92E-03 (7.32E-03)$-$ & 6.59E-03 (1.41E-02)$-$ & 1.06E+02 (1.42E-02)$-$ & 9.54E+00 (9.86E+00)$-$ \\
$F_{8}$ & \textbf{5.52E-03 (4.41E-03)} & 1.42E-01 (2.45E-01)$-$ & 5.15E-02 (2.29E-02)$-$ & 1.74E+01 (4.79E+00)$-$ & 6.02E-03 (1.02E-02)$-$ & 2.29E+00 (5.89E+00)$-$ & 2.19E+00 (5.89E+00)$-$ & 7.06E+00 (9.36E+00)$-$ \\
$F_{9}$ & \textbf{3.36E+00 (1.77E+01)} & 2.29E+02 (7.53E-06)$-$ & 2.29E+02 (6.38E-06)$-$ & 2.29E+02 (6.38E-06)$-$ & 2.29E+02 (7.17E-06)$-$ & 2.29E+02 (7.43E-06)$-$ & 2.29E+02 (8.19E-05)$-$ & 2.29E+02 (1.01E-05)$-$ \\
$F_{10}$ & \textbf{0.00E+00 (0.00E+00)} & 1.00E+02 (5.18E-02)$-$ & 1.03E+02 (1.83E+01)$-$ & 1.04E+02 (1.91E+01)$-$ & 1.00E+02 (6.83E-02)$-$ & 1.10E+02 (3.09E+01)$-$ & 1.03E+02 (1.77E+01)$-$ & 1.07E+02 (2.62E+01)$-$ \\
$F_{11}$ & \textbf{0.00E+00 (0.00E+00)} & \boldmath{}\textbf{0.00E+00 (0.00E+00)$\approx$}\unboldmath{} & 2.42E+01 (5.52E+01)$-$ & \boldmath{}\textbf{0.00E+00 (0.00E+00)$\approx$}\unboldmath{} & \boldmath{}\textbf{0.00E+00 (0.00E+00)$\approx$}\unboldmath{} & \boldmath{}\textbf{0.00E+00 (0.00E+00)$\approx$}\unboldmath{} & \boldmath{}\textbf{0.00E+00 (0.00E+00)$\approx$}\unboldmath{} & 4.84E+00 (2.65E+01)$-$ \\
$F_{12}$ & \textbf{1.39E+02 (4.63E+01)} & 1.62E+02 (1.04E+00)$-$ & 1.63E+02 (1.57E+00)$-$ & 1.62E+02 (2.22E+00)$-$ & 1.59E+02 (1.14E+00)$-$ & 1.63E+02 (1.25E+00)$-$ & 1.64E+02 (1.38E+00)$-$ & 1.62E+02 (1.71E+00)$-$ \\
\midrule
$+$ / $\approx$ / $-$ & --    & 0/4/8 & 0/3/9 & 0/5/7 & 0/5/7 & 0/4/8 & 0/4/8 & 0/3/9 \\
\bottomrule
\end{tabular}%
}
\footnotesize
\textsuperscript{*} The Wilcoxon rank-sum tests (with a significance level of 0.05) were conducted between MetaDE and each individually.
The final row displays the number of problems where the corresponding algorithm performs statistically better ($+$),  similar ($\thickapprox$), or worse ($-$) compared to MetaDE.\\
\label{tab:vsClass10D_supp}%

\end{table}%

\begin{table}[htbp]
  \centering
  \caption{Detailed results on 20D problems in CEC2022 benchmark suite. The peer DE variants are set with population size of 100.
  The mean and standard deviation (in parentheses) of the results over multiple runs are displayed in pairs. 
Results with the best mean values are highlighted.
  }
  {
  \resizebox{\textwidth}{!}{
   \renewcommand{\arraystretch}{1.2}
% Table generated by Excel2LaTeX from sheet 'Experiment1 60S'
\begin{tabular}{ccccccccc}
\toprule
Func  & MetaDE & DE    & SaDE  & JaDE  & CoDE  & SHADE & LSHADE-RSP & EDEV \\
\midrule
$F_{1}$ & \multicolumn{1}{l}{\textbf{0.00E+00 (0.00E+00)}} & \multicolumn{1}{l}{\boldmath{}\textbf{0.00E+00 (0.00E+00)$\approx$}\unboldmath{}} & \multicolumn{1}{l}{\boldmath{}\textbf{0.00E+00 (0.00E+00)$\approx$}\unboldmath{}} & \multicolumn{1}{l}{\boldmath{}\textbf{0.00E+00 (0.00E+00)$\approx$}\unboldmath{}} & \multicolumn{1}{l}{\boldmath{}\textbf{0.00E+00 (0.00E+00)$\approx$}\unboldmath{}} & \multicolumn{1}{l}{\boldmath{}\textbf{0.00E+00 (0.00E+00)$\approx$}\unboldmath{}} & \multicolumn{1}{l}{\boldmath{}\textbf{0.00E+00 (0.00E+00)$\approx$}\unboldmath{}} & \multicolumn{1}{l}{\boldmath{}\textbf{0.00E+00 (0.00E+00)$\approx$}\unboldmath{}} \\
$F_{2}$ & \multicolumn{1}{l}{\textbf{1.26E-02 (3.74E-02)}} & \multicolumn{1}{l}{4.69E+01 (2.09E+00)$-$} & \multicolumn{1}{l}{3.50E+01 (2.21E+01)$-$} & \multicolumn{1}{l}{4.75E+01 (8.67E+00)$-$} & \multicolumn{1}{l}{4.58E+01 (1.20E+01)$-$} & \multicolumn{1}{l}{4.75E+01 (8.67E+00)$-$} & \multicolumn{1}{l}{4.30E+01 (1.71E+01)$-$} & \multicolumn{1}{l}{4.13E+01 (1.78E+01)$-$} \\
$F_{3}$ & \multicolumn{1}{l}{\textbf{0.00E+00 (0.00E+00)}} & \multicolumn{1}{l}{\boldmath{}\textbf{0.00E+00 (0.00E+00)$\approx$}\unboldmath{}} & \multicolumn{1}{l}{\boldmath{}\textbf{0.00E+00 (0.00E+00)$\approx$}\unboldmath{}} & \multicolumn{1}{l}{\boldmath{}\textbf{0.00E+00 (0.00E+00)$\approx$}\unboldmath{}} & \multicolumn{1}{l}{\boldmath{}\textbf{0.00E+00 (0.00E+00)$\approx$}\unboldmath{}} & \multicolumn{1}{l}{1.03E-08 (5.64E-08)$\approx$} & \multicolumn{1}{l}{\boldmath{}\textbf{0.00E+00 (0.00E+00)$\approx$}\unboldmath{}} & \multicolumn{1}{l}{6.91E-05 (3.31E-04)$-$} \\
$F_{4}$ & \multicolumn{1}{l}{\textbf{2.02E+00 (8.56E-01)}} & \multicolumn{1}{l}{1.95E+01 (8.17E+00)$-$} & \multicolumn{1}{l}{7.42E+00 (2.14E+00)$-$} & \multicolumn{1}{l}{7.21E+01 (3.45E+01)$-$} & \multicolumn{1}{l}{1.11E+01 (2.22E+00)$-$} & \multicolumn{1}{l}{1.27E+01 (2.73E+00)$-$} & \multicolumn{1}{l}{9.05E+00 (1.44E+00)$-$} & \multicolumn{1}{l}{2.37E+01 (1.37E+01)$-$} \\
$F_{5}$ & \multicolumn{1}{l}{\textbf{0.00E+00 (0.00E+00)}} & \multicolumn{1}{l}{\boldmath{}\textbf{0.00E+00 (0.00E+00)$\approx$}\unboldmath{}} & \multicolumn{1}{l}{7.24E-01 (1.22E+00)$-$} & \multicolumn{1}{l}{\boldmath{}\textbf{0.00E+00 (0.00E+00)$\approx$}\unboldmath{}} & \multicolumn{1}{l}{\boldmath{}\textbf{0.00E+00 (0.00E+00)$\approx$}\unboldmath{}} & \multicolumn{1}{l}{\boldmath{}\textbf{0.00E+00 (0.00E+00)$\approx$}\unboldmath{}} & \multicolumn{1}{l}{\boldmath{}\textbf{0.00E+00 (0.00E+00)$\approx$}\unboldmath{}} & \multicolumn{1}{l}{1.78E-01 (3.12E-01)$-$} \\
$F_{6}$ & \multicolumn{1}{l}{\textbf{1.16E-01 (2.79E-02)}} & \multicolumn{1}{l}{4.99E-01 (4.11E-01)$-$} & \multicolumn{1}{l}{3.13E+01 (1.54E+01)$-$} & \multicolumn{1}{l}{5.34E+01 (3.33E+01)$-$} & \multicolumn{1}{l}{1.89E+01 (1.83E+01)$-$} & \multicolumn{1}{l}{5.06E+01 (3.16E+01)$-$} & \multicolumn{1}{l}{1.25E+01 (1.00E+01)$-$} & \multicolumn{1}{l}{4.91E+03 (6.53E+03)$-$} \\
$F_{7}$ & \multicolumn{1}{l}{\textbf{5.17E-02 (6.21E-02)}} & \multicolumn{1}{l}{4.23E+00 (7.90E+00)$-$} & \multicolumn{1}{l}{1.07E+01 (5.20E+00)$-$} & \multicolumn{1}{l}{2.98E+00 (3.61E+00)$-$} & \multicolumn{1}{l}{1.16E+00 (1.26E+00)$-$} & \multicolumn{1}{l}{7.77E+00 (6.65E+00)$-$} & \multicolumn{1}{l}{1.42E+01 (8.96E+00)$-$} & \multicolumn{1}{l}{2.29E+01 (8.80E+00)$-$} \\
$F_{8}$ & \multicolumn{1}{l}{\textbf{7.19E-01 (1.02E+00)}} & \multicolumn{1}{l}{8.24E+00 (1.00E+01)$-$} & \multicolumn{1}{l}{2.10E+01 (7.26E-01)$-$} & \multicolumn{1}{l}{2.64E+01 (9.73E-01)$-$} & \multicolumn{1}{l}{1.38E+01 (8.98E+00)$-$} & \multicolumn{1}{l}{2.02E+01 (8.29E-01)$-$} & \multicolumn{1}{l}{1.96E+01 (3.80E+00)$-$} & \multicolumn{1}{l}{2.08E+01 (3.81E-01)$-$} \\
$F_{9}$ & \multicolumn{1}{l}{\textbf{1.07E+02 (1.98E+01)}} & \multicolumn{1}{l}{1.81E+02 (9.39E-06)$-$} & \multicolumn{1}{l}{1.81E+02 (3.75E-06)$-$} & \multicolumn{1}{l}{1.81E+02 (9.02E-06)$-$} & \multicolumn{1}{l}{1.81E+02 (8.41E-06)$-$} & \multicolumn{1}{l}{1.81E+02 (1.04E-05)$-$} & \multicolumn{1}{l}{1.81E+02 (2.18E-05)$-$} & \multicolumn{1}{l}{1.81E+02 (5.43E-04)$-$} \\
$F_{10}$ & \multicolumn{1}{l}{\textbf{0.00E+00 (0.00E+00)}} & \multicolumn{1}{l}{1.13E+02 (3.52E+01)$-$} & \multicolumn{1}{l}{1.00E+02 (3.03E-02)$-$} & \multicolumn{1}{l}{1.13E+02 (3.76E+01)$-$} & \multicolumn{1}{l}{1.00E+02 (3.55E-02)$-$} & \multicolumn{1}{l}{1.12E+02 (3.47E+01)$-$} & \multicolumn{1}{l}{1.11E+02 (3.42E+01)$-$} & \multicolumn{1}{l}{1.07E+02 (3.80E+01)$-$} \\
$F_{11}$ & \multicolumn{1}{l}{\textbf{7.28E-05 (3.06E-04)}} & \multicolumn{1}{l}{3.39E+02 (4.87E+01)$-$} & \multicolumn{1}{l}{3.06E+02 (2.46E+01)$-$} & \multicolumn{1}{l}{3.19E+02 (3.95E+01)$-$} & \multicolumn{1}{l}{3.39E+02 (4.87E+01)$-$} & \multicolumn{1}{l}{3.16E+02 (3.68E+01)$-$} & \multicolumn{1}{l}{3.39E+02 (4.87E+01)$-$} & \multicolumn{1}{l}{3.19E+02 (3.95E+01)$-$} \\
$F_{12}$ & \multicolumn{1}{l}{\textbf{2.29E+02 (6.08E-01)}} & \multicolumn{1}{l}{2.37E+02 (3.11E+00)$-$} & \multicolumn{1}{l}{2.41E+02 (5.26E+00)$-$} & \multicolumn{1}{l}{2.37E+02 (5.10E+00)$-$} & \multicolumn{1}{l}{2.34E+02 (2.68E+00)$-$} & \multicolumn{1}{l}{2.39E+02 (4.46E+00)$-$} & \multicolumn{1}{l}{2.44E+02 (1.63E+01)$-$} & \multicolumn{1}{l}{2.42E+02 (8.38E+00)$-$} \\
\midrule
$+$ / $\approx$ / $-$ & --    & 0/3/9 & 0/2/10 & 0/3/9 & 0/3/9 & 0/3/9 & 0/3/9 & 0/1/11 \\
\bottomrule
\end{tabular}%
    }
\footnotesize
\textsuperscript{*} The Wilcoxon rank-sum tests (with a significance level of 0.05) were conducted between MetaDE and each individually.
The final row displays the number of problems where the corresponding algorithm performs statistically better ($+$),  similar ($\thickapprox$), or worse ($-$) compared to MetaDE.\\
\label{tab:vsClass20D_supp}%
}
\end{table}%

\clearpage

\begin{table}[htbp]
  \centering
  \caption{Detailed results on 10D problems in CEC2022 benchmark suite. The peer DE variants are set with population size of 1,000. 
The mean and standard deviation (in parentheses) of the results over multiple runs are displayed in pairs. 
Results with the best mean values are highlighted.
  }
  {
  \resizebox{\textwidth}{!}{
   \renewcommand{\arraystretch}{1.2}
% Table generated by Excel2LaTeX from sheet 'Exp2 NP1000'
\begin{tabular}{ccccccccc}
\toprule
Func  & MetaDE & DE    & SaDE  & JaDE  & CoDE  & SHADE & LSHADE-RSP & EDEV \\
\midrule
$F_{1}$ & \multicolumn{1}{l}{\textbf{0.00E+00 (0.00E+00)}} & \multicolumn{1}{l}{\boldmath{}\textbf{0.00E+00 (0.00E+00)$\approx$}\unboldmath{}} & \multicolumn{1}{l}{\boldmath{}\textbf{0.00E+00 (0.00E+00)$\approx$}\unboldmath{}} & \multicolumn{1}{l}{\boldmath{}\textbf{0.00E+00 (0.00E+00)$\approx$}\unboldmath{}} & \multicolumn{1}{l}{\boldmath{}\textbf{0.00E+00 (0.00E+00)$\approx$}\unboldmath{}} & \multicolumn{1}{l}{\boldmath{}\textbf{0.00E+00 (0.00E+00)$\approx$}\unboldmath{}} & \multicolumn{1}{l}{\boldmath{}\textbf{0.00E+00 (0.00E+00)$\approx$}\unboldmath{}} & \multicolumn{1}{l}{\boldmath{}\textbf{0.00E+00 (0.00E+00)$\approx$}\unboldmath{}} \\
$F_{2}$ & \multicolumn{1}{l}{\textbf{0.00E+00 (0.00E+00)}} & \multicolumn{1}{l}{3.60E+00 (1.20E+00)$-$} & \multicolumn{1}{l}{6.87E+00 (3.63E+00)$-$} & \multicolumn{1}{l}{8.15E+00 (2.11E+00)$-$} & \multicolumn{1}{l}{2.44E+00 (1.97E+00)$-$} & \multicolumn{1}{l}{8.02E+00 (2.47E+00)$-$} & \multicolumn{1}{l}{1.12E+00 (1.79E+00)$-$} & \multicolumn{1}{l}{6.27E+00 (2.93E+00)$-$} \\
$F_{3}$ & \multicolumn{1}{l}{\textbf{0.00E+00 (0.00E+00)}} & \multicolumn{1}{l}{\boldmath{}\textbf{0.00E+00 (0.00E+00)$\approx$}\unboldmath{}} & \multicolumn{1}{l}{\boldmath{}\textbf{0.00E+00 (0.00E+00)$\approx$}\unboldmath{}} & \multicolumn{1}{l}{\boldmath{}\textbf{0.00E+00 (0.00E+00)$\approx$}\unboldmath{}} & \multicolumn{1}{l}{\boldmath{}\textbf{0.00E+00 (0.00E+00)$\approx$}\unboldmath{}} & \multicolumn{1}{l}{\boldmath{}\textbf{0.00E+00 (0.00E+00)$\approx$}\unboldmath{}} & \multicolumn{1}{l}{\boldmath{}\textbf{0.00E+00 (0.00E+00)$\approx$}\unboldmath{}} & \multicolumn{1}{l}{\boldmath{}\textbf{0.00E+00 (0.00E+00)$\approx$}\unboldmath{}} \\
$F_{4}$ & \multicolumn{1}{l}{\textbf{0.00E+00 (0.00E+00)}} & \multicolumn{1}{l}{1.50E+01 (2.59E+00)$-$} & \multicolumn{1}{l}{4.75E-01 (5.04E-01)$-$} & \multicolumn{1}{l}{2.20E+00 (5.30E-01)$-$} & \multicolumn{1}{l}{\boldmath{}\textbf{0.00E+00 (0.00E+00)$\approx$}\unboldmath{}} & \multicolumn{1}{l}{\boldmath{}\textbf{0.00E+00 (0.00E+00)$\approx$}\unboldmath{}} & \multicolumn{1}{l}{9.63E-01 (6.54E-01)$-$} & \multicolumn{1}{l}{4.91E+00 (5.24E+00)$-$} \\
$F_{5}$ & \multicolumn{1}{l}{\textbf{0.00E+00 (0.00E+00)}} & \multicolumn{1}{l}{\boldmath{}\textbf{0.00E+00 (0.00E+00)$\approx$}\unboldmath{}} & \multicolumn{1}{l}{\boldmath{}\textbf{0.00E+00 (0.00E+00)$\approx$}\unboldmath{}} & \multicolumn{1}{l}{\boldmath{}\textbf{0.00E+00 (0.00E+00)$\approx$}\unboldmath{}} & \multicolumn{1}{l}{\boldmath{}\textbf{0.00E+00 (0.00E+00)$\approx$}\unboldmath{}} & \multicolumn{1}{l}{\boldmath{}\textbf{0.00E+00 (0.00E+00)$\approx$}\unboldmath{}} & \multicolumn{1}{l}{\boldmath{}\textbf{0.00E+00 (0.00E+00)$\approx$}\unboldmath{}} & \multicolumn{1}{l}{\boldmath{}\textbf{0.00E+00 (0.00E+00)$\approx$}\unboldmath{}} \\
$F_{6}$ & \multicolumn{1}{l}{\textbf{5.50E-04 (3.96E-04)}} & \multicolumn{1}{l}{1.49E-02 (4.74E-03)$-$} & \multicolumn{1}{l}{1.95E+00 (1.54E+00)$-$} & \multicolumn{1}{l}{1.01E-01 (6.20E-02)$-$} & \multicolumn{1}{l}{7.69E-04 (4.31E-04)$\approx$} & \multicolumn{1}{l}{4.62E-03 (8.17E-03)$-$} & \multicolumn{1}{l}{1.89E-03 (7.18E-04)$-$} & \multicolumn{1}{l}{4.37E-02 (6.48E-02)$-$} \\
$F_{7}$ & \multicolumn{1}{l}{\textbf{0.00E+00 (0.00E+00)}} & \multicolumn{1}{l}{9.60E-04 (5.35E-03)$-$} & \multicolumn{1}{l}{1.98E-02 (8.61E-03)$-$} & \multicolumn{1}{l}{\boldmath{}\textbf{0.00E+00 (0.00E+00)$\approx$}\unboldmath{}} & \multicolumn{1}{l}{\boldmath{}\textbf{0.00E+00 (0.00E+00)$\approx$}\unboldmath{}} & \multicolumn{1}{l}{\boldmath{}\textbf{0.00E+00 (0.00E+00)$\approx$}\unboldmath{}} & \multicolumn{1}{l}{\boldmath{}\textbf{0.00E+00 (0.00E+00)$\approx$}\unboldmath{}} & \multicolumn{1}{l}{1.35E+00 (4.98E+00)$-$} \\
$F_{8}$ & \multicolumn{1}{l}{5.52E-03 (4.41E-03)} & \multicolumn{1}{l}{1.19E-01 (2.50E-02)$-$} & \multicolumn{1}{l}{1.13E+00 (4.44E-01)$-$} & \multicolumn{1}{l}{1.03E-01 (2.62E-02)$-$} & \multicolumn{1}{l}{\boldmath{}\textbf{0.00E+00 (0.00E+00)$\approx$}\unboldmath{}} & \multicolumn{1}{l}{8.28E-02 (2.66E-02)$-$} & \multicolumn{1}{l}{9.13E-02 (1.11E-01)$-$} & \multicolumn{1}{l}{1.03E+00 (3.58E+00)$-$} \\
$F_{9}$ & \multicolumn{1}{l}{\textbf{3.36E+00 (1.77E+01)}} & \multicolumn{1}{l}{2.29E+02 (7.85E-06)$-$} & \multicolumn{1}{l}{2.29E+02 (8.58E-06)$-$} & \multicolumn{1}{l}{2.29E+02 (8.28E-06)$-$} & \multicolumn{1}{l}{2.29E+02 (8.67E-14)$-$} & \multicolumn{1}{l}{2.29E+02 (2.74E-06)$-$} & \multicolumn{1}{l}{2.29E+02 (9.39E-06)$-$} & \multicolumn{1}{l}{2.29E+02 (1.17E-05)$-$} \\
$F_{10}$ & \multicolumn{1}{l}{\textbf{0.00E+00 (0.00E+00)}} & \multicolumn{1}{l}{1.00E+02 (1.70E-02)$-$} & \multicolumn{1}{l}{1.00E+02 (3.93E-02)$-$} & \multicolumn{1}{l}{1.00E+02 (2.49E-02)$-$} & \multicolumn{1}{l}{1.00E+02 (1.01E-02)$-$} & \multicolumn{1}{l}{1.00E+02 (1.83E-02)$-$} & \multicolumn{1}{l}{1.00E+02 (8.05E-04)$-$} & \multicolumn{1}{l}{1.00E+02 (3.93E-02)$-$} \\
$F_{11}$ & \multicolumn{1}{l}{\textbf{0.00E+00 (0.00E+00)}} & \multicolumn{1}{l}{\boldmath{}\textbf{0.00E+00 (0.00E+00)$\approx$}\unboldmath{}} & \multicolumn{1}{l}{\boldmath{}\textbf{0.00E+00 (0.00E+00)$\approx$}\unboldmath{}} & \multicolumn{1}{l}{\boldmath{}\textbf{0.00E+00 (0.00E+00)$\approx$}\unboldmath{}} & \multicolumn{1}{l}{3.65E-07 (2.03E-06)$-$} & \multicolumn{1}{l}{\boldmath{}\textbf{0.00E+00 (0.00E+00)$\approx$}\unboldmath{}} & \multicolumn{1}{l}{\boldmath{}\textbf{0.00E+00 (0.00E+00)$\approx$}\unboldmath{}} & \multicolumn{1}{l}{\boldmath{}\textbf{0.00E+00 (0.00E+00)$\approx$}\unboldmath{}} \\
$F_{12}$ & \multicolumn{1}{l}{\textbf{1.39E+02 (4.63E+01)}} & \multicolumn{1}{l}{1.60E+02 (9.79E-01)$-$} & \multicolumn{1}{l}{1.60E+02 (1.56E+00)$-$} & \multicolumn{1}{l}{1.59E+02 (1.28E+00)$-$} & \multicolumn{1}{l}{1.59E+02 (8.67E-14)$-$} & \multicolumn{1}{l}{1.61E+02 (1.69E+00)$-$} & \multicolumn{1}{l}{1.63E+02 (7.62E-01)$-$} & \multicolumn{1}{l}{1.60E+02 (1.18E+00)$-$} \\
\midrule
$+$ / $\approx$ / $-$ & --    & 0/4/8 & 0/4/8 & 0/5/7 & 0/7/5 & 0/6/6 & 0/5/7 & 0/4/8 \\
\bottomrule
\end{tabular}%
}
\footnotesize
\textsuperscript{*} The Wilcoxon rank-sum tests (with a significance level of 0.05) were conducted between MetaDE and each individually.
The final row displays the number of problems where the corresponding algorithm performs statistically better ($+$),  similar ($\thickapprox$), or worse ($-$) compared to MetaDE.\\
\label{tab:NP10000 10D_supp}%
}
\end{table}%

\begin{table}[htbp]
  \centering
  \caption{Detailed results on 20D problems in CEC2022 benchmark suite. The peer DE variants are set with population size of 1,000. 
The mean and standard deviation (in parentheses) of the results over multiple runs are displayed in pairs. 
Results with the best mean values are highlighted.
  }
  {
  \resizebox{\textwidth}{!}{
   \renewcommand{\arraystretch}{1.2}
% Table generated by Excel2LaTeX from sheet 'Exp2 NP1000'
\begin{tabular}{ccccccccc}
\toprule
Func  & MetaDE & DE    & SaDE  & JaDE  & CoDE  & SHADE & LSHADE-RSP & EDEV \\
\midrule
$F_{1}$ & \multicolumn{1}{l}{\textbf{0.00E+00 (0.00E+00)}} & \multicolumn{1}{l}{\boldmath{}\textbf{0.00E+00 (0.00E+00)$\approx$}\unboldmath{}} & \multicolumn{1}{l}{\boldmath{}\textbf{0.00E+00 (0.00E+00)$\approx$}\unboldmath{}} & \multicolumn{1}{l}{\boldmath{}\textbf{0.00E+00 (0.00E+00)$\approx$}\unboldmath{}} & \multicolumn{1}{l}{\boldmath{}\textbf{0.00E+00 (0.00E+00)$\approx$}\unboldmath{}} & \multicolumn{1}{l}{\boldmath{}\textbf{0.00E+00 (0.00E+00)$\approx$}\unboldmath{}} & \multicolumn{1}{l}{\boldmath{}\textbf{0.00E+00 (0.00E+00)$\approx$}\unboldmath{}} & \multicolumn{1}{l}{\boldmath{}\textbf{0.00E+00 (0.00E+00)$\approx$}\unboldmath{}} \\
$F_{2}$ & \multicolumn{1}{l}{\textbf{1.26E-02 (3.74E-02)}} & \multicolumn{1}{l}{4.49E+01 (0.00E+00)$-$} & \multicolumn{1}{l}{4.91E+01 (7.67E-06)$-$} & \multicolumn{1}{l}{4.91E+01 (0.00E+00)$-$} & \multicolumn{1}{l}{4.91E+01 (0.00E+00)$-$} & \multicolumn{1}{l}{4.91E+01 (0.00E+00)$-$} & \multicolumn{1}{l}{4.52E+01 (1.05E+00)$-$} & \multicolumn{1}{l}{4.89E+01 (1.05E+00)$-$} \\
$F_{3}$ & \multicolumn{1}{l}{\textbf{0.00E+00 (0.00E+00)}} & \multicolumn{1}{l}{\boldmath{}\textbf{0.00E+00 (0.00E+00)$\approx$}\unboldmath{}} & \multicolumn{1}{l}{\boldmath{}\textbf{0.00E+00 (0.00E+00)$\approx$}\unboldmath{}} & \multicolumn{1}{l}{\boldmath{}\textbf{0.00E+00 (0.00E+00)$\approx$}\unboldmath{}} & \multicolumn{1}{l}{\boldmath{}\textbf{0.00E+00 (0.00E+00)$\approx$}\unboldmath{}} & \multicolumn{1}{l}{\boldmath{}\textbf{0.00E+00 (0.00E+00)$\approx$}\unboldmath{}} & \multicolumn{1}{l}{\boldmath{}\textbf{0.00E+00 (0.00E+00)$\approx$}\unboldmath{}} & \multicolumn{1}{l}{\boldmath{}\textbf{0.00E+00 (0.00E+00)$\approx$}\unboldmath{}} \\
$F_{4}$ & \multicolumn{1}{l}{\textbf{2.02E+00 (8.56E-01)}} & \multicolumn{1}{l}{8.33E+01 (5.85E+00)$-$} & \multicolumn{1}{l}{6.09E+00 (1.02E+00)$-$} & \multicolumn{1}{l}{1.07E+01 (1.95E+00)$-$} & \multicolumn{1}{l}{8.02E+00 (1.08E+00)$-$} & \multicolumn{1}{l}{7.39E+00 (1.17E+00)$-$} & \multicolumn{1}{l}{4.82E+00 (8.64E-01)$-$} & \multicolumn{1}{l}{1.81E+01 (1.52E+01)$-$} \\
$F_{5}$ & \multicolumn{1}{l}{\textbf{0.00E+00 (0.00E+00)}} & \multicolumn{1}{l}{\boldmath{}\textbf{0.00E+00 (0.00E+00)$\approx$}\unboldmath{}} & \multicolumn{1}{l}{6.16E-02 (2.71E-01)$-$} & \multicolumn{1}{l}{\boldmath{}\textbf{0.00E+00 (0.00E+00)$\approx$}\unboldmath{}} & \multicolumn{1}{l}{\boldmath{}\textbf{0.00E+00 (0.00E+00)$\approx$}\unboldmath{}} & \multicolumn{1}{l}{\boldmath{}\textbf{0.00E+00 (0.00E+00)$\approx$}\unboldmath{}} & \multicolumn{1}{l}{\boldmath{}\textbf{0.00E+00 (0.00E+00)$\approx$}\unboldmath{}} & \multicolumn{1}{l}{\boldmath{}\textbf{0.00E+00 (0.00E+00)$\approx$}\unboldmath{}} \\
$F_{6}$ & \multicolumn{1}{l}{1.16E-01 (2.79E-02)} & \multicolumn{1}{l}{8.86E+00 (1.34E+00)$-$} & \multicolumn{1}{l}{1.72E+02 (5.24E+02)$-$} & \multicolumn{1}{l}{1.80E+00 (7.49E-01)$-$} & \multicolumn{1}{l}{2.07E-01 (5.04E-02)$-$} & \multicolumn{1}{l}{1.35E-01 (5.23E-02)$\approx$} & \multicolumn{1}{l}{\textbf{7.35E-02 (3.33E-02)$+$}} & \multicolumn{1}{l}{2.70E+00 (1.13E+01)$-$} \\
$F_{7}$ & \multicolumn{1}{l}{5.17E-02 (6.21E-02)} & \multicolumn{1}{l}{3.28E+01 (1.82E+00)$-$} & \multicolumn{1}{l}{1.57E+01 (3.45E+00)$-$} & \multicolumn{1}{l}{1.31E+01 (2.39E+00)$-$} & \multicolumn{1}{l}{\textbf{2.02E-03 (1.44E-03)$+$}} & \multicolumn{1}{l}{9.48E+00 (1.76E+00)$-$} & \multicolumn{1}{l}{2.92E+00 (9.57E-01)$-$} & \multicolumn{1}{l}{1.68E+01 (1.04E+01)$-$} \\
$F_{8}$ & \multicolumn{1}{l}{\textbf{7.19E-01 (1.02E+00)}} & \multicolumn{1}{l}{2.41E+01 (2.25E+00)$-$} & \multicolumn{1}{l}{2.11E+01 (1.76E+00)$-$} & \multicolumn{1}{l}{2.15E+01 (1.27E+00)$-$} & \multicolumn{1}{l}{1.12E+01 (1.98E+00)$-$} & \multicolumn{1}{l}{1.99E+01 (2.81E+00)$-$} & \multicolumn{1}{l}{8.53E+00 (4.47E+00)$-$} & \multicolumn{1}{l}{1.88E+01 (5.20E+00)$-$} \\
$F_{9}$ & \multicolumn{1}{l}{\textbf{1.07E+02 (1.98E+01)}} & \multicolumn{1}{l}{1.81E+02 (2.68E-05)$-$} & \multicolumn{1}{l}{1.81E+02 (2.75E-05)$-$} & \multicolumn{1}{l}{1.81E+02 (1.08E-05)$-$} & \multicolumn{1}{l}{1.81E+02 (7.25E-06)$-$} & \multicolumn{1}{l}{1.81E+02 (9.14E-06)$-$} & \multicolumn{1}{l}{1.81E+02 (1.24E-05)$-$} & \multicolumn{1}{l}{1.81E+02 (1.01E-04)$-$} \\
$F_{10}$ & \multicolumn{1}{l}{\textbf{0.00E+00 (0.00E+00)}} & \multicolumn{1}{l}{1.00E+02 (3.21E-02)$-$} & \multicolumn{1}{l}{1.00E+02 (2.58E-02)$-$} & \multicolumn{1}{l}{1.00E+02 (3.51E-02)$-$} & \multicolumn{1}{l}{1.00E+02 (1.80E-02)$-$} & \multicolumn{1}{l}{1.00E+02 (1.80E-02)$-$} & \multicolumn{1}{l}{1.00E+02 (1.21E-02)$-$} & \multicolumn{1}{l}{1.00E+02 (4.13E-02)$-$} \\
$F_{11}$ & \multicolumn{1}{l}{\textbf{7.28E-05 (3.06E-04)}} & \multicolumn{1}{l}{3.97E+02 (1.80E+01)$-$} & \multicolumn{1}{l}{3.26E+02 (4.45E+01)$-$} & \multicolumn{1}{l}{3.16E+02 (3.74E+01)$-$} & \multicolumn{1}{l}{3.81E+02 (4.02E+01)$-$} & \multicolumn{1}{l}{3.16E+02 (3.74E+01)$-$} & \multicolumn{1}{l}{3.77E+02 (4.25E+01)$-$} & \multicolumn{1}{l}{3.19E+02 (9.80E+01)$-$} \\
$F_{12}$ & \multicolumn{1}{l}{\textbf{2.29E+02 (6.08E-01)}} & \multicolumn{1}{l}{2.35E+02 (2.45E+00)$-$} & \multicolumn{1}{l}{2.34E+02 (2.55E+00)$-$} & \multicolumn{1}{l}{2.30E+02 (1.99E+00)$-$} & \multicolumn{1}{l}{2.30E+02 (1.40E+00)$-$} & \multicolumn{1}{l}{2.32E+02 (1.09E+00)$-$} & \multicolumn{1}{l}{2.33E+02 (2.27E+00)$-$} & \multicolumn{1}{l}{2.38E+02 (3.81E+00)$-$} \\
\midrule
$+$ / $\approx$ / $-$ & --    & 0/3/9 & 0/2/10 & 0/3/9 & 1/3/8 & 0/4/8 & 1/3/8 & 0/3/9 \\
\bottomrule
\end{tabular}%
}
\footnotesize
\textsuperscript{*} The Wilcoxon rank-sum tests (with a significance level of 0.05) were conducted between MetaDE and each individually.
The final row displays the number of problems where the corresponding algorithm performs statistically better ($+$),  similar ($\thickapprox$), or worse ($-$) compared to MetaDE.\\
\label{tab:NP10000 20D_supp}%
}
\end{table}%

% \subsection{Supplementary results}\label{section:FEs}
 % This experiment leveraged parallel GPU computation and constrained the runtime: 30 seconds for the 10-dimensional problems and 60 seconds for the 20-dimensional problems. 
 
% We recorded the number of FEs each algorithm achieved within 60s, which are presented in Table \ref{tab:FEs}. 
 
%  In addition, when the population size of the comparative DE variants was increased to 10,000 (same level of concurrency as MetaDE), the FEs achieved by all algorithms are displayed in Table \ref{tab:FEs NP10000}.

% The results show that MetaDE achieved considerably more FEs within a given time than the other algorithms. This demonstrates that MetaDE has a high degree of parallelism, making it particularly well-suited for GPU computing.

\begin{table}[htbp]
  \centering
  \caption{The number of FEs achieved by each algorithm within \SI{60}{\second}. The peer DE variants are set with population size of 1,000.}
  {
           \renewcommand{\arraystretch}{1}
 \renewcommand{\tabcolsep}{10pt}
% Table generated by Excel2LaTeX from sheet 'Exp2 NP10000'
\begin{tabular}{cccccccccc}
\toprule
Dim   & Func  & MetaDE & DE    & SaDE  & JaDE  & CoDE  & SHADE &LSHADE-RSP&EDEV\\
\midrule
\multirow{12}[2]{*}{10D} & $F_{1}$ & \textbf{1.85E+09} & 3.91E+07 & 4.38E+06 & 6.26E+06 & 7.67E+07 & 4.62E+06&1.92E+07&1.91E+07 \\
      & $F_{2}$ & \textbf{1.84E+09} & 4.05E+07 &4.35E+06 & 6.20E+06 & 7.61E+07 & 4.63E+06&1.89E+07&1.96E+07 \\
      & $F_{3}$ & \textbf{1.50E+09} & 3.63E+07 & 4.31E+06 &6.18E+06& 6.87E+07& 4.54E+06&1.84E+07 &1.89E+07\\
      & $F_{4}$ & \textbf{1.84E+09} &3.77E+07 & 4.27E+06 & 6.16E+06 &7.42E+07 & 4.64E+06&1.89E+07&1.98E+07 \\
      & $F_{5}$ & \textbf{1.83E+09} & 3.85E+07 &4.33E+06 & 6.16E+06 & 7.50E+07& 4.64E+06&1.85E+07& 1.99E+07\\
      & $F_{6}$ & \textbf{1.84E+09} &3.71E+07 & 4.30E+06 & 6.01E+06& 7.38E+07 & 4.61E+06& 1.86E+07&1.99E+07\\
      & $F_{7}$ & \textbf{1.74E+09} &3.71E+07 & 4.15E+06 & 5.84E+06 & 7.05E+07& 4.37E+06&1.77E+07&1.91E+07\\
      & $F_{8}$ & \textbf{1.72E+09} & 3.69E+07 & 4.02E+06 & 5.75E+06 & 6.82E+07 & 4.35E+06& 1.71E+07&1.93E+07\\
      & $F_{9}$ & \textbf{1.78E+09} &3.81E+07 &4.35E+06 &6.10E+06 & 6.84E+07 & 4.56E+06&1.84E+07 &1.92E+07\\
      & $F_{10}$ & \textbf{1.44E+09} & 3.50E+07&4.12E+06 &  5.77E+06 & 6.94E+07 & 4.41E+06&1.76E+07& 1.83E+07\\
      & $F_{11}$ & \textbf{1.46E+09} & 3.43E+07& 4.11E+06 & 5.87E+06 & 6.89E+07 & 4.36E+06 &1.69E+07&1.83E+07\\
      & $F_{12}$ & \textbf{1.43E+09} &3.37E+07 & 3.97E+06 & 5.81E+06 & 6.56E+07 & 4.32E+06 &1.72E+07&1.92E+07\\
\midrule
\midrule
\multirow{12}[2]{*}{20D} & $F_{1}$ & \textbf{1.66E+09} & 3.93E+07 &4.29E+06 & 5.94E+06 & 7.20E+07 & 4.62E+06&1.87E+07&1.94E+07 \\
      & $F_{2}$ & \textbf{1.66E+09} & 3.89E+07 &4.17E+06 & 5.94E+06 & 7.33E+07 & 4.61E+06&1.91E+07& 1.95E+07\\
      & $F_{3}$ & \textbf{1.18E+09} & 3.38E+07 & 4.16E+06& 5.93E+06 &  6.77E+07& 4.49E+06& 1.72E+07&1.85E+07\\
      & $F_{4}$ & \textbf{1.65E+09} &3.56E+07 & 4.27E+06 & 5.85E+06 & 7.05E+07& 4.63E+06& 1.90E+07&1.94E+07\\
      & $F_{5}$ & \textbf{1.64E+09} & 3.88E+07 & 4.20E+06 & 6.04E+06& 7.16E+07& 4.62E+06&1.86E+07&1.95E+07 \\
      & $F_{6}$ & \textbf{1.64E+09} & 3.73E+07 & 4.25E+06 & 5.84E+06 & 7.10E+07 & 4.61E+06&1.86E+07& 1.92E+07\\
      & $F_{7}$ & \textbf{1.45E+09} & 3.13E+07 & 3.87E+06 & 5.27E+06 & 6.40E+07 & 4.26E+06& 1.72E+07&1.79E+07\\
      & $F_{8}$ & \textbf{1.44E+09} & 3.08E+07 & 3.53E+06& 5.29E+06 & 5.98E+07& 4.09E+06&1.69E+07&1.70E+07 \\
      & $F_{9}$ & \textbf{1.57E+09} & 3.62E+07 &4.05E+06 & 6.15E+06 &  6.91E+07& 4.54E+06&1.84E+07& 1.89E+07\\
      & $F_{10}$ & \textbf{9.80E+08} &2.94E+07 & 3.74E+06 & 5.28E+06& 5.69E+07& 4.12E+06& 1.54E+07& 1.63E+07\\
      & $F_{11}$ & \textbf{1.00E+09} & 2.37E+07 &3.76E+06 & 5.38E+06&5.86E+07& 4.04E+06&1.59E+07& 1.64E+07 \\
      & $F_{12}$ & \textbf{9.90E+08} & 2.33E+07 & 3.76E+06 &5.39E+06&5.47E+07& 4.09E+06& 1.57E+07& 1.60E+07\\
\bottomrule
\end{tabular}%
}
  \label{tab:FEs NP10000_supp}%
\end{table}%

\clearpage

\section{Supplementary information for the application of robotics control}\label{section_brax_supp}

\subsection{Illustrations of the robotics control tasks}\label{section_brax_image_supp}

% Figs. \ref{Figure_swimmer}-\ref{Figure_hopper}, illustrate the three robotics control tasks from the Brax \cite{brax} reinforcement learning library: swimmer, reacher, and hopper.

% \begin{enumerate}
%     \item Swimmer: A serpentine agent that must coordinate joint movements to navigate through a fluid environment, aiming for efficient propulsion and forward movement.
%     \item Reacher: A robotic arm environment where the goal is to control joint torques to reach a target point with precision, testing the fine control of the learning algorithm.
%     \item Hopper: A one-legged robot that must learn to balance and hop forward as far and as fast as possible, providing a benchmark for locomotion and stability in dynamic environments.
% \end{enumerate}

\begin{figure}[!h]
\centering
\includegraphics[width=6cm, height=3cm]{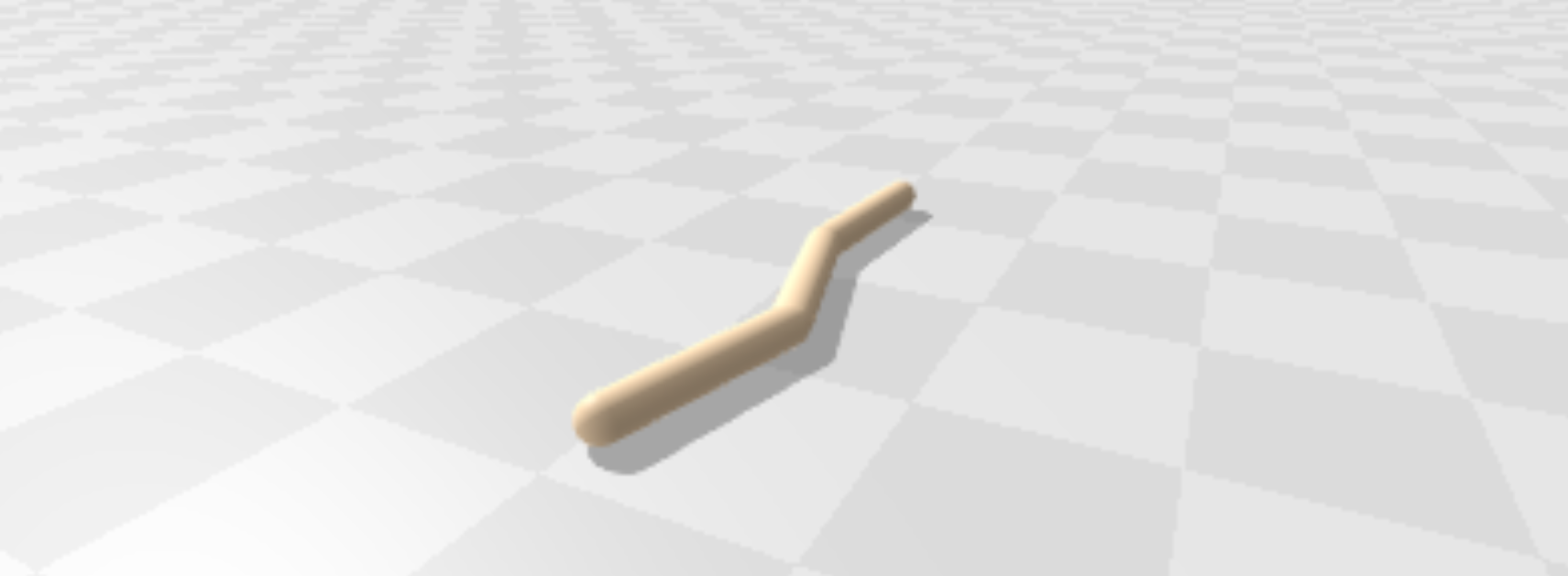}
\caption{The swimmer task in Brax. It is designed to simulate a multi-jointed creature navigating through a fluid medium.}
\label{Figure_swimmer_supp}
\end{figure}

\begin{figure}[!h]
\centering
\includegraphics[width=6cm, height=3cm]{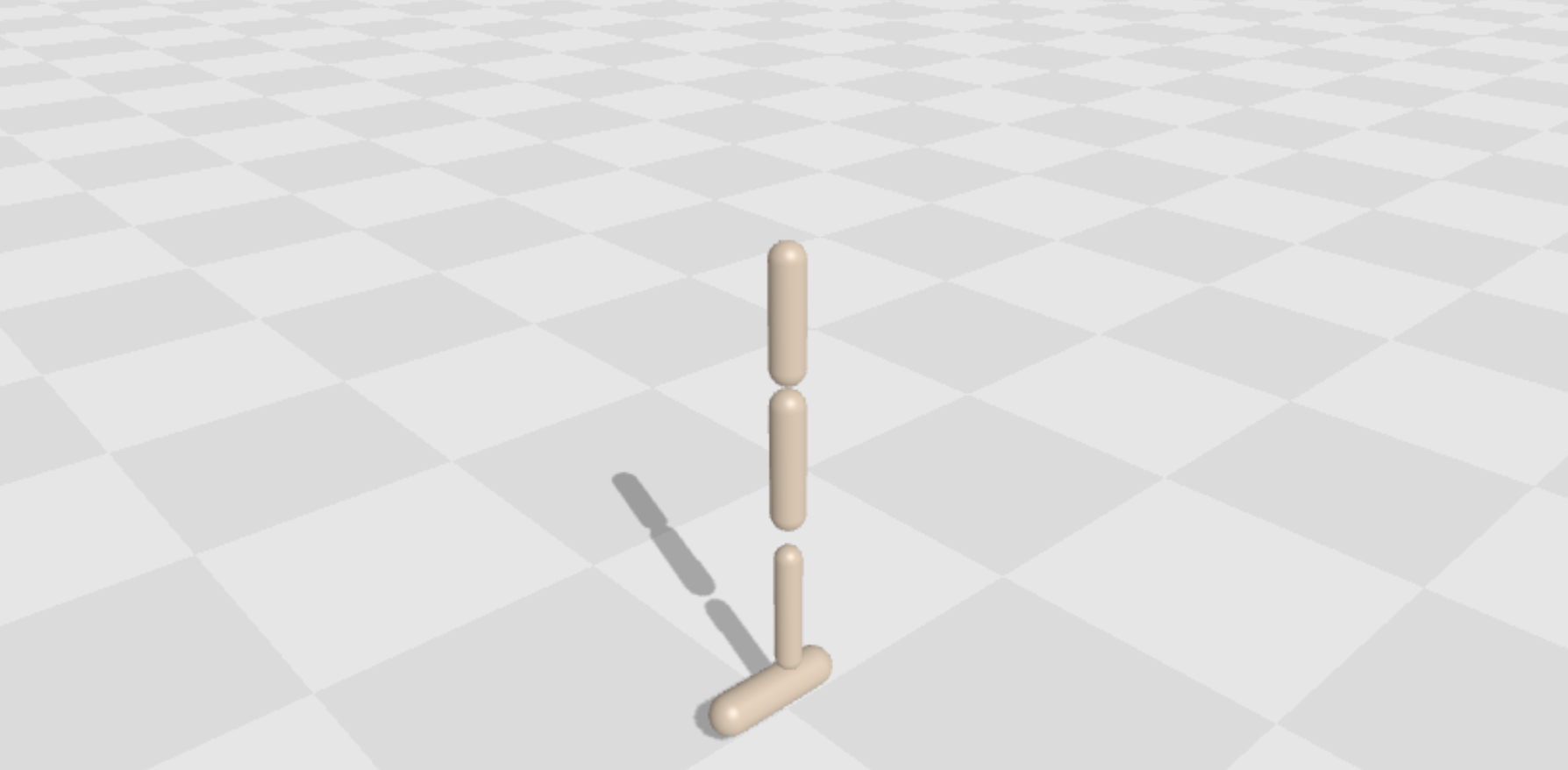}
\caption{The hopper task in Brax. It resembles a one-legged robotic creature with the objective to hop forward smoothly and quickly.}
\label{Figure_hopper_supp}
\end{figure}

\begin{figure}[!h]
\centering
\includegraphics[width=6cm, height=3cm]{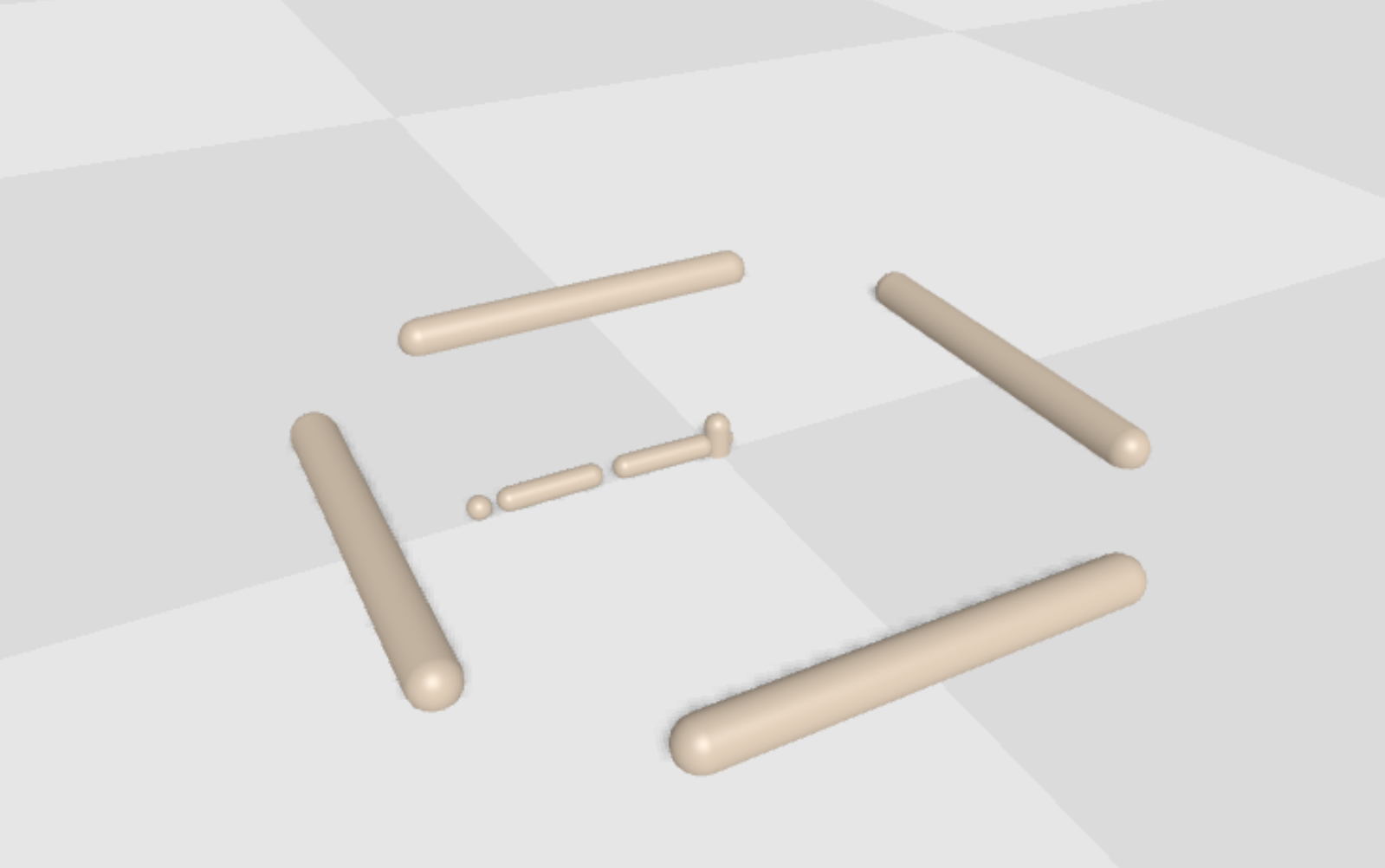}
\caption{The reacher task in Brax. It simulates a robotic arm tasked with reaching a target location.}
\label{Figure_reacher_supp}
\end{figure}

\subsection{Supplementary results}\label{section_brax_results_supp}
% Table \ref{tab:comparative-reward-analysis} displays the results of the neuroevolution experimrnt for 60 minutes.

\begin{table}[htbp]
\centering
\caption{Rewards achieved by MetaDE and peer evolutionary algorithms on the robotics tasks. 
The mean and standard deviation (in parentheses) of the results over multiple runs are displayed in pairs. 
Results with the best mean values are highlighted.
}
\label{tab:comparative-reward-analysis_supp}
{%
 \renewcommand{\arraystretch}{1}
\renewcommand{\tabcolsep}{3pt}
% Table generated by Excel2LaTeX from sheet 'Exp6 brax'
\begin{tabular}{cccccccc}
\toprule
Task & MetaDE  & \textbf{CSO } & \textbf{CMAES } & \textbf{SHADE } & \textbf{DE}&\textbf{LSHADE-RSP}&\textbf{EDEV} \\
\midrule
Swimmer  & \textbf{ 190.85 (2.39) } &  183.45 (1.15)  &  186.07 (2.68)  & 185.90 (3.31) &  182.15 (2.54) &186.36 (1.21)&145.52 (41.22)\\
%\midrule
Hopper   &  1187.53 (122.72) & 1330.45 (325.55) &  \textbf{1389.72 (474.41)}  &  1022.25 (122.94)  &  871.06 (147.12)&1102.23 (100.33)&457.22 (80.16) \\
%\midrule
Reacher  &  -21.05 (5.05)  &  -504.07 (112.18)  & \textbf{ -3.76 (1.05) } &  -343.18 (105.65)  &  -522.99 (142.78)    &    -342.74 (36.63)  &  -493.46 (173.67)\\
\bottomrule
\end{tabular}%

}
\end{table}

\clearpage
\section{More Evolvers}\label{subsection Evolver Comparison_supp}
We selected \texttt{DE/rand/1/bin} as the evolver for its simplicity and adaptability, hypothesizing its capability to self-evolve. Nonetheless, it is also interesting to evaluate the performance implications of utilizing other EAs as evolvers.

In this experiment, we maintained the identical framework and parameter settings for MetaDE as utilized in the primary experiments. 
The distinction lies in the comparative evaluation of the effectiveness of DE, PSO, Natural Evolution Strategies (NES) \citesupp{NES}, and Competitive Swarm Optimization (CSO) \citesupp{CSO} as evolvers. 
For PSO, parameters were set to the recommended values of $w=0.729$ and $c_1=c_2=1.49$ \citesupp{PSOParamSetting}. 
It is important to note that CSO and NES do not require parameter tuning.

As shown in Tables \ref{tab:diffTunner 10D}-\ref{tab:diffTunner 20D} and Figs. \ref{Figure_evolver_10D}-\ref{Figure_evolver_20D}, the experimental outcomes suggest minimal performance disparities among the EAs when employed as evolvers. 
Specifically, DE demonstrates a marginal advantage in 10-dimensional problems, whereas the performance is comparably uniform across all evolvers for 20-dimensional problems. 
These results imply that the selection of an evolver within the MetaDE framework is relatively flexible. 
Such findings highlight the adaptability and flexibility of MetaDE, illustrating that its efficiency is not significantly influenced by the particular choice of evolver.

\begin{table}[htbp]
  \centering
  \caption{
  Performance of MetaDE with different evolvers on 10D problems in CEC2022 benchmark suite. 
The mean and standard deviation (in parentheses) of the results over multiple runs are displayed in pairs. 
Results with the best mean values are highlighted.
  }
  {
         \renewcommand{\arraystretch}{1}
 \renewcommand{\tabcolsep}{10pt}
% Table generated by Excel2LaTeX from sheet 'Exp5 evolver'
\begin{tabular}{ccccc}
\toprule
Func  & DE    & PSO   & NES   & CSO \\
\midrule
$F_{1}$ & \textbf{0.00E+00 (0.00E+00)} & \boldmath{}\textbf{0.00E+00 (0.00E+00)$\approx$}\unboldmath{} & \boldmath{}\textbf{0.00E+00 (0.00E+00)$\approx$}\unboldmath{} & \boldmath{}\textbf{0.00E+00 (0.00E+00)$\approx$}\unboldmath{} \\
$F_{2}$ & \textbf{0.00E+00 (0.00E+00)} & \boldmath{}\textbf{0.00E+00 (0.00E+00)$\approx$}\unboldmath{} & 1.60E-06 (2.82E-06)$-$ & \boldmath{}\textbf{0.00E+00 (0.00E+00)$\approx$}\unboldmath{} \\
$F_{3}$ & \textbf{0.00E+00 (0.00E+00)} & \boldmath{}\textbf{0.00E+00 (0.00E+00)$\approx$}\unboldmath{} & \boldmath{}\textbf{0.00E+00 (0.00E+00)$\approx$}\unboldmath{} & \boldmath{}\textbf{0.00E+00 (0.00E+00)$\approx$}\unboldmath{} \\
$F_{4}$ & \textbf{0.00E+00 (0.00E+00)} & 1.41E-03 (7.70E-03)$\approx$ & 3.23E-02 (1.76E-01)$\approx$ & \boldmath{}\textbf{0.00E+00 (0.00E+00)$\approx$}\unboldmath{} \\
$F_{5}$ & \textbf{0.00E+00 (0.00E+00)} & \boldmath{}\textbf{0.00E+00 (0.00E+00)$\approx$}\unboldmath{} & \boldmath{}\textbf{0.00E+00 (0.00E+00)$\approx$}\unboldmath{} & \boldmath{}\textbf{0.00E+00 (0.00E+00)$\approx$}\unboldmath{} \\
$F_{6}$ & \textbf{5.50E-04 (3.96E-04)} & 1.45E-03 (1.61E-03)$\approx$ & 1.66E-03 (1.79E-03)$-$ & 1.37E-03 (9.80E-04)$-$ \\
$F_{7}$ & \textbf{0.00E+00 (0.00E+00)} & \boldmath{}\textbf{0.00E+00 (0.00E+00)$\approx$}\unboldmath{} & \boldmath{}\textbf{0.00E+00 (0.00E+00)$\approx$}\unboldmath{} & \boldmath{}\textbf{0.00E+00 (0.00E+00)$\approx$}\unboldmath{} \\
$F_{8}$ & 5.52E-03 (4.41E-03) & \textbf{1.77E-06 (1.48E-06)$+$} & 1.83E-02 (1.09E-02)$-$ & 1.57E-03 (2.26E-03)$+$ \\
$F_{9}$ & \textbf{3.36E+00 (1.77E+01)} & 1.40E+01 (4.63E+01)$-$ & 2.27E+01 (4.18E+01)$\approx$ & 1.75E+01 (4.86E+01)$-$ \\
$F_{10}$ & \textbf{0.00E+00 (0.00E+00)} & \boldmath{}\textbf{0.00E+00 (0.00E+00)$\approx$}\unboldmath{} & \boldmath{}\textbf{0.00E+00 (0.00E+00)$\approx$}\unboldmath{} & 7.38E-02 (3.99E-01)$\approx$ \\
$F_{11}$ & \textbf{0.00E+00 (0.00E+00)} & \boldmath{}\textbf{0.00E+00 (0.00E+00)$\approx$}\unboldmath{} & \boldmath{}\textbf{0.00E+00 (0.00E+00)$\approx$}\unboldmath{} & \boldmath{}\textbf{0.00E+00 (0.00E+00)$\approx$}\unboldmath{} \\
$F_{12}$ & \textbf{1.39E+02 (4.63E+01)} & 1.48E+02 (3.39E+01)$\approx$ & 1.59E+02 (6.82E-06)$-$ & 1.54E+02 (1.78E+01)$\approx$ \\
\midrule
$+$ / $\approx$ / $-$ & --    & 1/10/1 & 4/8/0 & 1/9/2 \\
\bottomrule
\end{tabular}%
}

\footnotesize
\textsuperscript{*} The Wilcoxon rank-sum tests (with a significance level of 0.05) were conducted between MetaDE and each individually.
The final row displays the number of problems where the corresponding evlover performs statistically better ($+$),  similar ($\thickapprox$), or worse ($-$) compared to DE.\\
\label{tab:diffTunner 10D_supp}%
\end{table}%

\begin{table}[htbp]
  \centering
  \caption{
  Performance of MetaDE with different evolvers on 20D problems in CEC2022 benchmark suite.
The mean and standard deviation (in parentheses) of the results over multiple runs are displayed in pairs. 
Results with the best mean values are highlighted.
  }
  {
        \renewcommand{\arraystretch}{1}
 \renewcommand{\tabcolsep}{10pt}
% Table generated by Excel2LaTeX from sheet 'Exp5 evolver'
\begin{tabular}{ccccc}
\toprule
Func  & DE    & PSO   & NES   & CSO \\
\midrule
$F_{1}$ & \multicolumn{1}{l}{\textbf{0.00E+00 (0.00E+00)}} & \multicolumn{1}{l}{\boldmath{}\textbf{0.00E+00 (0.00E+00)$\approx$}\unboldmath{}} & \multicolumn{1}{l}{\boldmath{}\textbf{0.00E+00 (0.00E+00)$\approx$}\unboldmath{}} & \multicolumn{1}{l}{\boldmath{}\textbf{0.00E+00 (0.00E+00)$\approx$}\unboldmath{}} \\
$F_{2}$ & \multicolumn{1}{l}{1.26E-02 (3.74E-02)} & \multicolumn{1}{l}{4.28E-02 (1.25E-01)$\approx$} & \multicolumn{1}{l}{1.11E+00 (1.30E+00)$-$} & \multicolumn{1}{l}{\textbf{0.00E+00 (0.00E+00)$+$}} \\
$F_{3}$ & \multicolumn{1}{l}{\textbf{0.00E+00 (0.00E+00)}} & \multicolumn{1}{l}{\boldmath{}\textbf{0.00E+00 (0.00E+00)$\approx$}\unboldmath{}} & \multicolumn{1}{l}{\boldmath{}\textbf{0.00E+00 (0.00E+00)$\approx$}\unboldmath{}} & \multicolumn{1}{l}{\boldmath{}\textbf{0.00E+00 (0.00E+00)$\approx$}\unboldmath{}} \\
$F_{4}$ & \multicolumn{1}{l}{\textbf{2.02E+00 (8.56E-01)}} & \multicolumn{1}{l}{2.51E+00 (1.19E+00)$-$} & \multicolumn{1}{l}{5.53E+00 (1.55E+00)$-$} & \multicolumn{1}{l}{2.85E+00 (1.48E+00)$\approx$} \\
$F_{5}$ & \multicolumn{1}{l}{\textbf{0.00E+00 (0.00E+00)}} & \multicolumn{1}{l}{\boldmath{}\textbf{0.00E+00 (0.00E+00)$\approx$}\unboldmath{}} & \multicolumn{1}{l}{\boldmath{}\textbf{0.00E+00 (0.00E+00)$\approx$}\unboldmath{}} & \multicolumn{1}{l}{\boldmath{}\textbf{0.00E+00 (0.00E+00)$\approx$}\unboldmath{}} \\
$F_{6}$ & \multicolumn{1}{l}{1.16E-01 (2.79E-02)} & \multicolumn{1}{l}{1.81E-01 (2.42E-01)$-$} & \multicolumn{1}{l}{9.32E-02 (2.65E-02)$+$} & \multicolumn{1}{l}{\boldmath{}\textbf{1.10E-01 (2.85E-02)$\approx$}\unboldmath{}} \\
$F_{7}$ & \multicolumn{1}{l}{5.17E-02 (6.21E-02)} & \multicolumn{1}{l}{1.61E-01 (3.84E-01)$\approx$} & \multicolumn{1}{l}{8.13E-01 (6.01E-01)$-$} & \multicolumn{1}{l}{\textbf{6.37E-02 (2.03E-01)$+$}} \\
$F_{8}$ & \multicolumn{1}{l}{\textbf{7.19E-01 (1.02E+00)}} & \multicolumn{1}{l}{2.31E+00 (4.16E+00)$\approx$} & \multicolumn{1}{l}{1.82E+01 (3.82E+00)$-$} & \multicolumn{1}{l}{2.19E+00 (4.05E+00)$\approx$} \\
$F_{9}$ & \multicolumn{1}{l}{\textbf{1.07E+02 (1.98E+01)}} & \multicolumn{1}{l}{1.34E+02 (3.72E+01)$-$} & \multicolumn{1}{l}{1.81E+02 (3.75E-06)$-$} & \multicolumn{1}{l}{1.75E+02 (1.81E+01)$-$} \\
$F_{10}$ & \multicolumn{1}{l}{\textbf{0.00E+00 (0.00E+00)}} & \multicolumn{1}{l}{\boldmath{}\textbf{0.00E+00 (0.00E+00)$\approx$}\unboldmath{}} & \multicolumn{1}{l}{\boldmath{}\textbf{0.00E+00 (0.00E+00)$\approx$}\unboldmath{}} & \multicolumn{1}{l}{7.86E-01 (2.39E+00)$-$} \\
$F_{11}$ & \multicolumn{1}{l}{7.28E-05 (3.06E-04)} & \multicolumn{1}{l}{1.01E-02 (3.90E-02)$\approx$} & \multicolumn{1}{l}{8.79E-03 (1.52E-02)$-$} & \multicolumn{1}{l}{\textbf{1.45E-05 (0.00E+00)$-$}} \\
$F_{12}$ & \multicolumn{1}{l}{\textbf{2.29E+02 (6.08E-01)}} & \multicolumn{1}{l}{2.29E+02 (1.17E+00)$\approx$} & \multicolumn{1}{l}{2.31E+02 (8.77E-01)$-$} & \multicolumn{1}{l}{2.30E+02 (9.60E-01)$-$} \\
\midrule
$+$ / $\approx$ / $-$ & --    & 0/9/3 & 1/4/7 & 2/6/4 \\
\bottomrule
\end{tabular}%
}

\footnotesize
\textsuperscript{*} The Wilcoxon rank-sum tests (with a significance level of 0.05) were conducted between MetaDE and each individually.
The final row displays the number of problems where the corresponding evlover performs statistically better ($+$),  similar ($\thickapprox$), or worse ($-$) compared to DE.\\
\label{tab:diffTunner 20D_supp}%
\end{table}%

\clearpage

\begin{figure*}[htpb]
\centering
\includegraphics[scale=0.29]{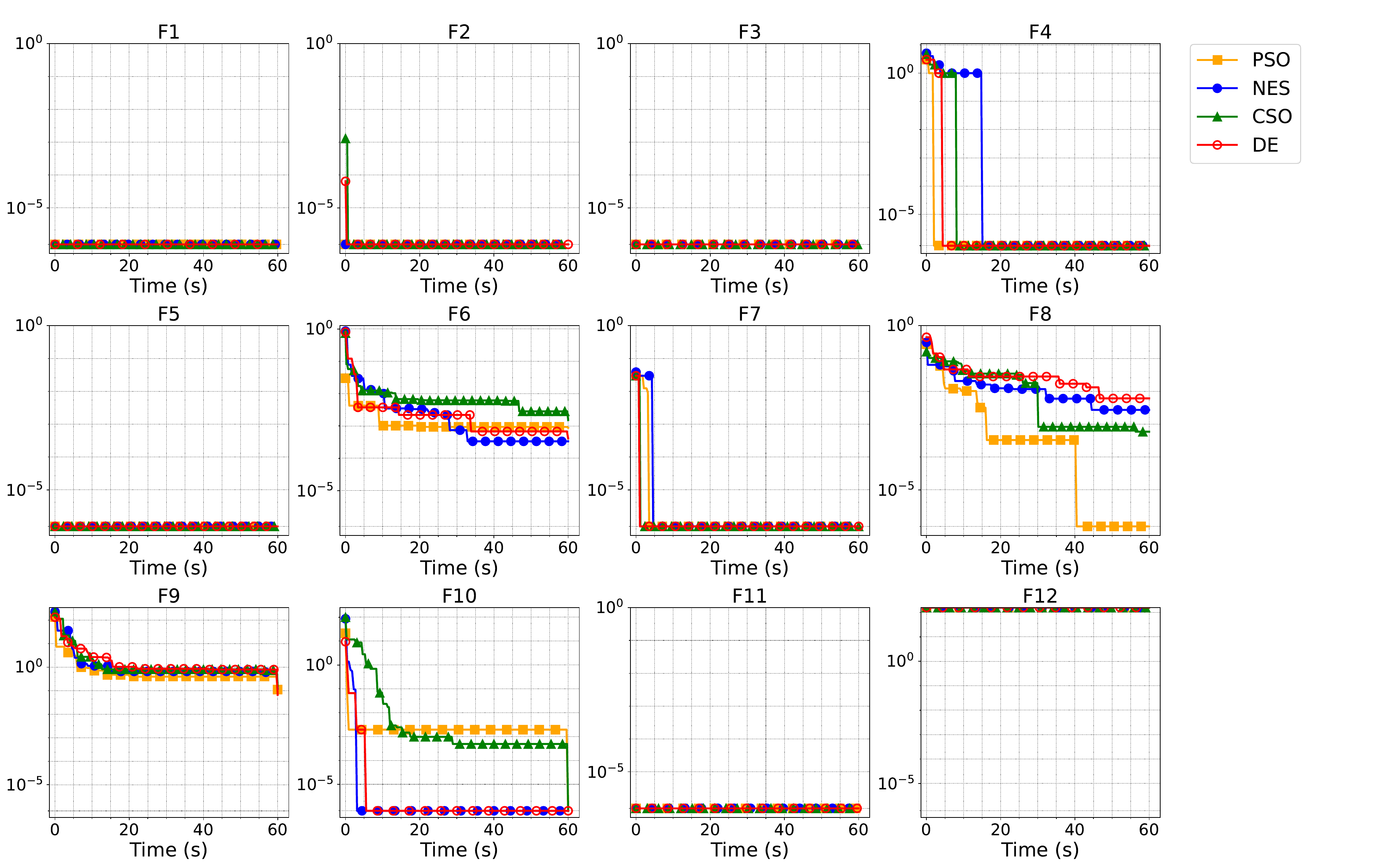}
\caption{Convergence curves with different evolvers on 10D problems in CEC2022 benchmark suite.}
\label{Figure_evolver_10D_supp}
\end{figure*}

\begin{figure*}[htpb]
\centering
\includegraphics[scale=0.29]{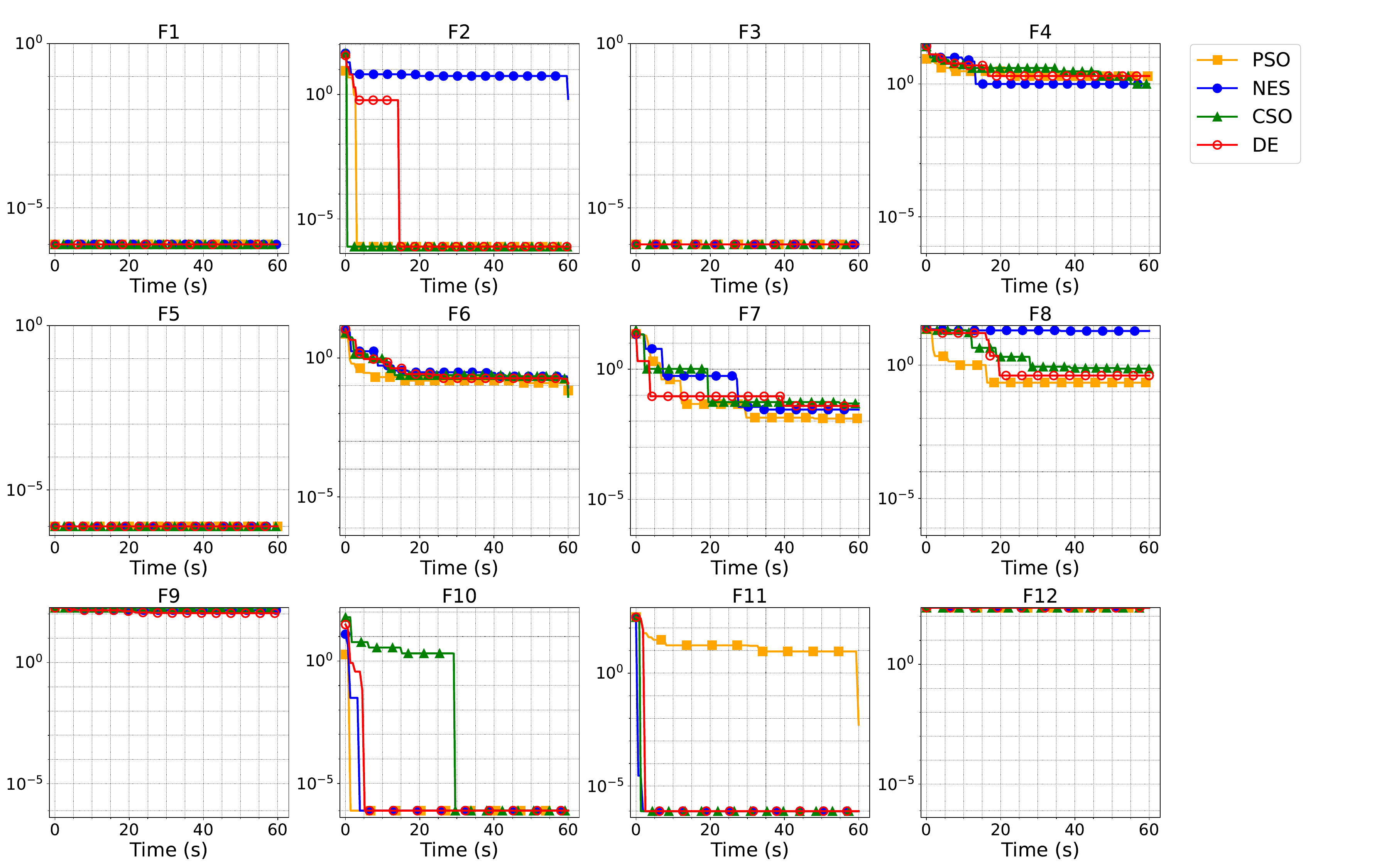}
\caption{Convergence curves with different evolvers on 20D problems in CEC2022 benchmark suite.}
\label{Figure_evolver_20D_supp}
\end{figure*}

\clearpage
% \bibliography{Supplement_references}

% \bibliographystylesupp{IEEEtran}
% \bibliographysupp{Supplement_references}

\end{document}